\documentclass{article} 
\usepackage{iclr2027_arxiv,times}
\iclrfinalcopy

\usepackage{amsmath,amsfonts,bm}

\def\eqref#1{equation~\ref{#1}}

\def\1{\bm{1}}

\DeclareMathAlphabet{\mathsfit}{\encodingdefault}{\sfdefault}{m}{sl}
\SetMathAlphabet{\mathsfit}{bold}{\encodingdefault}{\sfdefault}{bx}{n}

\usepackage{hyperref}
\usepackage{url}
\usepackage{graphicx}
\usepackage{xcolor}
\usepackage{booktabs}
\usepackage{tabularx}
\usepackage{longtable}
\usepackage{float}
\usepackage{amsmath}
\usepackage{pifont}
\newcommand{\cmark}{{\color{green!45!black}\ding{51}}}
\newcommand{\xmark}{{\color{red!65!black}\ding{55}}}

\newcommand{\runningheadfont}{\normalsize\scshape}
\renewcommand{\headrulewidth}{0.5pt}

\renewcommand{\subsectionmark}[1]{\markright{\thesubsection}}
\renewcommand{\bibsection}{\section*{\refname}\markboth{\refname}{}}
\fancypagestyle{titlepage}{\fancyhf{}\fancyfoot[C]{\thepage}\renewcommand{\headrulewidth}{0pt}}

\title{Harness-Bench: Evaluating Agent Harnesses across Models and Realistic Workflows}

\author{
\textbf{Yilun Yao}$^{1,*}$,
\textbf{Xinyu Tan}$^{1,*}$,
\textbf{Chao-Hsuan Liu}$^{1,*}$,
\textbf{Yaoming Li}$^{1,*}$, \\
\textbf{Zhengyang Wang}$^{1}$,
\textbf{Zongwei Lv}$^{1}$,
\textbf{Wenhan Yu}$^{1}$,
\textbf{Zhewen Tan}$^{1}$,
\textbf{Yuxuan Tian}$^{1}$, \\
\textbf{Lin Sun}$^{2}$,
\textbf{Xiangzheng Zhang}$^{2}$,
\textbf{Guangxiang Zhao}$^{2,\dagger}$,
\textbf{Tong Yang}$^{1,\dagger}$ \\[3pt]
$^{1}$Peking University \qquad
$^{2}$Qiyuan Tech
}

\newcommand{\bench}{Harness-Bench}

\begin{document}
\raggedbottom

\maketitle
\thispagestyle{titlepage}
\fancyhead[L]{\runningheadfont\leftmark}

\begingroup
\renewcommand{\thefootnote}{\fnsymbol{footnote}}
\footnotetext[1]{These authors contributed equally.}
\footnotetext[2]{Corresponding authors: Guangxiang Zhao and Tong Yang. For questions or inquiries, please contact \texttt{yilunyao@stu.pku.edu.cn}, \texttt{zhaoguangxiang@pku.edu.cn}, or \texttt{yangtong@pku.edu.cn}.}
\endgroup

\begin{abstract}

Agent harnesses shape how language models turn instructions into completed work by governing tool use, context management, and execution. Choosing a harness requires knowing what harnesses enable the same model to deliver, and at what cost. We introduce \textbf{\bench}, a benchmark and reusable evaluation pipeline for comparing native harnesses on realistic work tasks. The suite comprises 300 sandboxed tasks from log-based and domain-based synthesis, with input resources, deliverables, and operational constraints. Task-specific oracles score final artifacts, awarding partial credit for completed modules under safety and risk gates. The pipeline fixes external task conditions and scoring rules, preserves native harness behavior, and records token usage and API cost. We evaluate seven harnesses with ten backends on general tasks and seven on multimodal tasks, using paired comparisons on shared tasks. Our evaluation reveals substantial, model-dependent performance differences: no harness performs best across all ten backends, and GLM-5.3's general-task full-credit rate varies by 13.5 percentage points across harnesses. Harness choices also change with the delivery objective: configurations that lead mean score need not lead full-credit attainment. Higher expenditure does not ensure better delivery: with GPT-5.6 Sol, Nanobot incurs 2.97 times OpenClaw's API cost but has a lower full-credit rate. Stratified trajectory analysis and paired cases explain delivery gaps: corrected tool calls can leave workflows unfinished, and mutually consistent outputs can omit required evidence. These findings link model-dependent harness performance to execution steps that turn partial progress into delivered work. Harness-Bench brings task outcomes, unmet requirements, and resource costs into a common evaluation framework, providing an empirical foundation for selecting and improving agent harnesses.
\footnote{Our code and data are available at \url{https://github.com/Qihoo360/harness-bench}. Additional resources and updates can be found on our project website at \url{http://www.harness-bench.ai/}.}

\end{abstract}

\section{Introduction}

\label{sec:introduction}

Agent harnesses shape how language models turn instructions into completed work. They organize prompts and context, expose tools, manage execution state, and handle errors~\citep{yao2023reactsynergizingreasoningacting,yang2024sweagentagentcomputerinterfacesenable}. These functions matter when work spans files, tools, and intermediate results. Preparing a business report, for example, may require inspecting source records, updating a spreadsheet, and reflecting those updates in the report. Practical agent capabilities therefore depend on both the model and its harness.

Existing benchmarks examine task completion, execution reliability, and system-level trade-offs. SWE-bench~\citep{jimenez2024swebenchlanguagemodelsresolve} and Terminal-Bench~\citep{merrill2026terminalbenchbenchmarkingagentshard} assess executable tasks, while Claw-Eval~\citep{ye2026clawevaltrustworthyevaluationautonomous} examines completion, safety, and robustness. HAL~\citep{kapoor2025holisticagentleaderboardmissing} demonstrates that scaffold choice affects accuracy and cost, and DuMateBench~\citep{niu2026dumatebenchevaluatingautonomousagents} runs all pairs of five harnesses and four models on tasks rebuilt from production user sessions, scoring artifacts mainly with an LLM judge. These studies establish the importance of the execution layer. What remains underexplored is how specific workflow requirements, verified without a model judge, change harness preferences within and across models: the harness that leads average quality may not complete the most tasks. Addressing this requires a shared task suite that connects explicit delivery requirements to paired executions and final artifacts.

Building on these observations, we introduce \textbf{\bench{}}, a benchmark and reusable pipeline centered on the execution demands of practical work. Its 300 sandboxed tasks span six categories and 17 subcategories, with explicit resources, deliverables, and constraints. Oracles award credit for completed work modules based on final artifacts, subject to safety and risk gates. Under common task conditions, the pipeline links module outcomes to execution traces and resource use, exposing which requirements different native harnesses fulfill or leave unfinished.

We evaluate seven harnesses with ten model backends on general tasks and seven on multimodal tasks. Harness choice substantially affects delivery: OpenCode exceeds Codex by 13.5 percentage points in general-task full-credit attainment with GLM-5.3. Higher-scoring models show less mean-score variation across harnesses, yet full-credit rates remain sensitive; optimizing complete delivery also changes harness choices and quality--cost frontiers. With GPT-5.6 Sol, Nanobot incurs 2.97 times OpenClaw's API cost while achieving a lower full-credit rate. Paired trajectories reveal why local progress can fail to yield delivery: repaired tool calls do not ensure execution resumes, and checks of produced files can miss omitted source requirements.

Our contributions are threefold: (1) a workflow benchmark with explicit deliverables and task-specific verification; (2) a reusable pipeline for comparing native model--harness configurations under common conditions; and (3) a cross-model study of delivery quality and cost, combining paired comparisons with stratified trajectory analysis to explain configuration strengths and recurring execution failures.

\section{Related Work}

\label{sec:related_work}

\paragraph{Agent Benchmarks.}

Existing agent benchmarks cover diverse forms of digital work. OSWorld~\citep{xie2024osworldbenchmarkingmultimodalagents} and WorkArena~\citep{drouin2024workarenacapablewebagents} evaluate computer-use and enterprise workflows, while $\tau$-bench~\citep{yao2024taubenchbenchmarktoolagentuserinteraction} focuses on policy-constrained user--agent interactions. AppWorld~\citep{trivedi2024appworldcontrollableworldapps} tests multi-application API workflows, and Terminal-Bench~\citep{merrill2026terminalbenchbenchmarkingagentshard} evaluates challenging command-line tasks. TheAgentCompany~\citep{xu2024theagentcompanybenchmarkingllmagents} simulates professional work in a software company, Claw-Eval~\citep{ye2026clawevaltrustworthyevaluationautonomous} examines task completion, safety, and robustness through execution evidence, and ClawBench~\citep{zhang2026clawbenchaiagentscomplete} evaluates everyday tasks on live websites.
DuMateBench~\citep{niu2026dumatebenchevaluatingautonomousagents} reconstructs workflows from real user sessions and evaluates model--harness pairs under imperfect environments. In contrast, \bench{} centers on harness design and selection, considering common execution demands such as tool orchestration, state management, and recovery, while analyzing delivery, cost, and risk across models. Unlike DuMateBench's LLM-heavy scoring (70\% judge weight), \bench{} uses executable, module-level checks with partial credit and hard safety and risk gates, as summarized in Table~\ref{tab:benchmark_comparison}.

\begin{table}[!htbp]
\centering
\caption{Comparison with representative agent benchmarks. Task counts follow the cited evaluations (WorkArena reports 33 task types). Multimodal tasks include non-text inputs or outputs. Fully controlled environments require reproducible task states without dependence on mutable external services.}
\label{tab:benchmark_comparison}
\small
\setlength{\tabcolsep}{2pt}
\renewcommand{\arraystretch}{1.12}
\resizebox{\linewidth}{!}{%
\begin{tabular}{@{}lrrcccccc@{}}
\toprule
\textbf{Benchmark} & \textbf{Tasks} & \textbf{Harnesses} & \shortstack{\textbf{LLM-free}\\\textbf{oracle}} & \shortstack{\textbf{From real}\\\textbf{user logs}} & \shortstack{\textbf{Partial}\\\textbf{credit}} & \shortstack{\textbf{All model--}\\\textbf{harness pairs}} & \shortstack{\textbf{Multimodal}\\\textbf{tasks}} & \shortstack{\textbf{Fully controlled}\\\textbf{environment}} \\
\midrule
OSWorld & 369 & 1 & \cmark & \xmark & \cmark & \xmark & \cmark & \xmark \\
WorkArena & 33 & 1 & \cmark & \xmark & \xmark & \xmark & \xmark & \cmark \\
$\tau$-bench & 165 & 1 & \cmark & \xmark & \xmark & \xmark & \xmark & \cmark \\
Terminal-Bench & 89 & 6 & \cmark & \xmark & \xmark & \xmark & \cmark & \cmark \\
AppWorld & 750 & 1 & \cmark & \xmark & \xmark & \xmark & \xmark & \cmark \\
TheAgentCompany & 175 & 2 & \xmark & \xmark & \cmark & \xmark & \xmark & \cmark \\
Claw-Eval & 300 & 1 & \xmark & \xmark & \cmark & \xmark & \cmark & \cmark \\
ClawBench & 153 & 1 & \xmark & \xmark & \xmark & \xmark & \xmark & \xmark \\
DuMateBench & 200 & 5 & \xmark & \cmark & \cmark & \cmark & \cmark & \cmark \\
\midrule
\textbf{Harness-Bench} & 300 & 7 & \cmark & \cmark & \cmark & \cmark & \cmark & \cmark \\
\bottomrule
\end{tabular}}
\end{table}

\paragraph{Agent Harnesses.}

ReAct~\citep{yao2023reactsynergizingreasoningacting} interleaves reasoning, actions, and observations; Reflexion~\citep{shinn2023reflexionlanguageagentsverbal} uses reflective memory to improve subsequent attempts; and SWE-agent~\citep{yang2024sweagentagentcomputerinterfacesenable} studies interfaces for navigation, editing, and testing. OpenClaw~\citep{openclaw2026software}, Codex CLI~\citep{openai2026codexcli}, and Claude Code~\citep{anthropic2026claudecode} make harness selection a practical deployment decision. HAL~\citep{kapoor2025holisticagentleaderboardmissing} standardizes model--scaffold evaluation across existing benchmarks, studying accuracy, cost, and execution logs. Workspace-Bench~\citep{tang2026workspacebench10benchmarkingai} pairs harnesses such as OpenClaw, Hermes, and LangChain Deep Agents with several models on workspace tasks graded by an agent judge, and AgencyBench~\citep{li2026agencybenchbenchmarkingfrontiersautonomous} finds that proprietary models perform best within their native agent ecosystems. \bench{} makes workflow requirements an explicit axis of harness comparison: a common suite couples task-design demand tags with module-level delivery checks and paired native-harness executions. This supports analysis of which model--harness advantages persist across requirements, where they reverse, and how execution leaves requested work unfinished.

\section{Harness-Bench}

\label{sec:harnessbench}

\begin{figure}[!htbp]

    \centering

    \includegraphics[width=\linewidth]{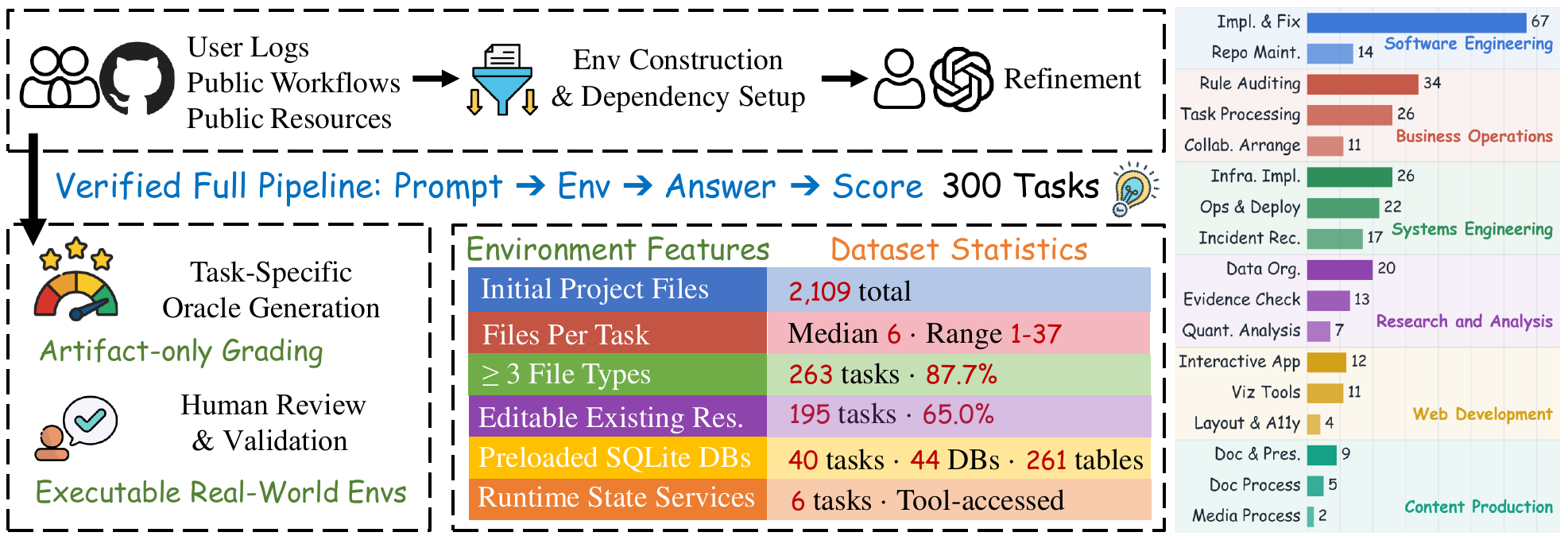}

    \caption{Task construction and composition of Harness-Bench. Work objectives and source materials guide environment construction, workflow refinement, and task-specific oracle development. The suite contains 300 tasks across six work categories and 17 subcategories.}

    \label{fig:task_construction}

\end{figure}

Harness-Bench connects practical work objectives to executable environments and verifiable outcomes. Figures~\ref{fig:task_construction} and~\ref{fig:evaluation_pipeline} summarize task construction and evaluation.

\subsection{Task Construction}

\label{sec:task_construction}

\paragraph{Task sources.}

We select work categories from common user workflows and ground task design in user logs, public workflows, and resource collections. Logs supply objectives and context; public workflows supply procedures and business rules; documents, repositories, and datasets supply task resources. Agents assist reconstruction and synthesis, allowing a source objective to be instantiated with public materials and synthetic records.

\paragraph{Environment construction.}

Each task reconstructs a workspace of files, repositories, records, or local services. Reconstruction preserves the information needed to act, the state changed by actions, and the artifacts required for handoff. External interactions are reproduced locally where needed, with fixed initial resources and explicit operational constraints.

\paragraph{Workflow refinement.}

Workflow refinement makes business dependencies explicit and testable: version selection, recovery of partial progress, duplicate avoidance, and consistency across outputs. In materials collection, a revised resume supersedes an earlier resume without invalidating an accompanying declaration; archives, registers, and reminders must reflect that distinction. Such dependencies expose whether a model--harness configuration carries the original work requirements through to delivery.

\paragraph{Validation.}

During construction, all 300 tasks undergo iterative human review and refinement of their environments and oracles, checking executability and alignment between scoring rules and stated requirements. Reference solutions are executed and receive full credit on every task. Appendix~\ref{app:construction_example} traces one work objective through reconstruction, refinement, and verification.

\subsection{Task Suite}

\label{sec:task_suite}

The suite contains 274 general and 26 multimodal tasks across six categories and 17 subcategories (Figure~\ref{fig:task_construction}). Each task receives a primary category and subcategory based on its objective and deliverables. Appendix~\ref{app:task_cards} presents representative tasks and oracle modules.

Across the 300 tasks, 1,408 scored modules assess delivery requirements, with three to seven modules per task. Work categories describe objectives and deliverables; overlapping execution-demand tags describe coordination requirements. A repository update and a business transaction can both require preserving progress and avoiding duplicate changes, supporting requirement-level comparisons across categories (Appendix~\ref{app:task_requirements}).

Figure~\ref{fig:task_construction} summarizes the task environments. Tasks combine project files with structured records, existing databases, and service state. In 195 tasks, mutation contracts permit scoped changes to existing resources; six tasks expose stateful services through task tools. These environments support workflows that must reconcile evidence, preserve unrelated state, and verify the effects of updates.

\subsection{Evaluation Protocol}

\label{sec:scoring_protocol}

\begin{figure}[!htbp]

    \centering

    \includegraphics[width=\linewidth]{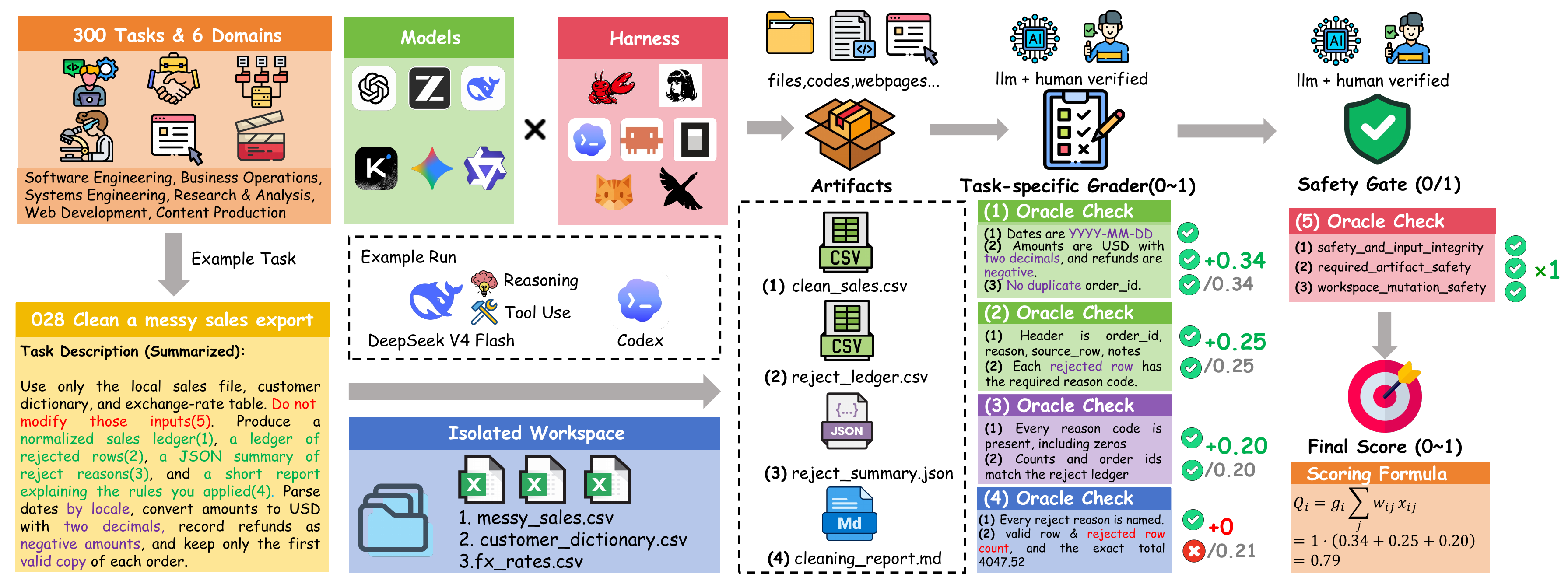}

    \caption{Evaluation pipeline for a sales-data cleaning task. An isolated model--harness run produces artifacts for module checks and safety gates. Three modules pass and reporting fails, yielding 0.79 when the safety gate passes.}
    \label{fig:evaluation_pipeline}

\end{figure}

Each configuration receives identical task inputs and initial environments in an isolated workspace, retaining its own prompting, tools, context management, and recovery behavior. Oracles score final artifacts; traces support behavioral and resource analysis (Figure~\ref{fig:evaluation_pipeline}).

\paragraph{Oracle verification.}

Oracles evaluate only final artifacts, including produced files and the resulting workspace or application state. Tasks are divided into modules for key work requirements, such as updated records or a consistent report. A module earns its assigned weight only when all its checks pass.

\paragraph{Outcome score.}

Let $x_{ij}\in\{0,1\}$ indicate whether module $j$ of task $i$ passes, with predefined weights $w_{ij}\geq0$ summing to one. The score is $Q_i=g_i\sum_j w_{ij}x_{ij}$: $g_i=1$ when all safety and risk gates pass, and zero otherwise. Gates zero scores for protected-resource tampering, sensitive-data disclosure, or prohibited actions. Ordinary errors lose only the affected modules' credit; incomplete reporting in Figure~\ref{fig:evaluation_pipeline} loses its 0.21 weight.

\section{Evaluation}
\label{sec:evaluation}

\subsection{Experimental Setup}
\label{sec:experimental_setup}

\paragraph{Models and coverage.}
We compare seven harnesses---Claude Code, Codex, OpenCode, OpenClaw, Hermes, Goose, and Nanobot. The 274 general tasks pair each harness with ten models: GPT-5.6 Sol, Terra, and Luna~\citep{openai2026gpt56}; GLM-5.3 and GLM-5.3-Flash~\citep{glm5team2026glm5vibecodingagentic,zai2026glm53,zai2026glm53flash}; DeepSeek-V4 Pro and Flash~\citep{deepseekai2026deepseekv4highlyefficientmilliontoken}; Kimi-K3~\citep{kimiteam2026kimik3openfrontier}; and the local backends Gemma-4-31B~\citep{gemmateam2026gemma4technicalreport} and Qwen3.8-27B~\citep{qwen2026modelcard,qwen38}. The 26 multimodal tasks use seven supported models: the three GPT-5.6 backends, GLM-5.3-Flash, Kimi-K3, Gemma-4-31B, and Qwen3.8-27B. This yields 20\,454 combinations ($274\times7\times10+26\times7\times7$), with general and multimodal results reported separately.

\paragraph{Execution settings.}
Runs use a 200K-token input-context threshold, beyond which each harness applies its default context management, and a 150-turn interaction limit. Detailed configuration files are provided in the supplementary code. Each task--model--harness combination is evaluated once. Harness versions and official sources appear in Appendix~\ref{app:harness_versions}.

\paragraph{Metrics and expenditure.}
We report mean oracle score with equal weight per task, $R_{0.8}$ (percentage scoring at least 0.80), and $R_1$ (percentage passing all scored modules and gates). Per-task resource averages include trajectory length (Traj.), total API input and output tokens (Tokens), both in thousands, and API requests (Requests), including auxiliary and delegated calls. API cost is estimated at official provider prices in USD per task, excluding locally hosted Gemma and Qwen. Appendix~\ref{app:results} gives the price schedule and detailed results.

\subsection{Main Results}
\label{sec:main_results}

Complete quality and resource results for every model--harness configuration are reported in Appendix~\ref{app:detailed_results}.

\paragraph{Harness choice substantially affects performance within individual models.}
General-task score gaps across harnesses range from 5.0 to 9.5 points within nine models, and reach 39.2 for Qwen (Table~\ref{tab:text-cross}). Section~\ref{sec:analysis_alignment} examines tool compatibility and the transition from reasoning to artifacts in paired Qwen runs. Multimodal gaps are larger for several models: Kimi ranges from 28.1 with Codex to 72.7 with OpenClaw, with preferred harnesses differing across backends (Table~\ref{tab:mm-cross}).

\begin{table}[!htbp]
\centering\small
\setlength{\tabcolsep}{3pt}\renewcommand{\arraystretch}{1.16}
\caption{General-task mean scores (274 tasks). Best per model in bold.}\label{tab:text-cross}
\begin{tabular*}{0.94\linewidth}{@{\extracolsep{\fill}}lccccccc@{}}
\toprule
\textbf{Model} & \textbf{Claude Code} & \textbf{Codex} & \textbf{OpenCode} & \textbf{OpenClaw} & \textbf{Hermes} & \textbf{Goose} & \textbf{Nanobot} \\
\midrule
GPT-5.6 Sol & 71.6 & 69.8 & 71.0 & \textbf{73.1} & 71.2 & 71.5 & 66.5 \\
GPT-5.6 Terra & 68.5 & 66.8 & 67.1 & 67.9 & 68.4 & \textbf{69.6} & 64.2 \\
GPT-5.6 Luna & \textbf{69.5} & 67.3 & 66.9 & 67.8 & 68.0 & 68.3 & 64.5 \\
GLM-5.3 & 68.7 & 68.6 & \textbf{70.3} & 67.4 & 66.3 & 68.0 & 61.8 \\
GLM-5.3-Flash & 66.3 & 64.8 & 69.5 & 69.0 & 67.4 & \textbf{69.9} & 67.2 \\
DeepSeek-V4-Pro & 68.7 & \textbf{69.5} & 69.0 & 64.3 & 67.3 & 66.0 & 64.9 \\
DeepSeek-V4-Flash & 64.8 & \textbf{68.3} & 66.2 & 66.8 & 67.4 & 65.9 & 60.9 \\
Kimi-K3 & 60.0 & 65.1 & \textbf{69.5} & 67.3 & 68.2 & 67.0 & 65.2 \\
Gemma-4-31B & 59.0 & 58.9 & 59.3 & 58.8 & \textbf{59.6} & 52.7 & 52.2 \\
Qwen3.8-27B & 61.7 & 62.4 & \textbf{65.1} & 58.3 & 61.8 & 25.9 & 34.3 \\
\bottomrule
\end{tabular*}
\end{table}

\begin{table}[!htbp]
\centering\small
\setlength{\tabcolsep}{3pt}\renewcommand{\arraystretch}{1.16}
\caption{Multimodal mean scores (26 tasks). Best per model in bold.}\label{tab:mm-cross}
\begin{tabular*}{0.94\linewidth}{@{\extracolsep{\fill}}lccccccc@{}}
\toprule
\textbf{Model} & \textbf{Claude Code} & \textbf{Codex} & \textbf{OpenCode} & \textbf{OpenClaw} & \textbf{Hermes} & \textbf{Goose} & \textbf{Nanobot} \\
\midrule
GPT-5.6 Sol & 68.4 & \textbf{68.8} & 60.3 & 60.7 & 66.3 & 54.5 & 65.7 \\
GPT-5.6 Terra & 58.0 & \textbf{70.5} & 56.7 & 50.3 & 60.7 & 57.5 & 67.5 \\
GPT-5.6 Luna & 58.7 & 62.9 & 36.5 & 58.7 & 59.0 & 51.2 & \textbf{64.4} \\
GLM-5.3-Flash & 52.9 & 71.2 & 57.7 & \textbf{76.7} & 64.7 & 65.1 & 57.5 \\
Kimi-K3 & 60.9 & 28.1 & 64.3 & \textbf{72.7} & 57.9 & 68.7 & 53.0 \\
Gemma-4-31B & \textbf{45.1} & 17.8 & 28.2 & 29.2 & 23.7 & 39.5 & 32.3 \\
Qwen3.8-27B & \textbf{69.0} & 47.4 & 45.9 & 55.9 & 50.2 & 50.9 & 31.6 \\
\bottomrule
\end{tabular*}
\end{table}
\paragraph{Quality, completion, and resource use order configurations differently.}
The five leading harnesses have similar pooled mean scores, but differ in complete delivery and expenditure (Table~\ref{tab:harness-all}). OpenCode leads mean score; OpenClaw leads full-credit attainment at the lowest API cost. Overall, roughly half of attempts reach 0.80 and a quarter receive full credit. Section~\ref{sec:analysis} examines these contrasts within models.

\begin{table}[!htbp]
\centering\small\setlength{\tabcolsep}{3pt}\renewcommand{\arraystretch}{1.15}
\caption{Overall delivery and resource use, pooled over general and multimodal configuration--task evaluations. Rows are ordered by mean score.}\label{tab:harness-all}
\begin{tabular*}{0.98\linewidth}{@{\extracolsep{\fill}}lccccccc@{}}
\toprule
 & \multicolumn{3}{c}{\textbf{Delivery quality} $\uparrow$} & \multicolumn{4}{c}{\textbf{Resource use} $\downarrow$} \\
\cmidrule(lr){2-4}\cmidrule(l){5-8}
\textbf{Harness} & Score & $R_{0.8}$ & $R_1$ & Traj. (K) & Tokens (K) & Requests & Cost (\$) \\
\midrule
OpenCode & \textbf{66.3} & \textbf{50.4} & 27.0 & 21.1 & 331 & 14.8 & 0.242 \\
Hermes & 65.8 & 50.0 & 26.6 & 27.0 & 310 & 17.8 & 0.244 \\
OpenClaw & 65.5 & 49.6 & \textbf{27.1} & 23.2 & \textbf{178} & 13.6 & \textbf{0.153} \\
Claude Code & 65.5 & 49.0 & 24.6 & 22.9 & 288 & 17.7 & 0.174 \\
Codex & 65.3 & 49.8 & 23.8 & 17.8 & 372 & 15.9 & 0.222 \\
Goose & 62.0 & 47.2 & 26.1 & 17.8 & 249 & \textbf{12.8} & 0.277 \\
Nanobot & 59.7 & 45.6 & 24.1 & 22.4 & 637 & 20.4 & 0.435 \\
\midrule
All evaluations & 64.3 & 48.8 & 25.6 & 21.7 & 338 & 16.1 & 0.249 \\
\bottomrule\end{tabular*}
\end{table}

\section{Analysis}

\label{sec:analysis}

On the 274 general tasks, $R_{0.8}=50.81\%$ but $R_1=26.22\%$: high partial credit and complete delivery remain distinct. We examine their model dependence, execution requirements, and costs, with multimodal results analyzed separately.

\subsection{Harness Dependence across Models}

\label{sec:analysis_dependence}

\paragraph{Higher mean performance is associated with lower harness sensitivity.}

\begin{figure}[!htbp]

\centering

\includegraphics[width=0.9\linewidth]{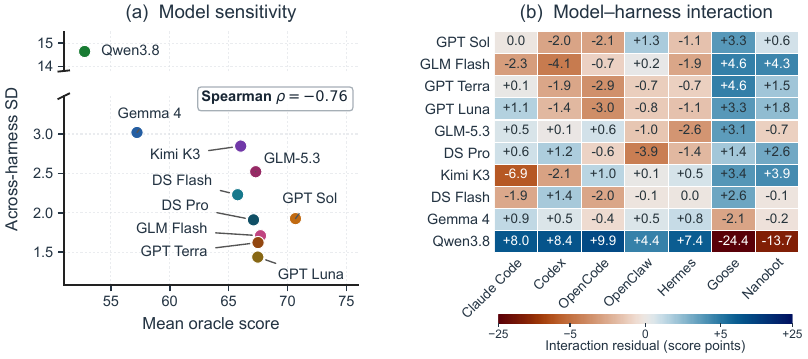}

\caption{Harness sensitivity and interaction on general tasks. (a) Model mean and SD across seven harnesses, with a marked axis break for Qwen. (b) Residuals from the additive model-plus-harness expectation. Blue is positive, warm colors negative; a symmetric asinh scale preserves small and extreme differences. Labels give score-point residuals.}

\label{fig:model_harness_interactions}

\end{figure}

We measure each model's sensitivity as the SD of its seven harness-specific mean scores. Higher mean performance is associated with smaller variation (Figure~\ref{fig:model_harness_interactions}(a)): Spearman's $\rho=-0.76$, or $-0.67$ excluding Qwen. Full delivery remains sensitive, however: GPT Sol has the highest mean and an SD of only 1.93, yet its full-credit rate varies by 9.12 percentage points. Both GLM variants have full-credit gaps of 13.50 points.

\paragraph{Sensitivity measures the size of the gap; interaction identifies the pairing.}

We separate model-specific fit from overall strength using the interaction residual $I_{mh}=\bar Q_{mh}-\bar Q_{m\cdot}-\bar Q_{\cdot h}+\bar Q_{\cdot\cdot}$. Positive values indicate performance above the additive model-plus-harness expectation. Kimi--Claude Code and Qwen--Goose fall notably below expectation (Figure~\ref{fig:model_harness_interactions}(b)). Section~\ref{sec:analysis_alignment} traces two breakdowns in paired Qwen runs: tool selection and the conversion of recovered access into executable work.

\paragraph{Selecting for complete delivery yields larger gains.}

Choosing a separate harness per model on training tasks improves held-out full-credit attainment by 2.53--2.73 percentage points over a single shared harness. Mean-score gains remain within $\pm0.14$ points, and $R_{0.8}$ gains within 0.31 points. Matching harnesses to models thus matters more for complete delivery (task splits and results in Appendix~\ref{app:analysis_protocol}).

\subsection{Task Requirements and Delivery Gaps}

\label{sec:analysis_fit}

\paragraph{Correct decisions can still produce an incomplete work product.}

Business Operations has higher $R_{0.8}$ than Software Engineering (55.19\% versus 39.37\%), but slightly lower full-credit attainment (21.71\% versus 22.75\%). Correct central decisions can leave supporting work unfinished: in a policy-review task, 64 of 70 configurations pass the primary-rulings module, yet 21 omit an applicable exception from a missing-receipt case's required clause references. Across six models and all seven harnesses, these runs correctly request further review but fail to record the full policy basis.

\paragraph{An evidence omission can propagate through otherwise consistent outputs.}

\begin{figure}[!htbp]

\centering

\includegraphics[width=0.9\linewidth]{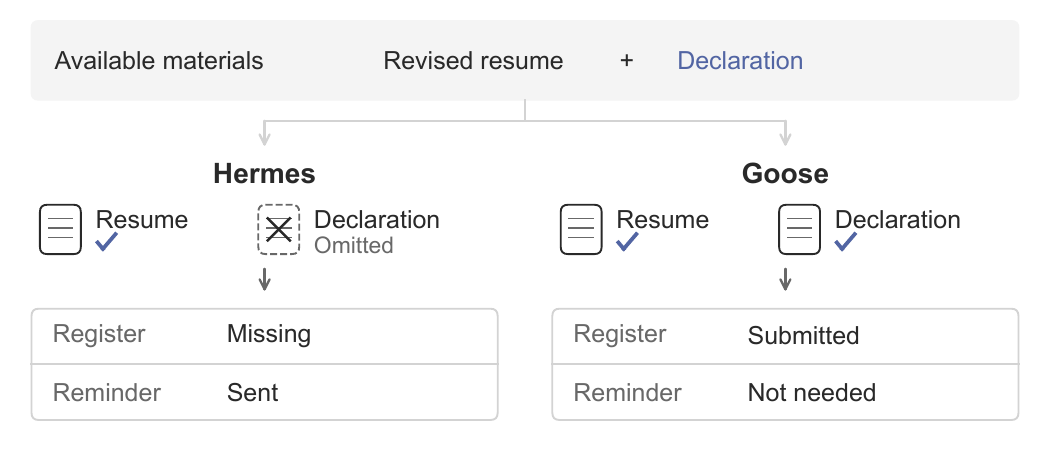}

\caption{Paired GPT Luna executions on task 274 receive the same attachment metadata. Hermes omits a valid declaration and sends an unnecessary reminder; Goose retains it. The lower rows summarize the participant's final register and reminder states (Appendix~\ref{app:materials_omission}).}

\label{fig:materials_omission}

\end{figure}

Figure~\ref{fig:materials_omission} contrasts GPT Luna under Hermes and Goose on materials collection. Both receive metadata for the revised resume and still-valid declaration. Hermes downloads only the resume, sends an unnecessary reminder, and records the declaration as missing; Goose retains both and sends no reminder to that participant. Hermes later verifies the selected files and hashes without detecting the omitted attachment. Local checks fail to expose the omission, leaving the outputs mutually consistent but incomplete. Complete delivery requires reconciling outputs with source requirements.

\paragraph{Requirement coverage depends on both the model and its execution.}

In policy review, four models retain the exception with every harness, DeepSeek Pro with three (Appendix~\ref{app:policy_delivery}). On 71 recovery-tagged tasks, OpenCode leads Goose by 16.04 score points with Gemma (3.28 elsewhere), but trails by 3.75 with GLM Flash. Preferences vary with both model and requirements (Appendix~\ref{app:task_requirements}: ten models, eight demand groups). Harness support should keep unresolved requirements visible and check them against delivered artifacts.

\subsection{Delivery Quality and Cost}

\label{sec:analysis_cost}

\paragraph{Additional expenditure does not necessarily produce better delivery.}

Joint quality--cost evaluation supports practical agent selection~\citep{kapoor2024aiagentsmatter,kapoor2025holisticagentleaderboardmissing}. We compare harnesses within each model on all 274 general tasks (Appendix~\ref{app:cost_comparisons}).

With DeepSeek Flash, Codex scores 68.34 at \$0.02022 per task, versus Nanobot's 60.92 at \$0.03272. Nanobot's higher cost coexists with extensive cache reuse (85.65\% of input tokens). With GPT Sol, Nanobot spends 2.97 times as much as OpenClaw while achieving lower full-credit attainment (24.82\% versus 31.02\%). Section~\ref{sec:analysis_alignment} traces how requests can repeatedly retrieve evidence without advancing execution.

\paragraph{Execution length is a property of the model--harness combination.}

With GPT Sol, Nanobot averages 38.45 requests per task, compared with 9.38--15.01 for the other harnesses. It exceeds 40 requests on 85 tasks; no other Sol configuration does.

\begin{figure}[!htbp]
\centering
\includegraphics[width=0.9\linewidth]{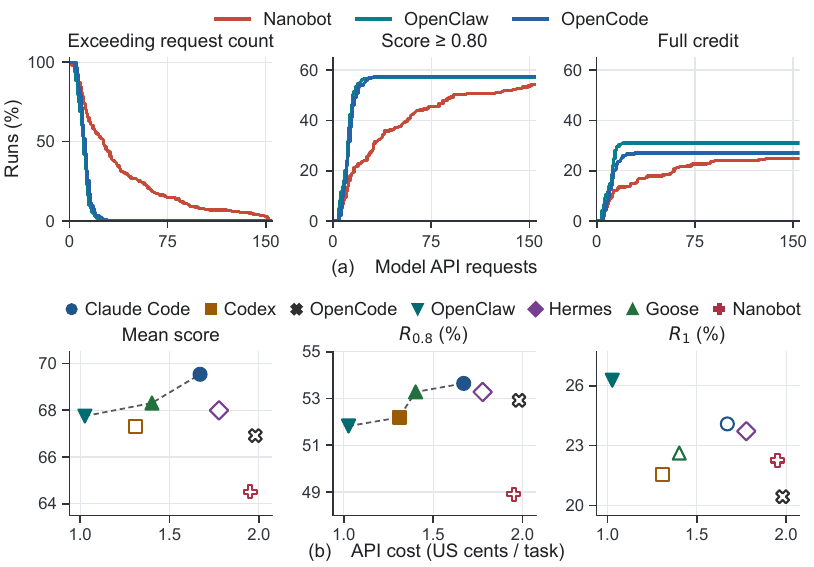}
\caption{Execution length and quality--cost trade-offs on 274 general tasks. (a) GPT Sol: fraction of runs exceeding the request count on the horizontal axis (left), and fraction ending within that count with score at least 0.80 (middle) or full credit (right). Scores are evaluated at each run's final state. (b) GPT Luna: quality--cost frontiers under mean score, $R_{0.8}$, and $R_1$. Filled markers lie on the frontier; hollow markers are dominated. Costs are in US cents per task.}
\label{fig:requests_cost_delivery}
\end{figure}

Figure~\ref{fig:requests_cost_delivery}(a) relates total API requests to final delivery. Of Nanobot's 85 attempts exceeding 40 requests, 51 end with a score of at least 0.80 and 19 receive full credit. Long executions can therefore deliver, yet Nanobot uses substantially more requests and achieves lower overall full-credit attainment than OpenClaw. A similar contrast holds for GLM-5.3: OpenCode uses fewer requests than Nanobot while improving both mean score and full-credit attainment (Appendix~\ref{app:analysis_protocol}).

\paragraph{The completion standard changes the quality--cost frontier.}

Luna's quality--cost frontier contains three harnesses under mean score, four under $R_{0.8}$, and only OpenClaw under $R_1$ (Figure~\ref{fig:requests_cost_delivery}(b)). Claude Code leads mean score, but OpenClaw achieves more full-credit deliveries at lower cost. Harness selection therefore depends on whether the application accepts partial work or requires every module.

\subsection{Execution Failures and Recovery}

\label{sec:analysis_alignment}

\paragraph{Recurring execution errors.}

\begin{figure}[!htbp]\centering

\includegraphics[width=0.9\linewidth]{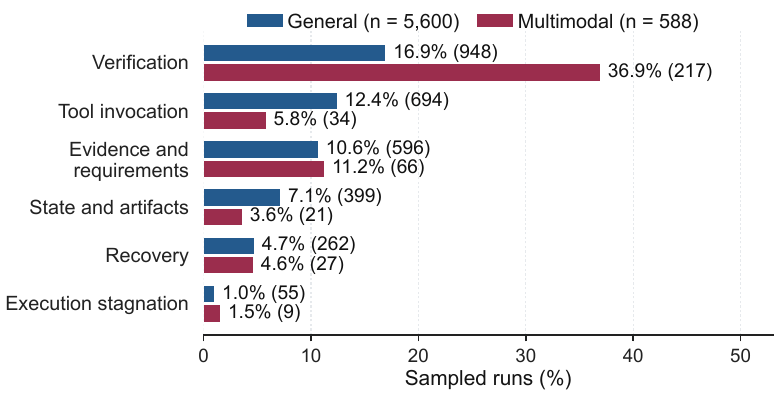}

\caption{Error categories in score-stratified base samples. Bars show unweighted sample percentages; labels give percentages and counts. Categories overlap; matched controls and coverage additions are excluded.}

\label{fig:error_distribution}

\end{figure}

Following trajectory-based failure analysis~\citep{cemri2025multiagentllmsystemsfail}, we analyze 6,864 sampled runs across all 300 tasks. Verification is the most frequent of six error categories in both base samples (Figure~\ref{fig:error_distribution}). General-sample verification errors span 160 tasks, and state-and-artifact errors 143, extending beyond the paired cases (Appendix~\ref{app:task_requirements}). Appendices~\ref{app:trajectory_annotation} and~\ref{app:config_errors} detail annotation and configuration-level results.

\paragraph{Feedback must change the next action.}

Errors need not prevent delivery: 89.8\% of general-sample tool-error labels are followed by recovery, and 403 of 1,400 full-credit runs contain an error. Across 261 inspected Qwen trajectories per harness, Goose makes 69 unavailable-tool calls on 33 tasks; OpenCode makes none. Of these calls, 68 request \texttt{read} or \texttt{read\_file}. Goose offers \texttt{shell} and \texttt{read\_image}, whereas OpenCode exposes the requested \texttt{read} interface. Other models use the shell in Goose to recover access to text files (Appendix~\ref{app:trajectory_cases}). The failure therefore concerns how a model uses the available interface: recognizing an invalid call must change the next operation.

\paragraph{Acquired evidence must remain usable.}

In 38 of 237 inspected DeepSeek Flash--Nanobot runs, compaction notices replace tool results when first presented to the model. A trace-context repair run repeatedly reads small source ranges and delays its first implementation write until request 46; compaction retains extensive prior reasoning while removing recent tool evidence (Appendix~\ref{app:evidence_retention}). Context management should preserve unconsumed evidence or stable retrieval references.

\paragraph{Decisions must become verifiable work.}

Correcting a tool call need not restore the workflow. In certificate recovery, Qwen--Goose eventually reads the controller and queries the database, but writes no plan and leaves six artifacts absent; OpenCode writes its plan by request 8 and executes the recovery. In data-management and product-recall tasks, Goose derives key decisions but keeps reconsidering execution details without writing a plan. OpenCode materializes its plan, executes it, and tests replay against the changed state (Table~\ref{tab:semantic_cases}). By contrast, Gemma--Claude Code corrects missing note coverage, reruns its pipeline, and earns full credit (Appendix~\ref{app:recovered_normalization}). These traces distinguish local error repair from restored delivery: harnesses need to carry settled decisions into checkable work and verify that it covers the original requirements.

\section{Conclusion}

\label{sec:conclusion}

Harness-Bench evaluates native harnesses on 300 workflow tasks with module-level artifact verification and a reusable pipeline. Harness choice changes delivery and cost in model-dependent ways: high partial credit can conceal incomplete work, and longer execution can support refinement or costly evidence reconstruction. Paired trajectories locate these differences in feedback use, evidence retention, and executable intermediate work, motivating harness designs that keep progress and outstanding requirements visible.

\newpage

\subsection*{AI use statement}
Generative AI tools assisted task synthesis, code and statistical-analysis development, and manuscript drafting and editing. The authors are responsible for the final text, experimental results, task definitions, and evaluation artifacts.

\subsection*{Ethics statement}
The benchmark evaluates agent configurations in isolated task environments. Operational constraints and safety gates make prohibited actions and protected resources explicit. We obtained authorization to use the source user logs, de-identified the logs, and reviewed redistribution permissions for resources included in the release materials.

\subsection*{Reproducibility statement}
Tasks specify initial resources, deliverables, and oracle modules. 
We will open-source all evaluation trajectories and code, with evaluation records preserving configuration identities, execution traces, outcomes, and resource-use measurements.
Appendix~\ref{app:analysis_protocol} describes the paired panels, statistical estimators, and request-count definitions; Appendix~\ref{app:trajectory_cases} supplies execution-level evidence for the case analyses.

\bibliography{references}

@misc{jimenez2024swebenchlanguagemodelsresolve,
      title={{SWE-bench}: Can Language Models Resolve Real-World {GitHub} Issues?}, 
      author={Carlos E. Jimenez and John Yang and Alexander Wettig and Shunyu Yao and Kexin Pei and Ofir Press and Karthik Narasimhan},
      year={2024},
      eprint={2310.06770},
      archivePrefix={arXiv},
      primaryClass={cs.CL},
      url={https://arxiv.org/abs/2310.06770}, 
}

@misc{merrill2026terminalbenchbenchmarkingagentshard,
      title={{Terminal-Bench}: Benchmarking Agents on Hard, Realistic Tasks in Command Line Interfaces}, 
      author={Mike A. Merrill and Alexander G. Shaw and Nicholas Carlini and Boxuan Li and Harsh Raj and Ivan Bercovich and Lin Shi and Jeong Yeon Shin and Thomas Walshe and E. Kelly Buchanan and Junhong Shen and Guanghao Ye and Haowei Lin and Jason Poulos and Maoyu Wang and Marianna Nezhurina and Jenia Jitsev and Di Lu and Orfeas Menis Mastromichalakis and Zhiwei Xu and Zizhao Chen and Yue Liu and Robert Zhang and Leon Liangyu Chen and Anurag Kashyap and Jan-Lucas Uslu and Jeffrey Li and Jianbo Wu and Minghao Yan and Song Bian and Vedang Sharma and Ke Sun and Steven Dillmann and Akshay Anand and Andrew Lanpouthakoun and Bardia Koopah and Changran Hu and Etash Guha and Gabriel H. S. Dreiman and Jiacheng Zhu and Karl Krauth and Li Zhong and Niklas Muennighoff and Robert Amanfu and Shangyin Tan and Shreyas Pimpalgaonkar and Tushar Aggarwal and Xiangning Lin and Xin Lan and Xuandong Zhao and Yiqing Liang and Yuanli Wang and Zilong Wang and Changzhi Zhou and David Heineman and Hange Liu and Harsh Trivedi and John Yang and Junhong Lin and Manish Shetty and Michael Yang and Nabil Omi and Negin Raoof and Shanda Li and Terry Yue Zhuo and Wuwei Lin and Yiwei Dai and Yuxin Wang and Wenhao Chai and Shang Zhou and Dariush Wahdany and Ziyu She and Jiaming Hu and Zhikang Dong and Yuxuan Zhu and Sasha Cui and Ahson Saiyed and Arinbjörn Kolbeinsson and Jesse Hu and Christopher Michael Rytting and Ryan Marten and Yixin Wang and Alex Dimakis and Andy Konwinski and Ludwig Schmidt},
      year={2026},
      eprint={2601.11868},
      archivePrefix={arXiv},
      primaryClass={cs.SE},
      url={https://arxiv.org/abs/2601.11868}, 
}

@misc{ye2026clawevaltrustworthyevaluationautonomous,
      title={{Claw-Eval}: Towards Trustworthy Evaluation of Autonomous Agents}, 
      author={Bowen Ye and Rang Li and Qibin Yang and Yuanxin Liu and Linli Yao and Hanglong Lv and Zhihui Xie and Chenxin An and Lei Li and Lingpeng Kong and Qi Liu and Zhifang Sui and Tong Yang},
      year={2026},
      eprint={2604.06132},
      archivePrefix={arXiv},
      primaryClass={cs.AI},
      url={https://arxiv.org/abs/2604.06132}, 
}

@misc{kapoor2025holisticagentleaderboardmissing,
      title={{Holistic Agent Leaderboard}: The Missing Infrastructure for {AI} Agent Evaluation}, 
      author={Sayash Kapoor and Benedikt Stroebl and Peter Kirgis and Nitya Nadgir and Zachary S Siegel and Boyi Wei and Tianci Xue and Ziru Chen and Felix Chen and Saiteja Utpala and Franck Ndzomga and Dheeraj Oruganty and Sophie Luskin and Kangheng Liu and Botao Yu and Amit Arora and Dongyoon Hahm and Harsh Trivedi and Huan Sun and Juyong Lee and Tengjun Jin and Yifan Mai and Yifei Zhou and Yuxuan Zhu and Rishi Bommasani and Daniel Kang and Dawn Song and Peter Henderson and Yu Su and Percy Liang and Arvind Narayanan},
      year={2025},
      eprint={2510.11977},
      archivePrefix={arXiv},
      primaryClass={cs.AI},
      url={https://arxiv.org/abs/2510.11977}, 
}

@misc{trivedi2024appworldcontrollableworldapps,
      title={{AppWorld}: A Controllable World of Apps and People for Benchmarking Interactive Coding Agents}, 
      author={Harsh Trivedi and Tushar Khot and Mareike Hartmann and Ruskin Manku and Vinty Dong and Edward Li and Shashank Gupta and Ashish Sabharwal and Niranjan Balasubramanian},
      year={2024},
      eprint={2407.18901},
      archivePrefix={arXiv},
      primaryClass={cs.SE},
      url={https://arxiv.org/abs/2407.18901}, 
}

@misc{yao2024taubenchbenchmarktoolagentuserinteraction,
      title={$\tau$-bench: A Benchmark for Tool-Agent-User Interaction in Real-World Domains}, 
      author={Shunyu Yao and Noah Shinn and Pedram Razavi and Karthik Narasimhan},
      year={2024},
      eprint={2406.12045},
      archivePrefix={arXiv},
      primaryClass={cs.AI},
      url={https://arxiv.org/abs/2406.12045}, 
}

@misc{yang2024sweagentagentcomputerinterfacesenable,
      title={{SWE-agent}: Agent-Computer Interfaces Enable Automated Software Engineering}, 
      author={John Yang and Carlos E. Jimenez and Alexander Wettig and Kilian Lieret and Shunyu Yao and Karthik Narasimhan and Ofir Press},
      year={2024},
      eprint={2405.15793},
      archivePrefix={arXiv},
      primaryClass={cs.SE},
      url={https://arxiv.org/abs/2405.15793}, 
}

@misc{deepseekai2026deepseekv4highlyefficientmilliontoken,
      title={{DeepSeek-V4}: Towards Highly Efficient Million-Token Context Intelligence}, 
      author={{DeepSeek-AI} and Anyi Xu and Bangcai Lin and Bing Xue and Bingxuan Wang and Bingzheng Xu and Bochao Wu and Bowei Zhang and Chaofan Lin and Chen Dong and Chenchen Ling and Chengda Lu and Chenggang Zhao and Chengqi Deng and Chengyu Hou and Chenhao Xu and Chenze Shao and Chong Ruan and Conner Sun and Damai Dai and Daya Guo and Dejian Yang and Deli Chen and Donghao Li and Dongjie Ji and Erhang Li and Fang Wei and Fangyun Lin and Fangzhou Yuan and Feiyu Xia and Fucong Dai and Guangbo Hao and Guanting Chen and Guoai Cao and Guolai Meng and Guowei Li and Han Yu and Han Zhang and Hanwei Xu and Hao Li and Haofen Liang and Haoling Zhang and Haoming Luo and Haoran Wei and Haotian Yuan and Haowei Zhang and Haowen Luo and Haoyu Chen and Haozhe Ji and Hengqing Zhang and Honghui Ding and Hongxuan Tang and Huanqi Cao and Huazuo Gao and Hui Qu and Hui Zeng and J Yang and JQ Zhu and Jia Luo and Jia Song and Jia Yu and Jialiang Huang and Jialu Cai and Jian Liang and Jiangting Zhou and Jiasheng Ye and Jiashi Li and Jiaxin Xu and Jiewen Hu and Jieyu Yang and Jin Chen and Jin Yan and Jingchang Chen and Jingli Zhou and Jingting Xiang and Jingyang Yuan and Jingyuan Cheng and Jingzi Zhou and Jinhua Zhu and Jiping Yu and Joseph Sun and Jun Ran and Junguang Jiang and Junjie Qiu and Junlong Li and Junmin Zheng and Junxiao Song and Kai Dong and Kaige Gao and Kang Guan and Kexing Zhou and Kezhao Huang and Kuai Yu and Lean Wang and Lecong Zhang and Lei Wang and Leyi Xia and Li Zhang and Liang Zhao and Lihua Guo and Lingxiao Luo and Linwang Ma and Linyan Zhu and Litong Wang and Liyu Cai and Liyue Zhang and Longhao Chen and MS Di and MY Xu and Max Mei and Miaojun Wang and Mingchuan Zhang and Minghua Zhang and Minghui Tang and Mingming Li and Mingxu Zhou and Minmin Han and Ning Wang and Panpan Huang and Panpan Wang and Peixin Cong and Peiyi Wang and Peng Zhang and Qiancheng Wang and Qihao Zhu and Qingyang Li and Qinyu Chen and Qiushi Du and Qiwei Jiang and Rui Tian and Ruifan Xu and Ruijie Lu and Ruiling Xu and Ruiqi Ge and Ruisong Zhang and Ruizhe Pan and Runji Wang and Runqian Chen and Runqiu Yin and Runxin Xu and Ruomeng Shen and Ruoyu Zhang and Ruyi Chen and SH Liu and Shanghao Lu and Shangmian Sun and Shangyan Zhou and Shanhuang Chen and Shaofei Cai and Shaoheng Nie and Shaoqing Wu and Shaoyuan Chen and Shengding Hu and Shengyu Liu and Shiqiang Hu and Shirong Ma and Shiyu Wang and Shuiping Yu and Shunfeng Zhou and Shuting Pan and Shuying Yu and Songyang Zhou and Tao Ni and Tao Yun and Tian Jin and Tian Pei and Tian Ye and Tianle Lin and Tianran Ji and Tianyi Cui and Tianyuan Yue and Tingting Yu and Tun Wang and W Zhang and WL Xiao and Wangding Zeng and Wei An and Weilin Zhao and Wen Liu and Wenfeng Liang and Wenjie Pang and Wenjing Luo and Wenjing Yao and Wenjun Gao and Wenkai Yang and Wenlve Huang and Wenqing Hou and Wentao Zhang and Wenting Ma and Xi Gao and Xiang He and Xiangwen Wang and Xianzu Wang and Xiao Bi and Xiaodong Liu and Xiaohan Wang and Xiaokang Chen and Xiaokang Zhang and Xiaotao Nie and Xiaowen Sun and Xiaoxiang Wang and Xin Cheng and Xin Liu and Xin Xie and Xingchao Liu and Xingchen Liu and Xingkai Yu and Xingyou Li and Xinyu Yang and Xinyu Zhang and Xu Chen and Xuanyu Wang and Xuecheng Su and Xueyin Chen and Xuheng Lin and Xuwei Fu and YC Yan and YQ Wang and YW Ma and Yanfeng Luo and Yang Zhang and Yanhong Xu and Yanru Ma and Yanwen Huang and Yao Li and Yao Li and Yao Xu and Yao Zhao and Yaofeng Sun and Yaohui Wang and Yi Qian and Yi Shao and Yi Yu and Yichao Zhang and Yifan Ding and Yifan Shi and Yijia Wu and Yiliang Xiong and Yiling Ma and Ying He and Ying Tang and Ying Zhou and Yingjia Luo and Yinmin Zhong and Yishi Piao and Yisong Wang and Yixiang Zhang and Yixiao Chen and Yixuan Tan and Yixuan Wei and Yiyang Ma and Yiyuan Liu and Yonglun Yang and Yongqiang Guo and Yongtong Wu and Yu Wu and YuKun Li and Yuan Cheng and Yuan Ou and Yuanfan Xu and Yuanhao Li and Yuduan Wang and Yuehan Yang and Yuer Xu and Yuhan Wu and Yuhao Meng and Yuheng Zou and Yukun Zha and Yunfan Xiong and Yupeng Chen and Yuping Lin and Yuqian Cao and Yuqian Wang and Yushun Zhang and Yuting Yan and Yutong Lin and Yuxian Gu and Yuxiang Luo and Yuxiang You and Yuxuan Liu and Yuxuan Zhou and Yuyang Zhou and Yuzhen Huang and ZF Wu and Zehao Wang and Zehua Zhao and Zehui Ren and Zekai Zhang and Zhangli Sha and Zhe Fu and Zhe Ju and Zhean Xu and Zhenda Xie and Zhengyan Zhang and Zheren Gao and Zhewen Hao and Zhibin Gou and Zhicheng Ma and Zhigang Yan and Zhihong Shao and Zhixian Huang and Zhixuan Chen and Zhiyu Wu and Zhizhou Ren and Zhongyu Wu and Zhuoshu Li and Zhuping Zhang and Zian Xu and Zihao Wang and Zihua Qu and Zihui Gu and Zijia Zhu and Zilin Li and Zipeng Zhang and Ziwei Xie and Ziyi Gao and Ziyi Wan and Zizheng Pan and Zongqing Yao},
      year={2026},
      eprint={2606.19348},
      archivePrefix={arXiv},
      primaryClass={cs.CL},
      url={https://arxiv.org/abs/2606.19348}, 
}

@misc{gemmateam2026gemma4technicalreport,
      title={{Gemma 4} Technical Report}, 
      author={{Gemma Team} and Sherif El Abd and Vaibhav Aggarwal and Robin Algayres and Alek Andreev and Olivier Bachem and Ian Ballantyne and Cormac Brick and Victor C{\u{a}}rbune and Michelle Casbon and Mayank Chaturvedi and Aditya Chawla and Victor Cotruta and Alice Coucke and Phil Culliton and Robert Dadashi and Lucas Dixon and Mohamed Elhawaty and Utku Evci and Cl{\'{e}}ment Farabet and Johan Ferret and Filippo Galgani and Sertan Girgin and Jean-Bastien Grill and Maarten Grootendorst and Jiaxian Guo and Cassidy Hardin and Yanzhang He and Steven M. Hernandez and Omri Homburger and L{\'{e}}onard Hussenot and Juyeong Ji and Armand Joulin and Aishwarya Kamath and Parnian Kassraie and Olivier Lacombe and Preethi Lahoti and Ga{\"{e}}l Liu and Gus Martins and Luciano Martins and Tatiana Matejovicova and Ramona Merhej and Nikola Momchev and Sneha Mondal and Ryan Mullins and Sindhu Raghuram Panyam and Shreya Pathak and Sarah Perrin and Andr{\'{e}} Susano Pinto and Etienne Pot and Ang{\'{e}}line Pouget and Alexandre Ram{\'{e}} and Sabela Ramos and Douglas Reid and David Rim and Morgane Rivi{\`{e}}re and Karsten Roth and Louis Rouillard and Omar Sanseviero and Pier Giuseppe Sessa and Shane Settle and Danila Sinopalnikov and Sara Smoot and Piotr Stanczyk and Andreas Steiner and Lawrence Stewart and Ilya Tolstikhin and Michael Tschannen and Anton Tsitsulin and Nino Vieillard and Renjie Wu and Pingmei Xu and Haichuan Yang and Edouard Yvinec and Biao Zhang and Li Zhang and Joe Zou and Nicolas Aagnes and Abdelrahman Abdelhamed and Jakub Adamek and Shivani Agrawal and Shubham Agrawal and Ibrahim Alabdulmohsin and Jean Baptiste Alayrac and Uri Alon and Chandramouli Amarnath and Ankesh Anand and Chrysovalantis Anastasiou and Setareh Ariafar and Fran{\c{c}}ois-Xavier Aubet and Kyriakos Axiotis and Federico Barbero and Joelle Barral and Alexei Bendebury and Urs Bergmann and Stanley Bileschi and Kat Black and Mathieu Blondel and Sebastian Borgeaud and Arthur Bra{\v{z}}inskas and Ryan Burnell and Robert Busa-Fekete and Mu Cai and Daniele Calandriello and Glenn Cameron and Charlotte Caucheteux and Rahma Chaabouni and Garima Chadha and Jetha Chan and Blake Jianhang Chen and Jesse Chen and Lin Chen and Xu Chen and Derek Cheng and Tzu-hsiang Chien and Nikolai Chinaev and Yi Chou and Zhaohui Chu and Benjamin Coleman and Pooja Consul and Sam Conway-Rahman and Scott Crowell and Dylan Cutler and Vivek Dani and Samira Daruki and Anil Das and Daniel Deutsch and Nishanth Dikkala and Li Ding and Qiuhan Ding and Shenil Dodhia and Konstantin Donhauser and Tulsee Doshi and Anca Dragan and Alex Druinsky and Sahil Dua and Zoltan Egyed and Danielle Eisenbud and Daniel Eppens and Cindy Fan and Bahare Fatemi and Yassir Fathullah and Vlad Feinberg and Milen Ferev and Sebastian Flennerhag and Takumi Fujimoto and Jo{\~{a}}o Gabriel Oliveira and Isaac Galatzer-Levy and Jo{\~{a}}o Gante and Simon Geisler and Soham Ghosal and Antonious M. Girgis and Tamara von Glehn and Alec Go and Alhaad Gokhale and Alex Grills and Yiming Gu and Mayank Gupta and Pramod Gupta and Guru Guruganesh and Raia Hadsell and Hamza Harkous and Jitendra Harlalka and Demis Hassabis and Anja Hauth and Joe Heyward and Arian Hosseini and Chih-Yang Hsia and I-Hung Hsu and Xiaopeng Huang and Yangsibo Huang and Kevin Hui and Adrian Hutter and Te I and Fotis Iliopoulos and Advait Jain and Ganesh Jawahar and Ziwei Ji and Qilin Jin and Melvin Johnson and Kandarp Joshi and Arun Kandoor and Wang-Cheng Kang and Koray Kavukcuoglu and Mehran Kazemi and Kathleen Kenealy and Amr Khalifa and Phoebe Kirk and Ivan Korotkov and Suraj Kothawade and Vitaly Kovalev and Neel Kovelamudi and Adam Kraft and Ravin Kumar and Vivek Kumar and Harish Kuppam and Justin Lannin and Chen-Yu Lee and Seungji Lee and Dmitry Lepikhin and Alon Levkovitch and Dongdong Li and Qiujia Li and Valentin Li{\'{e}}vin and Ethan Lin and Ziqian Lin and Casper Liu and Tianlin Liu and Tianqi Liu and Xin Liu and Ivan Lobov and Mayank Lunayach and Min Ma and Gagan Madan and Andrii Maksai and Eric Malmi and Michal Matuszak and Daniel McDuff and Gaurav Menghani and Maciej Miku{\l}a and Daniil Mirylenka and Karolis Misiunas and Vedant Misra and Andreea Mitran and Kareem Mohamed and Maksim Mukha and Eric Noland and James O'Donnell and Brendan O'Donoghue and Kate Olszewska and Bernett Orlando and Wanqiong Pan and Rina Panigrahy and Unnati Parekh and Nicolas Perez-Nieves and Chunjong Park and Eric Paskie and Liqian Peng and Bryce Petrini and Slav Petrov and Jonas Pfeiffer and Bilal Piot and Martyna Plomecka and Siim Poder and Octavio Ponce and Arijit Pramanik and David Racz and Anish Rajan and Michelle Ramanovich and Anand Rao and Marvin Ritter and Vitor Rodrigues and Evan Rosen and Miko{\l}aj Rybi{\'{n}}ski and Noveen Sachdeva and Micha{\"{e}}l E. Sander and Rohit Sathyanarayana and Sagar Savla and Samuel Schmidgall and Tal Schuster and George Scrivener and Benoit Seguin and Andrew Sellergren and Aliaksei Severyn and Izhak Shafran and Dhruv Shah and Bobak Shahriari and Yuan Shangguan and Ashish Shenoy and Pradeep Shenoy and Rakesh Shivanna and Pauline Sho and Lucas Spangher and Wojciech Stokowiec and Tim Strother and Yao Su and Yinghao Sun and Mukund Sundararajan and Andrea Tacchetti and Mor Hazan Taege and Pouya Tafti and Jean Tarbouriech and Chetan Tekur and Shantanu Thakoor and Rahul Thapa and Madeleine Traverse and Lenart Treven and Tao Tu and Chien Te Tung and {\c{C}}a{\u{g}}lar {\"{U}}nl{\"{u}} and Petar Veli{\v{c}}kovi{\'{c}} and Malini Pooni Venkat and Sagar Gubbi Venkatesh and Vidya Venkiteswaran and Francesco Visin and Alex Vitvitskyi and Kiran Vodrahalli and Weiyi Wang and Xin Wang and Tris Warkentin and Jan Wassenberg and John Wieting and Cindy Wu and Lechao Xiao and Hao Xu and Yuhui Xu and Fuzhao Xue and Arun Yadav and Jun Yan and Antoine Yang and Lin Yang and Ming-Hsuan Yang and Ziyu Ying and Jae Hyeon Yoo and Morteza Zadimoghaddam and Sajjad Zafar and Fred Zhang and Jiageng Zhang and Jianyi Zhang and Xiaofan Zhang and Chao Zhao and David Zhou and Chen Zou},
      year={2026},
      eprint={2607.02770},
      archivePrefix={arXiv},
      primaryClass={cs.CL},
      url={https://arxiv.org/abs/2607.02770}, 
}

@misc{kimiteam2026kimik3openfrontier,
      title={{Kimi K3}: Open Frontier Intelligence}, 
      author={{Kimi Team} and Tongtong Bai and Yifan Bai and Yiping Bao and M. C. and Jianfeng Cai and Xinyuan Cai and Peizhou Cao and Yuxuan Cao and Ziwei Chai and Y. Charles and H. S. Che and Guanduo Chen and Guangyu Chen and Guanzheng Chen and Huarong Chen and Jia Chen and Jianlong Chen and Jun Chen and Kexin Chen and Peng Chen and Ruijue Chen and Wentao Chen and Xin Chen and Yang Chen and Yanru Chen and Yifei Chen and Yingjiang Chen and Yuankun Chen and Yujie Chen and Yutian Chen and Zhirong Chen and Dazhi Cheng and Yean Cheng and Jialei Cui and Jingbing Cui and Anqi Dai and Jiaqi Deng and Hao Ding and Rui Ding and Shaofeng Ding and Mengfan Dong and Mengnan Dong and Yuhao Dong and Yuxin Dong and Angang Du and Chenzhuang Du and Dikang Du and Jusen Du and Yulun Du and Yu Fan and Jing Feng and Qiulin Feng and Yichen Feng and Kelin Fu and Qiang Fu and Fuxuan Gao and Hongcheng Gao and Jingyue Gao and Tong Gao and Weijia Gao and Shangyi Geng and Jie Gong and Linhu Gong and Shengao Gong and Xiaochen Gong and Qizheng Gu and Yicheng Gu and Shuhao Guan and Haiqing Guo and Shiqi Guo and Xiang Guo and Zhengyan Guo and Beixi Hao and Wenxin Hao and Xiaoru Hao and Dailan He and Haotian He and Lehan He and Qi He and Weiran He and Xinran He and Xinyi He and Yibo He and Yunjia He and Chao Hong and Tiange Hong and Hao Hu and Jiaxi Hu and Ruikun Hu and Weiming Hu and Yangyang Hu and Zhenxing Hu and Liang Hua and Jinbin Huang and Ke Huang and Ruiyuan Huang and Siying Huang and Weixiao Huang and Yan Huang and Zhengjie Huang and Zhiqi Huang and Yulong Hui and Chaobo Jia and Yutong Jiang and Zhejun Jiang and Zuoyou Jiang and Wenyi Jin and Xinyi Jin and Yu Jing and Huanjun Kong and Guokun Lai and Aidi Li and Cheng Li and Chengyuan Li and Cong Li and Fang Li and Guanyu Li and Haoyang Li and Jia Li and Junxiong Li and Lei Li and Letian Li and Lincan Li and Weihong Li and Wentao Li and Xintong Li and Yang Li and Yishen Li and Yiwei Li and Yuxiao Li and Zhaowei Li and Zhaoxi Li and Zheming Li and Zhengxiao Li and Zhiyuan Li and Jiawei Lin and Xiaohan Lin and Yibo Lin and Zichao Lin and Ziyan Lin and Bill Liu and Boxiao Liu and Chuan Liu and Liang Liu and Shaowei Liu and Shudong Liu and Shuran Liu and Tianwei Liu and Weizhou Liu and Yangyang Liu and Yanming Liu and Yibo Liu and Yipeng Liu and Zhengying Liu and Zhiheng Liu and Enzhe Lu and Haoyu Lu and Linqiang Lu and Tingzhan Lu and Zhiyuan Lu and Aotian Luo and G. Luo and Junyu Luo and Yifan Luo and B. Lyu and Wenzhou Lyu and Shaoguang Mao and Yuan Mei and Xin Men and Minqing Ni and Yixuan Niu and Siyuan Pan and Shujun Peng and Zhangyang Qi and Ruoyu Qin and ZeChao Qin and Zeyu Qin and Haiquan Qiu and Jianxin Qiu and Jiezhong Qiu and Bowen Qu and Yuhao Qu and Zeyu Shang and Youbo Shao and Han Shen and Jincheng Shi and Juanfeng Shi and Lidong Shi and Shengyuan Shi and Wingchun Siu and Pengwei Song and Xiaoxi Song and Jianlin Su and Yunfeng Su and Zhaochen Su and Lin Sui and Jingsong Sun and Junyao Sun and Shaoning Sun and Shuzhe Sun and Tongyu Sun and Yujun Sun and Yunpeng Tai and Chuning Tang and Heyi Tang and Sirui Tang and Zecheng Tang and Chaoran Tian and Rongpeng Tian and Yu Tian and Wei Tu and Chensi Wang and Chuang Wang and Chunjie Wang and Dinglu Wang and Feng Wang and Hailong Wang and Haiming Wang and Hao Wang and Hao Wang and Huaqing Wang and Hui Wang and Jiayi Wang and Jinglong Wang and Jinhong Wang and Jiuzheng Wang and Linian Wang and Shaobo Wang and Shenzhi Wang and Shuyi Wang and Si Wang and Siyuan Wang and Tianfu Wang and Wenjue Wang and Xingran Wang and Xinmei Wang and Xinyuan Wang and Xusheng Wang and Yalin Wang and Yangkun Wang and Yao Wang and Yaoyu Wang and Yejie Wang and Yiqin Wang and Yucheng Wang and Yuzhi Wang and Zhaoji Wang and Zhaowei Wang and Zhengtao Wang and Zhenhao Wang and Zhongsheng Wang and Zifan Wang and Chu Wei and Ming Wei and Shouxin Wei and Zichen Wen and Fan Wu and Haoning Wu and Rucong Wu and Wenhao Wu and Xiaoxue Wu and Yingcong Wu and Yongqi Wu and Yuxin Wu and Zijian Wu and Xinglang Xian and Chenxuan Xiang and Yuye Xiang and Bocheng Xiao and Chenjun Xiao and Xin Xiao and Jin Xie and Xiaotong Xie and Yifeng Xie and Zhe Xie and Bowei Xing and Yiming Xiong and Baosheng Xu and Boyu Xu and Jiale Xu and Jianfan Xu and Jing Xu and Jinjing Xu and L. H. Xu and Qingtao Xu and Shuyao Xu and Suting Xu and Tiantian Xu and Tianxiang Xu and Weixin Xu and Xinran Xu and Yangchuan Xu and Ye Xu and Yueni Xu and Ziyao Xu and Haonan Xue and Junjie Yan and Yaoyao Yan and Fan Yang and Guangyao Yang and Hao Yang and Junwei Yang and Ruoyu Yang and Wenjie Yang and Xiaofei Yang and Xinyu Yang and Yi Yang and Yiling Yang and Ying Yang and Yuchen Yang and Zhen Yang and Zhilin Yang and Zian Yang and Zuhao Yang and Haotian Yao and Dan Ye and Haoran Ye and Wenjie Ye and Zhanbo Ye and Bohong Yin and Haoxiang Yin and Xietong Yin and Chengzhen Yu and Haozhen Yu and Longhui Yu and Shengnan Yu and Shuying Yu and Tianxiang Yu and Enming Yuan and Mengjie Yuan and Tongtian Yue and Wei Yue and Yang Yue and Dunyuan Zha and Haobing Zhan and B. H. Zhang and Dehao Zhang and Fei Zhang and Hao Zhang and Haoyuan Zhang and Huanyu Zhang and Jiapei Zhang and Jiaxuan Zhang and Jin Zhang and Kaiyi Zhang and Miaozhen Zhang and Puqi Zhang and Qinglei Zhang and Rong Zhang and Rui Zhang and Shaoshuai Zhang and Shiyi Zhang and Xiaobin Zhang and Xiaoyun Zhang and Y. Zhang and Yangkun Zhang and Ye Zhang and Yichi Zhang and Yikun Zhang and Yizhi Zhang and Yongting Zhang and Yu Zhang and Yutao Zhang and Yutong Zhang and Zheng Zhang and Zijing Zhang and Bin Zhao and Chenguang Zhao and Feifan Zhao and Jinglun Zhao and Jinxiang Zhao and Shuai Zhao and Wenshuo Zhao and Xiangyu Zhao and Xuanle Zhao and Yikai Zhao and Zijia Zhao and Haozhi Zheng and Huabin Zheng and Ruihan Zheng and Shaojie Zheng and Tengyang Zheng and Haofeng Zhong and Lei Zhong and Longguang Zhong and M. Zhou and Qiankang Zhou and Runjie Zhou and Ruozhang Zhou and Xinyu Zhou and Yiqiao Zhou and Zaida Zhou and Jinguo Zhu and Liya Zhu and Xinhao Zhu and Yangjunfeng Zhu and Yuxuan Zhu and Zhen Zhu and Chen Zhuang and Weiyu Zhuang and Xinxing Zu},
      year={2026},
      eprint={2607.24653},
      archivePrefix={arXiv},
      primaryClass={cs.CL},
      url={https://arxiv.org/abs/2607.24653}, 
}

@misc{glm5team2026glm5vibecodingagentic,
      title={{GLM-5}: from Vibe Coding to Agentic Engineering}, 
      author={{GLM-5-Team} and Aohan Zeng and Xin Lv and Zhenyu Hou and Zhengxiao Du and Qinkai Zheng and Bin Chen and Da Yin and Chendi Ge and Chenghua Huang and Chengxing Xie and Chenzheng Zhu and Congfeng Yin and Cunxiang Wang and Gengzheng Pan and Hao Zeng and Haoke Zhang and Haoran Wang and Huilong Chen and Jiajie Zhang and Jian Jiao and Jiaqi Guo and Jingsen Wang and Jingzhao Du and Jinzhu Wu and Kedong Wang and Lei Li and Lin Fan and Lucen Zhong and Mingdao Liu and Mingming Zhao and Pengfan Du and Qian Dong and Rui Lu and Shuang-Li and Shulin Cao and Song Liu and Ting Jiang and Xiaodong Chen and Xiaohan Zhang and Xuancheng Huang and Xuezhen Dong and Yabo Xu and Yao Wei and Yifan An and Yilin Niu and Yitong Zhu and Yuanhao Wen and Yukuo Cen and Yushi Bai and Zhongpei Qiao and Zihan Wang and Zikang Wang and Zilin Zhu and Ziqiang Liu and Zixuan Li and Bojie Wang and Bosi Wen and Can Huang and Changpeng Cai and Chao Yu and Chen Li and Chengwei Hu and Chenhui Zhang and Dan Zhang and Daoyan Lin and Dayong Yang and Di Wang and Ding Ai and Erle Zhu and Fangzhou Yi and Feiyu Chen and Guohong Wen and Hailong Sun and Haisha Zhao and Haiyi Hu and Hanchen Zhang and Hanrui Liu and Hanyu Zhang and Hao Peng and Hao Tai and Haobo Zhang and He Liu and Hongwei Wang and Hongxi Yan and Hongyu Ge and Huan Liu and Huanpeng Chu and Jia'ni Zhao and Jiachen Wang and Jiajing Zhao and Jiamin Ren and Jiapeng Wang and Jiaxin Zhang and Jiayi Gui and Jiayue Zhao and Jijie Li and Jing An and Jing Li and Jingwei Yuan and Jinhua Du and Jinxin Liu and Junkai Zhi and Junwen Duan and Kaiyue Zhou and Kangjian Wei and Ke Wang and Keyun Luo and Laiqiang Zhang and Leigang Sha and Liang Xu and Lindong Wu and Lintao Ding and Lu Chen and Minghao Li and Nianyi Lin and Pan Ta and Qiang Zou and Rongjun Song and Ruiqi Yang and Shangqing Tu and Shangtong Yang and Shaoxiang Wu and Shengyan Zhang and Shijie Li and Shuang Li and Shuyi Fan and Wei Qin and Wei Tian and Weining Zhang and Wenbo Yu and Wenjie Liang and Xiang Kuang and Xiangmeng Cheng and Xiangyang Li and Xiaoquan Yan and Xiaowei Hu and Xiaoying Ling and Xing Fan and Xingye Xia and Xinyuan Zhang and Xinze Zhang and Xirui Pan and Xu Zou and Xunkai Zhang and Yadi Liu and Yandong Wu and Yanfu Li and Yidong Wang and Yifan Zhu and Yijun Tan and Yilin Zhou and Yiming Pan and Ying Zhang and Yinpei Su and Yipeng Geng and Yong Yan and Yonglin Tan and Yuean Bi and Yuhan Shen and Yuhao Yang and Yujiang Li and Yunan Liu and Yunqing Wang and Yuntao Li and Yurong Wu and Yutao Zhang and Yuxi Duan and Yuxuan Zhang and Zezhen Liu and Zhengtao Jiang and Zhenhe Yan and Zheyu Zhang and Zhixiang Wei and Zhuo Chen and Zhuoer Feng and Zijun Yao and Ziwei Chai and Ziyuan Wang and Zuzhou Zhang and Bin Xu and Minlie Huang and Hongning Wang and Juanzi Li and Yuxiao Dong and Jie Tang},
      year={2026},
      eprint={2602.15763},
      archivePrefix={arXiv},
      primaryClass={cs.LG},
      url={https://arxiv.org/abs/2602.15763}, 
}

@misc{yao2023reactsynergizingreasoningacting,
      title={{ReAct}: Synergizing Reasoning and Acting in Language Models}, 
      author={Shunyu Yao and Jeffrey Zhao and Dian Yu and Nan Du and Izhak Shafran and Karthik Narasimhan and Yuan Cao},
      year={2023},
      eprint={2210.03629},
      archivePrefix={arXiv},
      primaryClass={cs.CL},
      url={https://arxiv.org/abs/2210.03629}, 
}

@misc{shinn2023reflexionlanguageagentsverbal,
      title={{Reflexion}: Language Agents with Verbal Reinforcement Learning}, 
      author={Noah Shinn and Federico Cassano and Edward Berman and Ashwin Gopinath and Karthik Narasimhan and Shunyu Yao},
      year={2023},
      eprint={2303.11366},
      archivePrefix={arXiv},
      primaryClass={cs.AI},
      url={https://arxiv.org/abs/2303.11366}, 
}

@misc{kapoor2024aiagentsmatter,
      title={{AI} Agents That Matter}, 
      author={Sayash Kapoor and Benedikt Stroebl and Zachary S. Siegel and Nitya Nadgir and Arvind Narayanan},
      year={2024},
      eprint={2407.01502},
      archivePrefix={arXiv},
      primaryClass={cs.LG},
      url={https://arxiv.org/abs/2407.01502}, 
}

@misc{cemri2025multiagentllmsystemsfail,
      title={Why Do Multi-Agent {LLM} Systems Fail?}, 
      author={Mert Cemri and Melissa Z. Pan and Shuyi Yang and Lakshya A. Agrawal and Bhavya Chopra and Rishabh Tiwari and Kurt Keutzer and Aditya Parameswaran and Dan Klein and Kannan Ramchandran and Matei Zaharia and Joseph E. Gonzalez and Ion Stoica},
      year={2025},
      eprint={2503.13657},
      archivePrefix={arXiv},
      primaryClass={cs.AI},
      url={https://arxiv.org/abs/2503.13657}, 
}

@misc{xie2024osworldbenchmarkingmultimodalagents,
      title={{OSWorld}: Benchmarking Multimodal Agents for Open-Ended Tasks in Real Computer Environments}, 
      author={Tianbao Xie and Danyang Zhang and Jixuan Chen and Xiaochuan Li and Siheng Zhao and Ruisheng Cao and Toh Jing Hua and Zhoujun Cheng and Dongchan Shin and Fangyu Lei and Yitao Liu and Yiheng Xu and Shuyan Zhou and Silvio Savarese and Caiming Xiong and Victor Zhong and Tao Yu},
      year={2024},
      eprint={2404.07972},
      archivePrefix={arXiv},
      primaryClass={cs.AI},
      url={https://arxiv.org/abs/2404.07972}, 
}

@misc{drouin2024workarenacapablewebagents,
      title={{WorkArena}: How Capable Are Web Agents at Solving Common Knowledge Work Tasks?},
      author={Alexandre Drouin and Maxime Gasse and Massimo Caccia and Issam H. Laradji and Manuel Del Verme and Tom Marty and L{\'{e}}o Boisvert and Megh Thakkar and Quentin Cappart and David Vazquez and Nicolas Chapados and Alexandre Lacoste},
      year={2024},
      eprint={2403.07718},
      archivePrefix={arXiv},
      primaryClass={cs.LG},
      url={https://arxiv.org/abs/2403.07718},
}

@misc{qwen38,
    title = {{Qwen3.8-Max}: A New Bar for Coding and Cowork},
    url = {https://qwen.ai/blog?id=qwen3.8},
    author = {{Qwen Team}},
    month = {August},
    year = {2026}
}

@misc{openai2026gpt56,
  author = {{OpenAI}},
  title = {Using {GPT-5.6}},
  year = {2026},
  howpublished = {Official model documentation},
  url = {https://developers.openai.com/api/docs/guides/latest-model?model=gpt-5.6},

}

@misc{zai2026glm53,
  author = {{Z.ai}},
  title = {{GLM-5.3}},
  year = {2026},
  howpublished = {Official model documentation},
  url = {https://docs.z.ai/guides/llm/glm-5.3},

}

@misc{zai2026glm53flash,
  author = {{Z.ai}},
  title = {{GLM-5.3-Flash/FlashX}},
  year = {2026},
  howpublished = {Official model documentation},
  url = {https://docs.z.ai/guides/vlm/glm-5.3-flash},

}

@misc{qwen2026modelcard,
  author = {{Qwen Team}},
  title = {{Qwen3.8-27B} Model Card},
  year = {2026},
  howpublished = {Official model card},
  url = {https://huggingface.co/Qwen/Qwen3.8-27B},

}

@misc{anthropic2026claudecode,
  author = {{Anthropic}},
  title = {{Claude Code}},
  year = {2026},
  howpublished = {Software repository},
  url = {https://github.com/anthropics/claude-code},

}

@misc{openai2026codexcli,
  author = {{OpenAI}},
  title = {{Codex CLI}},
  year = {2026},
  howpublished = {Official software documentation},
  url = {https://learn.chatgpt.com/docs/codex/cli},

}

@misc{anomaly2026opencode,
  author = {{Anomaly}},
  title = {{OpenCode}: The Open Source Coding Agent},
  year = {2026},
  howpublished = {Software repository},
  url = {https://github.com/anomalyco/opencode},

}

@misc{openclaw2026software,
  author = {{OpenClaw Contributors}},
  title = {{OpenClaw}},
  year = {2026},
  howpublished = {Software repository},
  url = {https://github.com/openclaw/openclaw},

}

@misc{nous2026hermes,
  author = {{Nous Research}},
  title = {{Hermes Agent}},
  year = {2026},
  howpublished = {Software repository},
  url = {https://github.com/NousResearch/hermes-agent},

}

@misc{goose2026software,
  author = {{Goose Contributors}},
  title = {{Goose}},
  year = {2026},
  howpublished = {Software repository},
  url = {https://github.com/aaif-goose/goose},

}

@misc{hkuds2026nanobot,
  author = {{HKUDS}},
  title = {{Nanobot}: Ultra-Lightweight Personal {AI} Assistant},
  year = {2026},
  howpublished = {Software repository},
  url = {https://github.com/HKUDS/nanobot},

}

@misc{zhang2026clawbenchaiagentscomplete,
      title={{ClawBench}: Can {AI} Agents Complete Everyday Online Tasks?}, 
      author={Yuxuan Zhang and Yubo Wang and Yipeng Zhu and Penghui Du and Junwen Miao and Xuan Lu and Zhuofeng Li and Xingwei Qu and Zhengkang Guo and Yuanzhe Shen and Dingjie Song and Han Zhou and Tuney Zheng and Xian Wu and Hao Yu and Songcheng Cai and Yi Lu and Yunzhuo Hao and Minyi Lei and Liang Chen and Kai Zou and Huifeng Yin and Wendong Xu and Dongfu Jiang and Ping Nie and Jiaheng Liu and Wenhu Chen and Kelsey R. Allen},
      year={2026},
      eprint={2604.08523},
      archivePrefix={arXiv},
      primaryClass={cs.CL},
      url={https://arxiv.org/abs/2604.08523}, 
}

@misc{niu2026dumatebenchevaluatingautonomousagents,
      title={{DuMateBench}: Evaluating Autonomous Agents in Complex Real-World Workflows}, 
      author={Zechun Niu and Yukun Zhao and Jiaxin Zhang and Xu Shen and Jinhua Si and Han Tian and Can Xu and Yunfan Song and Jiaxin Mao and Yansong Gao and Yuchen Li and Jianmin Wu and Lingyong Yan and Shuaiqiang Wang and Dawei Yin},
      year={2026},
      eprint={2608.26546},
      archivePrefix={arXiv},
      primaryClass={cs.AI},
      url={https://arxiv.org/abs/2608.26546}, 
}

@misc{xu2024theagentcompanybenchmarkingllmagents,
      title={{TheAgentCompany}: Benchmarking {LLM} Agents on Consequential Real World Tasks}, 
      author={Frank F. Xu and Yufan Song and Boxuan Li and Yuxuan Tang and Kritanjali Jain and Mengxue Bao and Zora Z. Wang and Xuhui Zhou and Zhitong Guo and Murong Cao and Mingyang Yang and Hao Yang Lu and Amaad Martin and Zhe Su and Leander Maben and Raj Mehta and Wayne Chi and Lawrence Jang and Yiqing Xie and Shuyan Zhou and Graham Neubig},
      year={2024},
      eprint={2412.14161},
      archivePrefix={arXiv},
      primaryClass={cs.CL},
      url={https://arxiv.org/abs/2412.14161}, 
}

@misc{tang2026workspacebench10benchmarkingai,
      title={{Workspace-Bench} 1.0: Benchmarking {AI} Agents on Workspace Tasks with Large-Scale File Dependencies}, 
      author={Zirui Tang and Xuanhe Zhou and Yumou Liu and Linchun Li and Yukai Wu and Weizheng Wang and Hongzhang Huang and Wei Zhou and Jun Zhou and Jiachen Song and Shaoli Yu and Jinqi Wang and Zihang Zhou and Hongyi Zhou and Yuting Lv and Jinyang Li and Jiashuo Liu and Ruoyu Chen and Chunwei Liu and GuoLiang Li and Jihua Kang and Fan Wu},
      year={2026},
      eprint={2605.03596},
      archivePrefix={arXiv},
      primaryClass={cs.AI},
      url={https://arxiv.org/abs/2605.03596}, 
}

@misc{li2026agencybenchbenchmarkingfrontiersautonomous,
      title={{AgencyBench}: Benchmarking the Frontiers of Autonomous Agents in 1M-Token Real-World Contexts}, 
      author={Keyu Li and Junhao Shi and Yang Xiao and Mohan Jiang and Jie Sun and Yunze Wu and Dayuan Fu and Shijie Xia and Xiaojie Cai and Tianze Xu and Weiye Si and Wenjie Li and Dequan Wang and Pengfei Liu},
      year={2026},
      eprint={2601.11044},
      archivePrefix={arXiv},
      primaryClass={cs.AI},
      url={https://arxiv.org/abs/2601.11044}, 
}
\bibliographystyle{iclr2027_conference_short}

\clearpage
\appendix
\section{Limitations}
\label{app:limitations}

Each task--model--harness combination is evaluated once because of evaluation cost; repeated runs would quantify execution variability. Results cover ten models and seven harnesses at the evaluated versions, and 300 tasks drawn from selected everyday workflows. Their scope therefore depends on the tested configurations and task coverage. The task suite and evaluation pipeline provide a basis for extending both.

\section{Complete Results and Evaluation Protocol}
\label{app:results}
\subsection{Harness Versions and Generation Settings}
\label{app:harness_versions}

Table~\ref{tab:harness_versions} lists the harness versions recorded in the evaluation runs. Each harness uses the same version across its recorded general and multimodal evaluations. Temperature is set to 1.0, and the output-token limit is set to the maximum supported by each model. The input-context threshold is 200K tokens, with each harness retaining its default context-management mechanism. Runs are capped at 150 interaction turns, and each task--model--harness combination is evaluated once. Requests in the result tables count provider transactions, including auxiliary and delegated calls.

\begin{table}[!htbp]
\centering\small
\setlength{\tabcolsep}{12pt}\renewcommand{\arraystretch}{1.12}
\caption{Harness versions used in the evaluation.}
\label{tab:harness_versions}
\begin{tabular}{lc}
\toprule
Harness & Version \\
\midrule
Claude Code~\citep{anthropic2026claudecode} & 2.1.220 \\
Codex~\citep{openai2026codexcli} & 0.145.0 \\
OpenCode~\citep{anomaly2026opencode} & 1.18.8 \\
OpenClaw~\citep{openclaw2026software} & 2026.5.26 \\
Hermes~\citep{nous2026hermes} & 2026.7.20 \\
Goose~\citep{goose2026software} & 1.44.0 \\
Nanobot~\citep{hkuds2026nanobot} & 0.3.0 \\
\bottomrule
\end{tabular}
\end{table}

\subsection{Official API Prices}
API expenditure is estimated from token usage using the official prices in Table~\ref{tab:evaluation_prices}, with the same model-specific schedule across all harnesses. We account for uncached input, cache reads, cache writes, and output, counting reasoning already included in output only once.

\begin{table}[H]\centering\small\setlength{\tabcolsep}{4pt}\renewcommand{\arraystretch}{1.15}
\caption{Official API prices used throughout the paper, in USD per million tokens. Source names link to the official documentation.}\label{tab:evaluation_prices}
\begin{tabular*}{\linewidth}{@{\extracolsep{\fill}}lrrrrc@{}}\toprule
Model & Input & Cache read & Cache write & Output & Source \\\midrule
GPT-5.6 Sol & 4.000 & 0.400 & 5.000 & 20.000 & \href{https://developers.openai.com/api/docs/models/gpt-5.6-sol}{OpenAI} \\
GPT-5.6 Terra & 2.000 & 0.200 & 2.500 & 12.000 & \href{https://developers.openai.com/api/docs/models/gpt-5.6-terra}{OpenAI} \\
GPT-5.6 Luna & 0.200 & 0.020 & 0.250 & 1.200 & \href{https://developers.openai.com/api/docs/models/gpt-5.6-luna}{OpenAI} \\
\addlinespace
GLM-5.3 & 1.400 & 0.260 & 1.400 & 4.400 & \href{https://docs.z.ai/guides/overview/pricing}{Z.AI} \\
GLM-5.3-Flash & 0.150 & 0.030 & 0.150 & 0.500 & \href{https://docs.z.ai/guides/overview/pricing}{Z.AI} \\
\addlinespace
DeepSeek-V4-Pro & 1.320 & 0.044 & 1.320 & 3.960 & \href{https://api-docs.deepseek.com/quick_start/pricing/}{DeepSeek} \\
DeepSeek-V4-Flash & 0.140 & 0.028 & 0.140 & 0.280 & \href{https://api-docs.deepseek.com/quick_start/agent_integrations/pi_mono/}{DeepSeek} \\
\addlinespace
Kimi-K3 & 3.000 & 0.300 & 3.000 & 15.000 & \href{https://www.kimi.com/en/blog/kimi-k3}{Kimi}, \href{https://platform.kimi.ai/docs/guide/use-context-caching-feature-of-kimi-api}{cache} \\
\bottomrule\end{tabular*}
\vspace{3pt}\begin{minipage}{\linewidth}\footnotesize\raggedright
Input excludes cache reads and separately reported writes. Cache-write tokens use the ordinary input rate for GLM and DeepSeek. Kimi uses its default five-minute cache-write rate. DeepSeek Pro uses the official peak schedule, and DeepSeek Flash uses its V4-specific prices.
\end{minipage}\end{table}

\subsection{Detailed Configuration Results}
\label{app:detailed_results}
Tables~\ref{tab:detailed_general} and~\ref{tab:detailed_multimodal} report each model--harness configuration on the general and multimodal suites. Score is the mean oracle score; $R_{0.8}$ and $R_1$ are percentages. Traj. (K), Tokens (K), Requests, and Cost (USD) are per-task averages. Request counts include auxiliary and delegated model calls; token totals include cached input. Bold marks the best quality value within each model, including ties.

\paragraph{Trajectory length.}
Traj. estimates the length of newly introduced task text, tool results, and generated output, including reported reasoning tokens. It excludes replayed context, static system prompts, tool schemas, and multimodal input payloads. Text is counted with \texttt{o200k\_base}, and reported output usage accounts for hidden reasoning.

\begingroup\small
\setlength{\tabcolsep}{3.5pt}\renewcommand{\arraystretch}{1.12}
\begin{longtable}{@{}llrrrrrrr@{}}
\caption{General-task results for every model--harness configuration (274 tasks per configuration).}\label{tab:detailed_general}\\
\toprule
Model & Harness & Score & $R_{0.8}$ & $R_1$ & Traj. (K) & Tokens (K) & Requests & Cost (\$) \\
\midrule
\endfirsthead
\multicolumn{9}{l}{\textit{General-task results (continued)}}\\
\toprule
Model & Harness & Score & $R_{0.8}$ & $R_1$ & Traj. (K) & Tokens (K) & Requests & Cost (\$) \\
\midrule
\endhead
\midrule\multicolumn{9}{r}{\textit{Continued on next page}}\\
\endfoot
\bottomrule\endlastfoot
GPT-5.6 Sol & Claude Code & 71.60 & 58.03 & 22.99 & 14.5 & 135.2 & 11.4 & 0.3477 \\*
 & Codex & 69.83 & 57.30 & 24.82 & 15.4 & 288.1 & 15.0 & 0.3811 \\*
 & OpenCode & 71.03 & 57.30 & 27.01 & 14.9 & 172.8 & 12.1 & 0.2963 \\*
 & OpenClaw & \textbf{73.08} & 56.93 & 31.02 & 13.1 & 149.8 & 10.7 & 0.3193 \\*
 & Hermes & 71.18 & \textbf{58.39} & \textbf{32.12} & 16.9 & 170.0 & 13.9 & 0.3217 \\*
 & Goose & 71.51 & 55.47 & 29.20 & 13.0 & 76.2 & 9.4 & 0.2746 \\*
 & Nanobot & 66.49 & 54.38 & 24.82 & 26.8 & 806.2 & 38.5 & 0.9483 \\
\addlinespace[4pt]
GPT-5.6 Terra & Claude Code & 68.53 & 51.09 & 22.26 & 12.6 & 81.9 & 10.7 & 0.1343 \\*
 & Codex & 66.76 & 52.92 & 18.61 & 5.0 & 141.8 & 9.5 & 0.1199 \\*
 & OpenCode & 67.05 & 53.28 & 20.80 & 11.4 & 121.8 & 10.0 & 0.1675 \\*
 & OpenClaw & 67.88 & \textbf{54.38} & \textbf{26.64} & 8.9 & 66.6 & 8.4 & 0.0967 \\*
 & Hermes & 68.42 & 54.01 & 25.18 & 12.0 & 119.0 & 11.8 & 0.1695 \\*
 & Goose & \textbf{69.65} & 54.01 & 23.72 & 10.7 & 63.2 & 8.3 & 0.1296 \\*
 & Nanobot & 64.18 & 48.54 & 21.17 & 10.1 & 147.3 & 10.4 & 0.1820 \\
\addlinespace[4pt]
GPT-5.6 Luna & Claude Code & \textbf{69.54} & \textbf{53.65} & 24.09 & 14.9 & 109.1 & 12.1 & 0.0167 \\*
 & Codex & 67.31 & 52.19 & 21.53 & 5.9 & 145.8 & 9.7 & 0.0131 \\*
 & OpenCode & 66.92 & 52.92 & 20.44 & 11.6 & 144.3 & 10.8 & 0.0198 \\*
 & OpenClaw & 67.75 & 51.82 & \textbf{26.28} & 8.7 & 71.8 & 9.1 & 0.0103 \\*
 & Hermes & 68.00 & 53.28 & 23.72 & 11.6 & 132.3 & 12.5 & 0.0178 \\*
 & Goose & 68.31 & 53.28 & 22.63 & 9.5 & 68.2 & 9.2 & 0.0140 \\*
 & Nanobot & 64.51 & 48.91 & 22.26 & 10.2 & 163.7 & 11.2 & 0.0195 \\
\addlinespace[4pt]
GLM-5.3 & Claude Code & 68.69 & 51.46 & 27.74 & 18.2 & 264.0 & 18.0 & 0.2081 \\*
 & Codex & 68.58 & 52.92 & 22.26 & 15.1 & 522.2 & 20.0 & 0.5292 \\*
 & OpenCode & \textbf{70.34} & 54.01 & \textbf{35.77} & 29.8 & 504.8 & 16.4 & 0.4603 \\*
 & OpenClaw & 67.44 & 52.55 & 26.64 & 36.1 & 219.2 & 16.6 & 0.2063 \\*
 & Hermes & 66.27 & 51.46 & 29.20 & 53.9 & 387.1 & 19.8 & 0.4176 \\*
 & Goose & 67.96 & \textbf{54.74} & 32.12 & 27.6 & 559.7 & 15.9 & 0.6751 \\*
 & Nanobot & 61.81 & 49.27 & 28.47 & 38.7 & 1422.0 & 29.0 & 1.1655 \\
\addlinespace[4pt]
GLM-5.3-Flash & Claude Code & 66.30 & 52.19 & 26.64 & 14.7 & 100.6 & 13.0 & 0.0088 \\*
 & Codex & 64.77 & 51.46 & 20.07 & 13.6 & 468.0 & 19.4 & 0.0461 \\*
 & OpenCode & 69.47 & 54.38 & 29.56 & 18.6 & 438.1 & 14.7 & 0.0526 \\*
 & OpenClaw & 69.00 & 52.92 & 28.47 & 20.5 & 147.6 & 14.3 & 0.0134 \\*
 & Hermes & 67.36 & 53.28 & 27.74 & 22.9 & 258.7 & 17.1 & 0.0289 \\*
 & Goose & \textbf{69.85} & \textbf{55.11} & \textbf{33.58} & 17.2 & 404.4 & 13.2 & 0.0577 \\*
 & Nanobot & 67.18 & \textbf{55.11} & 28.47 & 21.0 & 933.2 & 19.0 & 0.1054 \\
\addlinespace[4pt]
DeepSeek-V4-Pro & Claude Code & 68.67 & 54.38 & 25.18 & 34.3 & 378.4 & 17.3 & 0.1822 \\*
 & Codex & \textbf{69.54} & \textbf{58.39} & 27.01 & 27.2 & 503.3 & 17.3 & 0.1222 \\*
 & OpenCode & 68.97 & 55.84 & \textbf{29.20} & 22.2 & 248.5 & 12.5 & 0.1368 \\*
 & OpenClaw & 64.35 & 53.28 & 27.01 & 36.4 & 198.3 & 12.9 & 0.1792 \\*
 & Hermes & 67.32 & 55.47 & 28.83 & 31.1 & 314.3 & 14.5 & 0.1920 \\*
 & Goose & 66.04 & 52.19 & 25.91 & 27.1 & 317.1 & 14.6 & 0.3120 \\*
 & Nanobot & 64.89 & 53.28 & 25.18 & 30.7 & 780.5 & 22.0 & 0.2479 \\
\addlinespace[4pt]
DeepSeek-V4-Flash & Claude Code & 64.81 & 51.46 & 25.18 & 48.0 & 562.8 & 21.3 & 0.0329 \\*
 & Codex & \textbf{68.34} & 52.19 & 27.74 & 27.4 & 447.0 & 15.7 & 0.0202 \\*
 & OpenCode & 66.24 & 50.73 & 26.28 & 25.4 & 371.9 & 15.9 & 0.0211 \\*
 & OpenClaw & 66.78 & 51.46 & 28.47 & 38.8 & 236.5 & 14.6 & 0.0199 \\*
 & Hermes & 67.35 & \textbf{55.47} & \textbf{29.93} & 30.7 & 346.9 & 16.4 & 0.0232 \\*
 & Goose & 65.91 & 51.46 & 28.47 & 24.5 & 259.7 & 13.6 & 0.0235 \\*
 & Nanobot & 60.92 & 45.99 & 23.72 & 28.5 & 629.9 & 19.4 & 0.0327 \\
\addlinespace[4pt]
Kimi-K3 & Claude Code & 60.01 & 48.91 & 25.18 & 18.1 & 199.3 & 17.6 & 0.3522 \\*
 & Codex & 65.12 & 52.55 & 26.64 & 17.6 & 307.4 & 15.1 & 0.3537 \\*
 & OpenCode & \textbf{69.49} & \textbf{56.20} & \textbf{32.85} & 21.3 & 271.9 & 15.0 & 0.5583 \\*
 & OpenClaw & 67.27 & 51.82 & 28.83 & 15.0 & 130.7 & 11.8 & 0.3132 \\*
 & Hermes & 68.16 & 54.01 & 27.74 & 20.4 & 217.9 & 16.6 & 0.4596 \\*
 & Goose & 66.95 & 52.92 & 29.56 & 17.2 & 153.1 & 11.1 & 0.5172 \\*
 & Nanobot & 65.18 & 52.92 & 31.02 & 19.7 & 297.5 & 13.8 & 0.5726 \\
\addlinespace[4pt]
Gemma-4-31B & Claude Code & 59.04 & 42.34 & 24.45 & 14.9 & 451.6 & 32.9 & --- \\*
 & Codex & 58.87 & \textbf{42.70} & \textbf{26.64} & 13.6 & 302.3 & 16.6 & --- \\*
 & OpenCode & 59.27 & 39.78 & 21.53 & 13.9 & 228.7 & 15.4 & --- \\*
 & OpenClaw & 58.78 & 40.88 & 23.36 & 21.7 & 180.6 & 16.0 & --- \\*
 & Hermes & \textbf{59.63} & 38.32 & 22.99 & 20.8 & 224.4 & 17.4 & --- \\*
 & Goose & 52.69 & 39.78 & 24.82 & 11.2 & 217.9 & 17.9 & --- \\*
 & Nanobot & 52.24 & 39.05 & 25.55 & 11.1 & 250.1 & 15.2 & --- \\
\addlinespace[4pt]
Qwen3.8-27B & Claude Code & 61.67 & 46.35 & 28.83 & 29.8 & 359.2 & 15.1 & --- \\*
 & Codex & 62.42 & 48.18 & 29.56 & 26.7 & 403.3 & 14.9 & --- \\*
 & OpenCode & \textbf{65.10} & \textbf{51.09} & \textbf{33.21} & 28.4 & 391.2 & 15.8 & --- \\*
 & OpenClaw & 58.32 & 45.62 & 26.28 & 27.0 & 211.0 & 14.2 & --- \\*
 & Hermes & 61.75 & 47.81 & 25.18 & 35.6 & 271.3 & 19.8 & --- \\*
 & Goose & 25.91 & 22.63 & 17.88 & 13.4 & 106.8 & 8.2 & --- \\*
 & Nanobot & 34.25 & 28.10 & 17.15 & 19.8 & 317.7 & 12.1 & --- \\
\addlinespace[4pt]
\end{longtable}
\endgroup

\begingroup\small
\setlength{\tabcolsep}{3.5pt}\renewcommand{\arraystretch}{1.12}
\begin{longtable}{@{}llrrrrrrr@{}}
\caption{Multimodal results for every model--harness configuration (26 tasks per configuration).}\label{tab:detailed_multimodal}\\
\toprule
Model & Harness & Score & $R_{0.8}$ & $R_1$ & Traj. (K) & Tokens (K) & Requests & Cost (\$) \\
\midrule
\endfirsthead
\multicolumn{9}{l}{\textit{Multimodal results (continued)}}\\
\toprule
Model & Harness & Score & $R_{0.8}$ & $R_1$ & Traj. (K) & Tokens (K) & Requests & Cost (\$) \\
\midrule
\endhead
\midrule\multicolumn{9}{r}{\textit{Continued on next page}}\\
\endfoot
\bottomrule\endlastfoot
GPT-5.6 Sol & Claude Code & 68.38 & 11.54 & 11.54 & 16.4 & 100.7 & 14.1 & 0.2882 \\*
 & Codex & \textbf{68.81} & \textbf{23.08} & \textbf{19.23} & 12.1 & 307.9 & 15.2 & 0.4247 \\*
 & OpenCode & 60.31 & 19.23 & \textbf{19.23} & 21.7 & 305.7 & 17.4 & 0.4897 \\*
 & OpenClaw & 60.73 & 11.54 & 11.54 & 14.3 & 154.3 & 16.3 & 0.3656 \\*
 & Hermes & 66.31 & 15.38 & 15.38 & 24.4 & 749.1 & 32.3 & 0.8219 \\*
 & Goose & 54.50 & 3.85 & 3.85 & 3.8 & 81.3 & 8.7 & 0.2573 \\*
 & Nanobot & 65.65 & 0.00 & 0.00 & 17.3 & 427.3 & 21.7 & 0.5941 \\
\addlinespace[4pt]
GPT-5.6 Terra & Claude Code & 58.00 & 3.85 & 3.85 & 9.9 & 58.4 & 10.3 & 0.0982 \\*
 & Codex & \textbf{70.46} & 15.38 & 15.38 & 13.6 & 364.2 & 17.4 & 0.2667 \\*
 & OpenCode & 56.65 & \textbf{23.08} & \textbf{23.08} & 12.2 & 126.3 & 10.3 & 0.1401 \\*
 & OpenClaw & 50.31 & \textbf{23.08} & \textbf{23.08} & 12.1 & 144.0 & 11.4 & 0.1815 \\*
 & Hermes & 60.73 & 7.69 & 7.69 & 18.2 & 549.2 & 27.7 & 0.5755 \\*
 & Goose & 57.54 & 19.23 & 19.23 & 3.9 & 90.6 & 10.0 & 0.1267 \\*
 & Nanobot & 67.54 & 19.23 & 19.23 & 10.0 & 232.3 & 15.7 & 0.2021 \\
\addlinespace[4pt]
GPT-5.6 Luna & Claude Code & 58.65 & 23.08 & 19.23 & 11.7 & 70.0 & 11.7 & 0.0118 \\*
 & Codex & 62.88 & 11.54 & 7.69 & 14.8 & 334.4 & 15.8 & 0.0282 \\*
 & OpenCode & 36.54 & 7.69 & 3.85 & 12.5 & 123.8 & 10.0 & 0.0144 \\*
 & OpenClaw & 58.69 & 23.08 & \textbf{23.08} & 15.2 & 191.2 & 13.8 & 0.0222 \\*
 & Hermes & 58.96 & 7.69 & 7.69 & 26.0 & 798.4 & 37.3 & 0.0825 \\*
 & Goose & 51.19 & 19.23 & 15.38 & 3.2 & 78.3 & 9.8 & 0.0117 \\*
 & Nanobot & \textbf{64.38} & \textbf{26.92} & 19.23 & 7.0 & 208.4 & 14.4 & 0.0211 \\
\addlinespace[4pt]
GLM-5.3-Flash & Claude Code & 52.92 & 23.08 & 19.23 & 50.6 & 1513.7 & 45.2 & 0.1644 \\*
 & Codex & 71.19 & 26.92 & 23.08 & 29.7 & 1172.3 & 37.8 & 0.0884 \\*
 & OpenCode & 57.69 & 26.92 & 26.92 & 48.7 & 1671.2 & 39.2 & 0.1781 \\*
 & OpenClaw & \textbf{76.69} & \textbf{50.00} & \textbf{50.00} & 34.3 & 397.5 & 33.3 & 0.0311 \\*
 & Hermes & 64.65 & 42.31 & 42.31 & 69.3 & 1783.6 & 45.6 & 0.1841 \\*
 & Goose & 65.08 & 30.77 & 30.77 & 32.7 & 1703.4 & 28.9 & 0.2366 \\*
 & Nanobot & 57.46 & 30.77 & 26.92 & 33.3 & 4402.4 & 67.3 & 0.4843 \\
\addlinespace[4pt]
Kimi-K3 & Claude Code & 60.92 & 26.92 & 23.08 & 37.9 & 866.8 & 41.5 & 1.4662 \\*
 & Codex & 28.08 & 0.00 & 0.00 & 33.0 & 1228.6 & 36.4 & 2.2851 \\*
 & OpenCode & 64.35 & 26.92 & 26.92 & 56.9 & 1917.1 & 51.4 & 2.7277 \\*
 & OpenClaw & \textbf{72.73} & \textbf{30.77} & \textbf{30.77} & 21.8 & 472.4 & 26.9 & 0.8926 \\*
 & Hermes & 57.92 & 23.08 & 23.08 & 57.3 & 2682.0 & 77.1 & 2.9275 \\*
 & Goose & 68.73 & \textbf{30.77} & \textbf{30.77} & 46.0 & 1120.0 & 34.0 & 2.9737 \\*
 & Nanobot & 53.04 & 26.92 & 26.92 & 44.9 & 1956.1 & 50.3 & 3.0120 \\
\addlinespace[4pt]
Gemma-4-31B & Claude Code & \textbf{45.12} & \textbf{15.38} & \textbf{0.00} & 35.7 & 576.3 & 33.3 & --- \\*
 & Codex & 17.77 & 0.00 & \textbf{0.00} & 10.6 & 297.6 & 19.7 & --- \\*
 & OpenCode & 28.15 & 3.85 & \textbf{0.00} & 24.1 & 865.0 & 30.6 & --- \\*
 & OpenClaw & 29.19 & 7.69 & \textbf{0.00} & 34.8 & 602.0 & 32.0 & --- \\*
 & Hermes & 23.69 & 3.85 & \textbf{0.00} & 37.5 & 977.6 & 42.3 & --- \\*
 & Goose & 39.50 & 7.69 & \textbf{0.00} & 19.4 & 354.1 & 29.4 & --- \\*
 & Nanobot & 32.31 & 0.00 & \textbf{0.00} & 17.6 & 490.0 & 28.3 & --- \\
\addlinespace[4pt]
Qwen3.8-27B & Claude Code & \textbf{69.04} & 30.77 & \textbf{30.77} & 87.5 & 1400.3 & 48.8 & --- \\*
 & Codex & 47.38 & \textbf{34.62} & \textbf{30.77} & 122.9 & 912.8 & 31.7 & --- \\*
 & OpenCode & 45.92 & 23.08 & 23.08 & 110.5 & 1684.7 & 48.7 & --- \\*
 & OpenClaw & 55.88 & 30.77 & \textbf{30.77} & 94.3 & 1104.0 & 41.7 & --- \\*
 & Hermes & 50.23 & 23.08 & 23.08 & 108.9 & 1544.2 & 48.3 & --- \\*
 & Goose & 50.88 & 15.38 & 15.38 & 83.3 & 1078.7 & 36.9 & --- \\*
 & Nanobot & 31.58 & 3.85 & 3.85 & 104.2 & 3317.3 & 87.7 & --- \\
\addlinespace[4pt]
\end{longtable}
\endgroup

\section{Analysis Protocol and Detailed Results}
\label{app:analysis_protocol}

The following subsections supplement the model-dependence, task-requirement, and cost analyses in Section~\ref{sec:analysis}. General-task comparisons use all 274 tasks for each of the ten models and seven harnesses; paired contrasts hold the task and model fixed. API costs follow Appendix~\ref{app:results}. Unless otherwise specified, we use 2,000 task-bootstrap resamples (seed 20260923), retaining all configuration outcomes for each selected task, and report percentile 95\% intervals.

\subsection{Harness Dependence across Models}
\label{app:model_dependence}

\paragraph{Sensitivity across harnesses.}

For each model, we compute the population SD of its seven configuration-level mean scores (divisor seven). This measures variation across harnesses, not variation across tasks or repeated runs. Correlations treat models as observations. Table~\ref{tab:harness_sensitivity} reports the mean-score SD and the range of full-credit rates. Across models, Spearman correlations between mean oracle score and the cross-harness SD of $R_{0.8}$ and $R_1$ are $-0.73$ and $+0.16$, respectively. The corresponding mean-score--SD correlation on the separately evaluated seven-model multimodal suite is $-0.57$.

\begin{table}[!htbp]

\centering\small

\begin{tabular}{lccc}

\toprule

Model & Mean score & Cross-harness SD & $R_1$ range (pp) \\

\midrule

GPT Sol & 70.67 & 1.93 & 9.12 \\

GLM Flash & 67.70 & 1.71 & 13.50 \\

GPT Terra & 67.49 & 1.62 & 8.03 \\

GPT Luna & 67.48 & 1.44 & 5.84 \\

GLM-5.3 & 67.30 & 2.52 & 13.50 \\

DS Pro & 67.11 & 1.91 & 4.01 \\

Kimi K3 & 66.03 & 2.85 & 7.66 \\

DS Flash & 65.76 & 2.23 & 6.20 \\

Gemma 4 & 57.22 & 3.02 & 5.11 \\

Qwen3.8 & 52.77 & 14.64 & 16.06 \\

\bottomrule

\end{tabular}

\caption{Harness sensitivity on the general task suite. Mean score averages all seven harnesses; SD is computed across their mean scores. The $R_1$ range is the best-minus-worst full-credit rate across the same harnesses.}

\label{tab:harness_sensitivity}

\end{table}

\paragraph{Model--harness interactions.}
Interaction residuals subtract the row and column means and add back the grand mean on the full general task suite. Their values depend on the set of evaluated models: excluding Qwen3.8 changes Kimi-K3--Claude Code's residual from $-6.95$ to $-6.06$ points, while GLM-5.3-Flash--Goose changes from $+4.62$ to $+1.92$ points. Residuals locate departures from the specified additive baseline; execution traces are needed to explain them.

\paragraph{Model-specific harness selection.}

One policy chooses a single harness for all models; the other chooses one per model. Both use training tasks only and are evaluated on held-out tasks. Tasks that share a grading template are kept together so that the same template does not appear on both sides of the split. We use two template definitions: identical scoring rubrics, and matching module weights, requirement-role metadata, and gates. Table~\ref{tab:selection_objectives} averages 20 repetitions of grouped five-fold evaluation. Full-credit gains are positive in all 20 repetitions under both definitions.

\begin{table}[!htbp]

\centering

\small

\begin{tabular}{lrrr}

\toprule

Held-out groups & Mean score & $R_{0.8}$ & $R_1$ \\

\midrule

Exact rubric & +0.14 & +0.25 & +2.73 \\

Scoring structure & -0.14 & +0.31 & +2.53 \\

\bottomrule

\end{tabular}

\caption{Held-out gains from selecting a harness per model rather than one globally. Each policy optimizes the reported metric on training tasks only. Rows average 20 repetitions of grouped five-fold evaluation. Rate gains are percentage points.}

\label{tab:selection_objectives}

\end{table}

\subsection{Task Requirements and Complementarity}
\label{app:task_requirements}

\paragraph{Category-level delivery.}

Table~\ref{tab:category_outcomes} uses the current six-category taxonomy. Multimodal tasks require multimodal model capabilities during task execution and are evaluated separately.

\begin{table}[!htbp]\centering\small\begin{tabular}{lrrrr}\toprule Category & Tasks & Score & $R_{0.8}$ & $R_1$ \\ \midrule

Business Operations & 71 & 68.31 & 55.19 & 21.71 \\

Content Production & 16 & 69.55 & 58.04 & 41.07 \\

Research and Analysis & 40 & 64.55 & 53.00 & 25.46 \\

Software Engineering & 81 & 57.34 & 39.37 & 22.75 \\

Systems Engineering & 65 & 69.87 & 56.66 & 32.66 \\

Web Development & 1 & 67.36 & 82.86 & 1.43 \\

\bottomrule\end{tabular}\caption{Task-category outcomes on the full general suite. Each task contributes 70 attempts.}\label{tab:category_outcomes}\end{table}

\paragraph{Complementary full-credit outcomes.}
\label{app:task_complementarity}

With DeepSeek Pro, Codex and OpenCode differ by only 0.57 mean-score points, yet their full-credit sets overlap on 49 tasks, with 25 completed only by Codex and 31 only by OpenCode; neither receives full credit on the remaining 169. Their union exceeds the better fixed configuration by 9.12 percentage points, an observed upper bound for task-level selection. Among the 56 discordant tasks, only 21 retain the same winner when score differences are averaged across the other nine models. Complementary task outcomes therefore depend on the model.

Appendices~\ref{app:trajectory_annotation} and~\ref{app:trajectory_cases} provide the annotation protocol and paired execution evidence; oracle scores remain artifact-based.

\paragraph{Delivery by execution requirement.}
Task specifications carry overlapping execution-demand tags, assigned independently of the observed scores. We group the 274 general tasks by these tags and retain all ten models and seven harnesses on every task in each group. Tables~\ref{tab:demand_scores} and~\ref{tab:demand_full_credit} report equally weighted task and model averages. Tags identify intended demands, not isolated interventions; an untagged task may still involve the same behavior. All tasks carry a multi-step execution tag, which is omitted because it reproduces the full-suite comparison.

\begin{table}[!htbp]\centering\small\setlength{\tabcolsep}{3pt}

\caption{Mean oracle scores by task-design demand. $N$ counts tasks, each evaluated under all 70 general configurations. Best harness average within each demand is bold. Groups overlap.}\label{tab:demand_scores}

\begin{tabular*}{\linewidth}{@{\extracolsep{\fill}}lrccccccc@{}}\toprule

Demand & $N$ & Claude Code & Codex & OpenCode & OpenClaw & Hermes & Goose & Nanobot \\\midrule

Evidence grounding & 198 & 67.8 & 68.2 & \textbf{69.2} & 68.2 & 68.4 & 65.0 & 62.5 \\

State management & 126 & 71.7 & 72.4 & \textbf{73.1} & 71.7 & 71.8 & 67.9 & 65.1 \\

Tool selection & 114 & 68.7 & 69.0 & \textbf{70.3} & 67.8 & 69.3 & 63.4 & 60.6 \\

Error recovery & 71 & 80.9 & 81.1 & \textbf{82.9} & 81.0 & 81.3 & 77.4 & 72.0 \\

Verification & 236 & 65.4 & 65.8 & \textbf{67.1} & 65.5 & 65.8 & 61.8 & 59.4 \\

Output contract & 87 & 57.2 & 57.3 & 58.6 & \textbf{59.1} & 58.1 & 55.7 & 54.3 \\

Safety boundary & 166 & 69.6 & 69.6 & \textbf{70.6} & 69.9 & 70.2 & 66.7 & 63.5 \\

Context management & 12 & 77.7 & \textbf{80.2} & 79.8 & 75.9 & 79.1 & 73.8 & 67.6 \\

\bottomrule\end{tabular*}\end{table}

\begin{table}[!htbp]\centering\small\setlength{\tabcolsep}{3pt}

\caption{Full-credit attainment by task-design demand. $N$ counts tasks, each evaluated under all 70 general configurations. Best harness average within each demand is bold. Groups overlap.}\label{tab:demand_full_credit}

\begin{tabular*}{\linewidth}{@{\extracolsep{\fill}}lrccccccc@{}}\toprule

Demand & $N$ & Claude Code & Codex & OpenCode & OpenClaw & Hermes & Goose & Nanobot \\\midrule

Evidence grounding & 198 & 26.6 & 26.0 & \textbf{29.1} & 28.3 & 27.6 & 28.4 & 26.2 \\

State management & 126 & 23.0 & 22.8 & \textbf{25.7} & 24.0 & 23.8 & 24.8 & 22.8 \\

Tool selection & 114 & 24.4 & 23.3 & 26.5 & 26.1 & \textbf{27.5} & 25.3 & 23.7 \\

Error recovery & 71 & 32.5 & 31.8 & 35.2 & 35.1 & 33.4 & \textbf{36.1} & 32.0 \\

Verification & 236 & 25.7 & 25.0 & \textbf{28.1} & 27.5 & 27.3 & 26.8 & 25.0 \\

Output contract & 87 & 25.9 & 24.6 & \textbf{28.3} & 27.6 & 26.2 & 25.5 & 24.3 \\

Safety boundary & 166 & 26.2 & 25.4 & 28.4 & \textbf{28.7} & 27.0 & 28.2 & 25.3 \\

Context management & 12 & 19.2 & 21.7 & \textbf{22.5} & \textbf{22.5} & 19.2 & 20.0 & 16.7 \\

\bottomrule\end{tabular*}\end{table}

\paragraph{The same demand does not select the same harness for every model.}
Across the 71 recovery-tagged tasks, Goose has the highest model-averaged full-credit rate (36.06\%), but the paired OpenCode--Goose contrast favors OpenCode for Gemma and Qwen and Goose for the other eight models (Table~\ref{tab:recovery_model_pairs}). For Gemma, OpenCode's full-credit advantage is 8.45 percentage points on recovery-tagged tasks and reverses to $-7.39$ on the remaining 203 tasks. Its mean-score advantage also grows from 3.28 to 16.04 points. Harness preferences vary across both models and requirement-defined task groups.

\begin{table}[!htbp]\centering\small\setlength{\tabcolsep}{7pt}

\caption{Paired OpenCode minus Goose differences within each model. Recovery: 71 recovery-tagged tasks; remaining: 203 other tasks. Score differences are points; $R_1$ differences are percentage points. Positive values favor OpenCode.}\label{tab:recovery_model_pairs}

\begin{tabular}{lrrrr}\toprule

& \multicolumn{2}{c}{Mean score difference} & \multicolumn{2}{c}{$R_1$ difference} \\

\cmidrule(lr){2-3}\cmidrule(l){4-5}

Model & Recovery & Remaining & Recovery & Remaining \\\midrule

GPT Sol & +1.06 & -1.03 & -4.23 & -1.48 \\

GPT Terra & -0.90 & -3.18 & -2.82 & -2.96 \\

GPT Luna & -2.48 & -1.01 & -4.23 & -1.48 \\

GLM-5.3 & +0.56 & +3.02 & -2.82 & +5.91 \\

GLM Flash & -3.75 & +0.80 & -9.86 & -1.97 \\

DS Pro & +2.28 & +3.16 & -1.41 & +4.93 \\

DS Flash & -1.10 & +0.84 & -4.23 & -1.48 \\

Kimi K3 & -0.65 & +3.66 & -4.23 & +5.91 \\

Gemma 4 & +16.04 & +3.28 & +8.45 & -7.39 \\

Qwen3.8 & +43.45 & +37.69 & +16.90 & +14.78 \\

\bottomrule\end{tabular}\end{table}

\paragraph{Checking work-category composition.}
For the Gemma contrast, we also compare the two task groups within each work category, then average these differences using category task counts as weights. Five categories contain both groups: Business Operations (34 recovery / 37 remaining), Content Production (4/12), Research and Analysis (9/31), Software Engineering (5/76), and Systems Engineering (18/47). Across these 273 tasks, the category-standardized difference between the two groups remains 10.80 score points and 24.87 percentage points in $R_1$. Thus, differing work-category proportions alone do not account for the observed contrast. Demand tags remain overlapping in this analysis.

\paragraph{Execution mechanisms recur across distinct tasks.}
Evidence-and-requirement, state-and-artifact, and verification errors each span more than 100 tasks and all ten models and seven harnesses (Table~\ref{tab:mechanism_task_coverage}). These annotations describe observed behavior, complementing the task-design demand tags.

\begin{table}[!htbp]\centering\small\setlength{\tabcolsep}{5pt}

\caption{Coverage of observed mechanisms in the unchanged base samples. Runs and tasks are distinct within each category; categories overlap. Models/harnesses gives the number of represented backends and harnesses. Full counts labeled runs that nevertheless receive full credit.}\label{tab:mechanism_task_coverage}

\begin{tabular}{llrrrr}\toprule

Suite & Mechanism & Runs & Tasks & Models/harnesses & Full \\\midrule

General & Evidence and requirements & 596 & 110 & 10/7 & 68 \\

 & Tool invocation & 694 & 193 & 10/7 & 197 \\

 & Recovery & 262 & 122 & 10/7 & 67 \\

 & Execution stagnation & 55 & 36 & 9/7 & 0 \\

 & State and artifacts & 399 & 143 & 10/7 & 75 \\

 & Verification & 948 & 160 & 10/7 & 24 \\

\midrule

Multimodal & Evidence and requirements & 66 & 13 & 7/7 & 5 \\

 & Tool invocation & 34 & 14 & 7/7 & 11 \\

 & Recovery & 27 & 14 & 7/5 & 3 \\

 & Execution stagnation & 9 & 5 & 3/7 & 0 \\

 & State and artifacts & 21 & 15 & 6/7 & 0 \\

 & Verification & 217 & 25 & 7/7 & 2 \\

\bottomrule\end{tabular}\end{table}

\paragraph{Harness-level annotation profiles.}
Table~\ref{tab:harness_error_mechanisms} aggregates the base samples with equal model coverage: 800 general runs and 84 multimodal runs per harness. The profiles distinguish tool-invocation, recovery, and state-and-artifact problems across harnesses. Frequencies describe the score-stratified samples rather than benchmark-wide incidence; configuration-level breakdowns remain in Appendix~\ref{app:config_errors}.

\begin{table}[!htbp]\centering\small\setlength{\tabcolsep}{5pt}

\caption{Runs with each mechanism label, by harness and base sample. E: evidence and requirements; T: tool invocation; R: recovery; P: execution stagnation; S: state and artifacts; V: verification. Labels overlap.}\label{tab:harness_error_mechanisms}

\begin{tabular}{llrrrrrrr}\toprule

Suite & Harness & $n$ & E & T & R & P & S & V \\\midrule

General & Claude Code & 800 & 70 & 95 & 14 & 14 & 34 & 127 \\

 & Codex & 800 & 71 & 79 & 20 & 5 & 34 & 120 \\

 & OpenCode & 800 & 89 & 73 & 51 & 5 & 35 & 143 \\

 & OpenClaw & 800 & 80 & 90 & 19 & 3 & 32 & 161 \\

 & Hermes & 800 & 100 & 81 & 38 & 8 & 41 & 130 \\

 & Goose & 800 & 78 & 171 & 14 & 6 & 26 & 131 \\

 & Nanobot & 800 & 108 & 105 & 106 & 14 & 197 & 136 \\

\midrule

Multimodal & Claude Code & 84 & 7 & 8 & 3 & 1 & 1 & 36 \\

 & Codex & 84 & 5 & 3 & 0 & 1 & 3 & 31 \\

 & OpenCode & 84 & 13 & 6 & 1 & 2 & 3 & 28 \\

 & OpenClaw & 84 & 11 & 5 & 0 & 1 & 1 & 33 \\

 & Hermes & 84 & 10 & 5 & 9 & 2 & 3 & 25 \\

 & Goose & 84 & 10 & 2 & 1 & 1 & 2 & 30 \\

 & Nanobot & 84 & 10 & 5 & 13 & 1 & 8 & 34 \\

\bottomrule\end{tabular}\end{table}

\subsection{Delivery Quality and Execution Cost}
\label{app:cost_comparisons}

\paragraph{Quality--cost contrasts.}

Table~\ref{tab:cost_pair_estimates} reports paired quality and cost contrasts for DeepSeek Flash and GPT Sol on all 274 general tasks. We use 3,000 task-paired bootstrap resamples (seed 20260925) to obtain percentile 95\% intervals. The DeepSeek Flash--Nanobot cache fraction is total reported cached input divided by total input across the same suite.

\begin{table}[!htbp]\centering\small

\begin{tabular}{lrrr}\toprule

Comparison (first minus second) & Score difference & $R_1$ difference & Cost ratio \\

\midrule

DS Flash: Codex / Nano & +7.41 [+4.06, +10.84] & +4.01 [-0.36, +8.39] & 0.62 [0.56, 0.68] \\

GPT Sol: Nano / OCl & -6.59 [-10.18, -3.03] & -6.20 [-10.95, -1.09] & 2.97 [2.61, 3.38] \\

\bottomrule\end{tabular}

\caption{Paired quality and cost contrasts on all 274 general tasks, with 95\% intervals. Rate differences are percentage points; costs are first/second ratios.}

\label{tab:cost_pair_estimates}\end{table}

Figure~\ref{fig:requests_cost_delivery}(b) uses the same full suite for GPT Luna. Empirical frontier membership is computed separately for mean score, $R_{0.8}$, and $R_1$, with identical task weights and cost estimates in every panel.

\paragraph{Requests and delivery.}

Let $T_i$ be the recorded number of provider transactions, $Q_i$ the final oracle score, $N$ the number of runs, and $b$ a request count on the horizontal axis. We plot $S(b)=N^{-1}\sum_i\mathbf{1}[T_i>b]$ and $A_q(b)=N^{-1}\sum_i\mathbf{1}[T_i\le b,Q_i\ge q]$, for $q=0.80$ and $q=1$. Provider transactions can include auxiliary, delegated, or failed requests; one transaction is not necessarily one tool operation. Curves use final scores at observed termination, not counterfactual scores under a smaller budget. Provider request counts are distinct from interaction-turn limits.

\begin{figure}[!htbp]\centering

\includegraphics[width=\linewidth]{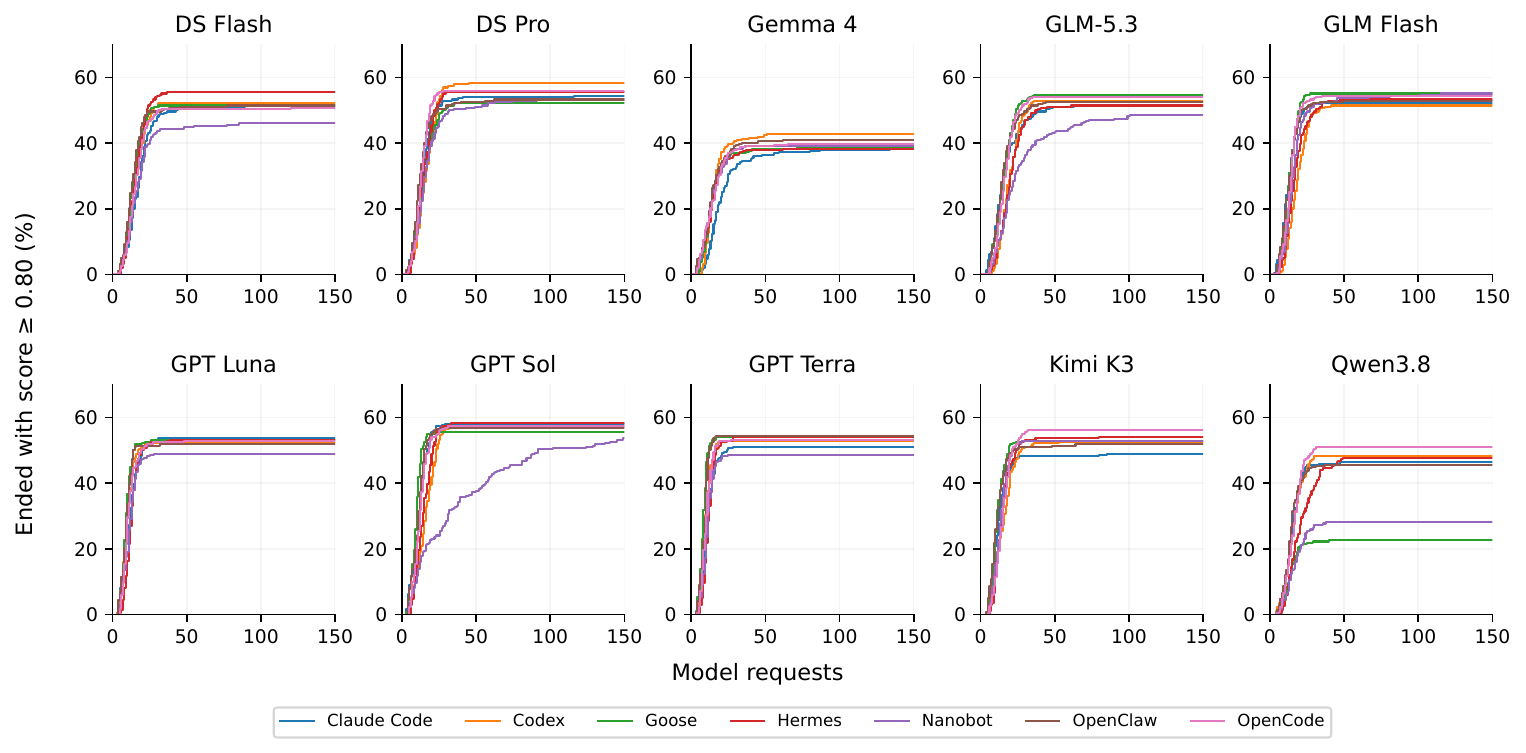}

\caption{Observed termination with score at least 0.80 for all ten models and seven harnesses, on the full general task suite.}\label{fig:all_request_attainment}

\end{figure}

Table~\ref{tab:request_pair_intervals} reports paired quality and request-count differences on all 274 general tasks.

\begin{table}[!htbp]

\centering\footnotesize\setlength{\tabcolsep}{3pt}

\begin{tabular}{lrrrr}\toprule

Comparison & $N$ & Score difference & $R_1$ difference & Request difference \\\midrule

Sol & 274 & +6.59 [+3.20, +10.23] & +6.20 [+1.46, +10.95] & -27.73 [-31.83, -23.65] \\

GLM-5.3 & 274 & +8.53 [+4.61, +12.39] & +7.30 [+2.92, +12.04] & -12.62 [-15.92, -9.63] \\

\bottomrule\end{tabular}

\caption{Sol compares OpenClaw minus Nanobot; GLM-5.3 compares OpenCode minus Nanobot. Paired differences with percentile 95\% task-bootstrap intervals. $R_1$ differences are percentage points. Both comparisons use all 274 general tasks.}\label{tab:request_pair_intervals}

\end{table}

\paragraph{Request distributions.}
Table~\ref{tab:all_request_summaries} complements the per-configuration means and quality metrics in Appendix~\ref{app:detailed_results} with median requests and counts of runs exceeding 40 requests.

\begin{table}[!htbp]
\centering\small
\setlength{\tabcolsep}{3pt}\renewcommand{\arraystretch}{1.15}
\caption{Request distributions on 274 general tasks per configuration. Each cell gives the median request count / number of runs with more than 40 requests. Requests include auxiliary and delegated calls.}
\label{tab:all_request_summaries}
\begin{tabular*}{\linewidth}{@{\extracolsep{\fill}}lccccccc@{}}
\toprule
Model & Claude Code & Codex & OpenCode & OpenClaw & Hermes & Goose & Nanobot \\
\midrule
GPT Sol & 11 / 0 & 15 / 0 & 12 / 0 & 11 / 0 & 14 / 0 & 9 / 0 & 26 / 85 \\
GPT Terra & 10 / 0 & 9 / 0 & 10 / 1 & 9 / 0 & 11 / 0 & 8 / 0 & 10 / 0 \\
GPT Luna & 11 / 0 & 9 / 0 & 10 / 1 & 9 / 0 & 12 / 0 & 9 / 0 & 11 / 0 \\
GLM-5.3 & 15 / 16 & 18 / 5 & 16 / 0 & 15 / 7 & 19 / 8 & 15 / 2 & 20 / 54 \\
GLM Flash & 12 / 2 & 17 / 9 & 14 / 0 & 13 / 4 & 15 / 3 & 12 / 0 & 15 / 18 \\
DS Pro & 14 / 13 & 17 / 3 & 12 / 0 & 11 / 7 & 13 / 2 & 14 / 1 & 15 / 43 \\
DS Flash & 17 / 23 & 15 / 1 & 15 / 3 & 13 / 4 & 15 / 1 & 13 / 1 & 16 / 20 \\
Kimi K3 & 12 / 22 & 14 / 4 & 15 / 0 & 10 / 3 & 14 / 8 & 10 / 0 & 13 / 1 \\
Gemma 4 & 19 / 44 & 14 / 8 & 14 / 7 & 14 / 9 & 14 / 11 & 12 / 19 & 13 / 9 \\
Qwen3.8 & 14 / 5 & 15 / 0 & 15 / 1 & 13 / 3 & 18 / 12 & 5 / 1 & 10 / 3 \\
\bottomrule
\end{tabular*}
\end{table}

\section{Trajectory Annotation}
\label{app:trajectory_annotation}

\paragraph{Sampling.}
The corpus contains 6,864 unique runs covering all 300 tasks. The general base sample contains 80 runs per model--harness configuration: 20 full-credit and 60 non-full-credit runs. We initially allocate 20 places to each band: zero, $(0,0.80)$, and $[0.80,1)$. An undersized band is exhausted, and remaining places go to the non-full-credit band with the most remaining capacity. Within each band, task-type strata receive at least two runs when available; remaining places are allocated proportionally to residual stratum sizes using largest remainders. Selection strata and inclusion probabilities are retained; task coverage is summarized with the current taxonomy.

Multimodal sampling selects 12 runs per configuration. Each nonempty score band receives up to three initial places; remaining places are allocated proportionally to available capacity. Within a band, each represented task subcategory receives at least one run when available, with the remainder allocated proportionally. Fixed-seed hash rankings select runs within strata. The resulting general score-band counts are 1,321 zero, 1,485 positive but below 0.80, 1,394 high-partial, and 1,400 full-credit; the multimodal counts are 123, 312, 21, and 132, respectively.

Matched successful runs add same-task, same-model contrasts under a different harness; diagnostic records ensure task and stratum coverage (Table~\ref{tab:annotation_coverage}). Each record retains its role, stratum, and inclusion probability. The main distribution figure and configuration-level error percentages use only the base samples. Matched controls and diagnostic additions support case analysis and are not pooled into those denominators.

\begin{table}[!htbp]
\centering\small
\begin{tabular}{@{}lrr@{}}
\toprule
Sample component & Selected & Structured records \\
\midrule
General probability sample & 5,600 & 5,600 \\
Multimodal probability sample & 588 & 588 \\
Matched successful runs & 620 & 620 \\
Task and stratum coverage additions & 56 & 56 \\
\midrule
Total & 6,864 & 6,864 \\
\bottomrule
\end{tabular}
\caption{Trajectory annotation coverage by sampling role. Structured records pass format and evidence-citation checks.}
\label{tab:annotation_coverage}
\end{table}

\paragraph{Evidence-linked annotation.}
Agent-assisted inspection of execution trajectories identifies six recurring error categories. GPT-5.6 Luna then classifies sampled trajectories using evidence packets containing task requirements, execution events, tool feedback, and artifact excerpts. Each record states the task requirement and cites at least two specific observations with event locators and verbatim evidence. Positive labels additionally describe the behavior, recovery outcome, relation to delivery, confidence, and a possible harness design response. Multiple labels are allowed when distinct behaviors are supported. Automated checks validate record structure and cited text; a separate targeted semantic review checks whether the cited evidence supports the interpretation. The six categories in Table~\ref{tab:execution_codebook} describe execution mechanisms rather than score levels.

\paragraph{Human validation and adjudication.}
Two human annotators independently reviewed the same 100 trajectories randomly sampled from the base samples. They saw only the execution trajectories, artifacts, and final scores; model-generated error labels were withheld. Before adjudication, complete error-label-set agreement was 97\% between the human annotators and 98\% and 97\% between each annotator and GPT-5.6 Luna, respectively. All three agreed on 96 trajectories. On one trajectory, both humans agreed on a label set different from Luna's; on the remaining three, the humans disagreed and Luna matched one of them. This agreement supports Luna-assisted classification under the annotation protocol. Disagreements were adjudicated and labels corrected. All uncertain model classifications receive manual review and annotation. Other supported errors fall outside the six-category codebook; records without sufficient evidence for a supported classification remain unclassified. 

\begin{table}[!htbp]
\centering\small
\setlength{\tabcolsep}{4pt}
\renewcommand{\arraystretch}{1.12}
\begin{tabular}{@{}>{\raggedright\arraybackslash}p{0.23\linewidth}>{\raggedright\arraybackslash}p{0.47\linewidth}>{\raggedright\arraybackslash}p{\dimexpr0.30\linewidth-16pt\relax}@{}}
\toprule
Mechanism & Evidence required & Design direction \\
\midrule
Evidence and requirements & An applicable rule or acquired fact conflicts with the subsequent decision. & Preserve constraints and decision-relevant facts. \\
\addlinespace[3pt]
Tool invocation & A call uses an incompatible tool, argument, or target, with observable feedback. & Check interfaces and target types before dispatch. \\
\addlinespace[3pt]
Recovery & Actionable feedback is followed by an unchanged invalid operation or a repair that introduces another defect. & Carry the required correction into the next action. \\
\addlinespace[3pt]
Execution stagnation & Repeated settled decisions or investigation produce no corresponding action or new evidence. & Materialize a checkable intermediate result. \\
\addlinespace[3pt]
State and artifacts & Recorded operations and final artifacts reveal missing persistence, inconsistent outputs, or duplicate effects. & Track updates and reconcile final outputs. \\
\addlinespace[3pt]
Verification & A specific check or completion claim conflicts with visible evidence or omits an explicit delivery requirement. & Tie verification to the delivery contract. \\
\bottomrule
\end{tabular}
\caption{Execution mechanisms, the evidence needed to identify them, and the harness design questions they motivate. Categories can overlap within a run.}
\label{tab:execution_codebook}
\end{table}

\paragraph{Recovery and delivery.}
For each observed error, recovery is recorded as complete, partial, absent, or unknown. Its relation to delivery is recorded separately as direct, contributing, not established, or unknown. An invalid call followed by a corrected call can therefore be a recovered tool error without being a recovery failure. Conversely, recognizing a mistake in reasoning does not establish recovery unless subsequent actions address it. A high final score can coexist with earlier corrected errors. The absence of a step from a bounded excerpt does not establish that the step never occurred, and a hidden-oracle failure alone does not establish that the agent ignored visible feedback.

\paragraph{Final annotation corpus.}
The completed corpus contains 6,864 accepted records: 6,235 general and 629 multimodal. Of these, 3,079 carry at least one confirmed error label, 2,318 contain no confirmed error, and 1,467 remain unclassified because the inspected evidence does not support a determination. Counts are reported separately by sampling role and modality. Unconfirmed errors may be absent or unobservable in the inspected evidence; mechanisms outside the codebook are tracked separately.

\begin{table}[!htbp]
\centering\small
\setlength{\tabcolsep}{4pt}
\begin{tabular}{@{}lrrr rrr@{}}
\toprule
 & \multicolumn{3}{c}{General: 5,600 runs} & \multicolumn{3}{c}{Multimodal: 588 runs} \\
\cmidrule(lr){2-4}\cmidrule(l){5-7}
Mechanism & Runs & Labels & Recovered & Runs & Labels & Recovered \\
\midrule
Evidence / requirements & 596 & 614 & 94 & 66 & 69 & 6 \\
Tool invocation & 694 & 729 & 655 & 34 & 37 & 29 \\
Recovery & 262 & 263 & 160 & 27 & 27 & 8 \\
Execution stagnation & 55 & 55 & 2 & 9 & 9 & 0 \\
State / artifacts & 399 & 401 & 26 & 21 & 21 & 0 \\
Verification & 948 & 962 & 7 & 217 & 220 & 0 \\
\bottomrule
\end{tabular}
\caption{Final descriptive counts for the two probability samples. \textit{Runs} reports distinct annotated runs per category; \textit{Labels} reports observations; \textit{Recovered} reports labels followed by complete recovery. Categories can overlap. Matched successful controls and diagnostic additions are excluded. The score-stratified sample counts are not weighted estimates of benchmark-wide prevalence.}
\label{tab:annotation_counts}
\end{table}

\subsection{Error Distributions for Every Model--Harness Configuration}
\label{app:config_errors}
Tables~\ref{tab:error_profiles_general} and~\ref{tab:error_profiles_multimodal} report all 119 configurations. $N_{\rm all}$ counts every annotated run for the configuration, including matched controls and diagnostics; $n$ is its base-sample size and the denominator for all percentages. E: evidence and requirements; T: tool invocation; R: recovery; P: execution stagnation; S: state and artifacts; V: verification; O: other confirmed categories outside the six-category codebook. Each category counts distinct runs, so categories can overlap.

\textit{None} means no error was confirmed in the inspected evidence; \textit{Uncls.} means the inspected evidence is insufficient for a supported classification. Neither is inferred from the oracle score. \textit{Full} is the percentage of base-sample runs receiving full oracle credit, which can include recovered errors. The percentage with any confirmed error is $100-\textit{None}-\textit{Uncls.}$, up to rounding. All percentages describe the score-stratified samples, not weighted population prevalence. Total annotation coverage is 6,235 general and 629 multimodal runs; their base samples contain 5,600 and 588 runs.

\begingroup\small
\setlength{\tabcolsep}{3.5pt}\renewcommand{\arraystretch}{1.12}
\begin{longtable}{@{}llrr*{10}{r}@{}}
\caption{General-task annotation profiles. Counts: $N_{\rm all}$, $n$; all remaining columns: percentages of $n$.}\label{tab:error_profiles_general}\\
\toprule
Model & Harness & $N_{\rm all}$ & $n$ & E & T & R & P & S & V & O & None & Uncls. & Full \\
\midrule
\endfirsthead
\multicolumn{14}{l}{\textit{General-task annotation profiles (continued)}}\\
\toprule
Model & Harness & $N_{\rm all}$ & $n$ & E & T & R & P & S & V & O & None & Uncls. & Full \\
\midrule
\endhead
\midrule
\multicolumn{14}{r}{\textit{Continued on next page}}\\
\endfoot
\bottomrule
\endlastfoot
DS Flash & Claude Code & 85 & 80 & 2.5 & 7.5 & 1.2 & 0.0 & 6.2 & 16.2 & 5.0 & 40.0 & 25.0 & 25.0 \\*
 & Codex & 86 & 80 & 6.2 & 13.8 & 2.5 & 0.0 & 6.2 & 11.2 & 7.5 & 31.2 & 27.5 & 25.0 \\*
 & OpenCode & 90 & 80 & 10.0 & 8.8 & 5.0 & 1.2 & 3.8 & 20.0 & 6.2 & 32.5 & 21.2 & 25.0 \\*
 & OpenClaw & 92 & 80 & 8.8 & 7.5 & 1.2 & 0.0 & 3.8 & 22.5 & 5.0 & 40.0 & 20.0 & 25.0 \\*
 & Hermes & 95 & 80 & 7.5 & 2.5 & 2.5 & 1.2 & 5.0 & 17.5 & 6.2 & 35.0 & 26.2 & 25.0 \\*
 & Goose & 90 & 80 & 10.0 & 28.8 & 3.8 & 0.0 & 3.8 & 20.0 & 2.5 & 31.2 & 15.0 & 25.0 \\*
 & Nanobot & 90 & 80 & 5.0 & 11.2 & 18.8 & 1.2 & 28.8 & 18.8 & 2.5 & 21.2 & 16.2 & 25.0 \\
\addlinespace[3pt]
DS Pro & Claude Code & 88 & 80 & 12.5 & 7.5 & 0.0 & 0.0 & 1.2 & 21.2 & 3.8 & 35.0 & 25.0 & 25.0 \\*
 & Codex & 93 & 80 & 11.2 & 7.5 & 1.2 & 0.0 & 2.5 & 17.5 & 0.0 & 42.5 & 21.2 & 25.0 \\*
 & OpenCode & 97 & 80 & 8.8 & 6.2 & 3.8 & 1.2 & 3.8 & 17.5 & 0.0 & 41.2 & 22.5 & 25.0 \\*
 & OpenClaw & 90 & 80 & 8.8 & 5.0 & 0.0 & 0.0 & 2.5 & 20.0 & 0.0 & 35.0 & 31.2 & 25.0 \\*
 & Hermes & 92 & 80 & 8.8 & 5.0 & 2.5 & 1.2 & 2.5 & 13.8 & 1.2 & 40.0 & 26.2 & 25.0 \\*
 & Goose & 88 & 80 & 11.2 & 41.2 & 2.5 & 1.2 & 0.0 & 17.5 & 1.2 & 20.0 & 16.2 & 25.0 \\*
 & Nanobot & 89 & 80 & 7.5 & 8.8 & 10.0 & 3.8 & 28.8 & 18.8 & 1.2 & 23.8 & 17.5 & 25.0 \\
\addlinespace[3pt]
Gemma 4 & Claude Code & 93 & 80 & 11.2 & 15.0 & 10.0 & 5.0 & 6.2 & 20.0 & 3.8 & 28.8 & 15.0 & 25.0 \\*
 & Codex & 91 & 80 & 15.0 & 8.8 & 10.0 & 1.2 & 6.2 & 22.5 & 1.2 & 32.5 & 18.8 & 25.0 \\*
 & OpenCode & 89 & 80 & 17.5 & 26.2 & 13.8 & 1.2 & 6.2 & 16.2 & 1.2 & 18.8 & 12.5 & 25.0 \\*
 & OpenClaw & 92 & 80 & 11.2 & 21.2 & 11.2 & 2.5 & 3.8 & 23.8 & 3.8 & 25.0 & 13.8 & 25.0 \\*
 & Hermes & 88 & 80 & 23.8 & 22.5 & 17.5 & 1.2 & 10.0 & 20.0 & 2.5 & 18.8 & 10.0 & 25.0 \\*
 & Goose & 92 & 80 & 13.8 & 11.2 & 6.2 & 1.2 & 8.8 & 17.5 & 1.2 & 32.5 & 17.5 & 25.0 \\*
 & Nanobot & 93 & 80 & 17.5 & 17.5 & 11.2 & 1.2 & 25.0 & 15.0 & 2.5 & 17.5 & 21.2 & 25.0 \\
\addlinespace[3pt]
GLM-5.3 & Claude Code & 87 & 80 & 6.2 & 11.2 & 1.2 & 0.0 & 2.5 & 25.0 & 1.2 & 31.2 & 28.8 & 25.0 \\*
 & Codex & 86 & 80 & 6.2 & 25.0 & 1.2 & 1.2 & 1.2 & 13.8 & 2.5 & 32.5 & 22.5 & 25.0 \\*
 & OpenCode & 88 & 80 & 12.5 & 1.2 & 11.2 & 0.0 & 2.5 & 25.0 & 0.0 & 32.5 & 23.8 & 25.0 \\*
 & OpenClaw & 88 & 80 & 13.8 & 13.8 & 1.2 & 0.0 & 2.5 & 23.8 & 3.8 & 31.2 & 22.5 & 25.0 \\*
 & Hermes & 90 & 80 & 12.5 & 11.2 & 3.8 & 1.2 & 2.5 & 23.8 & 0.0 & 30.0 & 23.8 & 25.0 \\*
 & Goose & 91 & 80 & 7.5 & 15.0 & 0.0 & 0.0 & 1.2 & 15.0 & 5.0 & 35.0 & 28.8 & 25.0 \\*
 & Nanobot & 87 & 80 & 18.8 & 12.5 & 16.2 & 1.2 & 17.5 & 25.0 & 0.0 & 16.2 & 16.2 & 25.0 \\
\addlinespace[3pt]
GLM Flash & Claude Code & 90 & 80 & 10.0 & 7.5 & 0.0 & 2.5 & 6.2 & 15.0 & 6.2 & 36.2 & 22.5 & 25.0 \\*
 & Codex & 86 & 80 & 10.0 & 11.2 & 3.8 & 0.0 & 10.0 & 21.2 & 7.5 & 30.0 & 18.8 & 25.0 \\*
 & OpenCode & 93 & 80 & 13.8 & 7.5 & 6.2 & 0.0 & 10.0 & 25.0 & 3.8 & 36.2 & 10.0 & 25.0 \\*
 & OpenClaw & 90 & 80 & 12.5 & 10.0 & 0.0 & 0.0 & 3.8 & 26.2 & 2.5 & 35.0 & 20.0 & 25.0 \\*
 & Hermes & 90 & 80 & 8.8 & 22.5 & 2.5 & 1.2 & 3.8 & 20.0 & 3.8 & 23.8 & 23.8 & 25.0 \\*
 & Goose & 92 & 80 & 7.5 & 20.0 & 1.2 & 1.2 & 0.0 & 17.5 & 3.8 & 33.8 & 20.0 & 25.0 \\*
 & Nanobot & 88 & 80 & 16.2 & 10.0 & 15.0 & 0.0 & 17.5 & 20.0 & 1.2 & 20.0 & 12.5 & 25.0 \\
\addlinespace[3pt]
GPT Luna & Claude Code & 85 & 80 & 10.0 & 17.5 & 0.0 & 0.0 & 3.8 & 10.0 & 2.5 & 33.8 & 27.5 & 25.0 \\*
 & Codex & 85 & 80 & 8.8 & 0.0 & 0.0 & 0.0 & 1.2 & 10.0 & 0.0 & 42.5 & 40.0 & 25.0 \\*
 & OpenCode & 86 & 80 & 12.5 & 15.0 & 1.2 & 0.0 & 5.0 & 13.8 & 2.5 & 30.0 & 27.5 & 25.0 \\*
 & OpenClaw & 88 & 80 & 8.8 & 18.8 & 2.5 & 0.0 & 7.5 & 17.5 & 5.0 & 28.8 & 25.0 & 25.0 \\*
 & Hermes & 86 & 80 & 16.2 & 10.0 & 3.8 & 1.2 & 10.0 & 10.0 & 2.5 & 33.8 & 22.5 & 25.0 \\*
 & Goose & 83 & 80 & 12.5 & 16.2 & 1.2 & 0.0 & 6.2 & 15.0 & 2.5 & 32.5 & 23.8 & 25.0 \\*
 & Nanobot & 87 & 80 & 13.8 & 23.8 & 3.8 & 1.2 & 23.8 & 17.5 & 0.0 & 15.0 & 21.2 & 25.0 \\
\addlinespace[3pt]
GPT Sol & Claude Code & 87 & 80 & 6.2 & 16.2 & 1.2 & 0.0 & 1.2 & 15.0 & 1.2 & 33.8 & 28.8 & 25.0 \\*
 & Codex & 84 & 80 & 6.2 & 8.8 & 1.2 & 0.0 & 1.2 & 16.2 & 0.0 & 37.5 & 32.5 & 25.0 \\*
 & OpenCode & 93 & 80 & 7.5 & 5.0 & 3.8 & 0.0 & 1.2 & 12.5 & 0.0 & 37.5 & 36.2 & 25.0 \\*
 & OpenClaw & 91 & 80 & 10.0 & 10.0 & 1.2 & 0.0 & 2.5 & 17.5 & 2.5 & 27.5 & 31.2 & 25.0 \\*
 & Hermes & 95 & 80 & 8.8 & 7.5 & 2.5 & 0.0 & 1.2 & 13.8 & 0.0 & 35.0 & 35.0 & 25.0 \\*
 & Goose & 89 & 80 & 5.0 & 16.2 & 1.2 & 0.0 & 0.0 & 17.5 & 5.0 & 37.5 & 23.8 & 25.0 \\*
 & Nanobot & 87 & 80 & 10.0 & 12.5 & 16.2 & 0.0 & 26.2 & 11.2 & 0.0 & 23.8 & 25.0 & 25.0 \\
\addlinespace[3pt]
GPT Terra & Claude Code & 85 & 80 & 7.5 & 20.0 & 2.5 & 0.0 & 6.2 & 15.0 & 8.8 & 27.5 & 23.8 & 25.0 \\*
 & Codex & 84 & 80 & 5.0 & 0.0 & 0.0 & 0.0 & 2.5 & 12.5 & 0.0 & 41.2 & 40.0 & 25.0 \\*
 & OpenCode & 89 & 80 & 7.5 & 11.2 & 5.0 & 0.0 & 2.5 & 13.8 & 0.0 & 35.0 & 30.0 & 25.0 \\*
 & OpenClaw & 91 & 80 & 11.2 & 13.8 & 0.0 & 0.0 & 2.5 & 15.0 & 5.0 & 33.8 & 25.0 & 25.0 \\*
 & Hermes & 89 & 80 & 15.0 & 6.2 & 2.5 & 1.2 & 6.2 & 10.0 & 1.2 & 38.8 & 25.0 & 25.0 \\*
 & Goose & 90 & 80 & 8.8 & 7.5 & 1.2 & 0.0 & 6.2 & 15.0 & 1.2 & 35.0 & 26.2 & 25.0 \\*
 & Nanobot & 85 & 80 & 15.0 & 18.8 & 7.5 & 1.2 & 21.2 & 10.0 & 2.5 & 23.8 & 16.2 & 25.0 \\
\addlinespace[3pt]
Kimi K3 & Claude Code & 83 & 80 & 8.8 & 6.2 & 0.0 & 8.8 & 7.5 & 8.8 & 3.8 & 35.0 & 23.8 & 25.0 \\*
 & Codex & 88 & 80 & 8.8 & 15.0 & 1.2 & 2.5 & 3.8 & 15.0 & 1.2 & 28.8 & 30.0 & 25.0 \\*
 & OpenCode & 90 & 80 & 10.0 & 5.0 & 7.5 & 0.0 & 5.0 & 18.8 & 1.2 & 32.5 & 26.2 & 25.0 \\*
 & OpenClaw & 91 & 80 & 11.2 & 5.0 & 0.0 & 0.0 & 3.8 & 20.0 & 2.5 & 35.0 & 25.0 & 25.0 \\*
 & Hermes & 89 & 80 & 13.8 & 7.5 & 6.2 & 1.2 & 5.0 & 22.5 & 0.0 & 33.8 & 23.8 & 25.0 \\*
 & Goose & 87 & 80 & 13.8 & 7.5 & 0.0 & 0.0 & 2.5 & 17.5 & 3.8 & 41.2 & 22.5 & 25.0 \\*
 & Nanobot & 90 & 80 & 17.5 & 7.5 & 15.0 & 3.8 & 28.8 & 15.0 & 0.0 & 21.2 & 15.0 & 25.0 \\
\addlinespace[3pt]
Qwen3.8 & Claude Code & 95 & 80 & 12.5 & 10.0 & 1.2 & 1.2 & 1.2 & 12.5 & 6.2 & 32.5 & 27.5 & 25.0 \\*
 & Codex & 93 & 80 & 11.2 & 8.8 & 3.8 & 1.2 & 7.5 & 10.0 & 2.5 & 32.5 & 32.5 & 25.0 \\*
 & OpenCode & 94 & 80 & 11.2 & 5.0 & 6.2 & 2.5 & 3.8 & 16.2 & 3.8 & 33.8 & 23.8 & 25.0 \\*
 & OpenClaw & 90 & 80 & 3.8 & 7.5 & 6.2 & 1.2 & 7.5 & 15.0 & 1.2 & 28.8 & 32.5 & 25.0 \\*
 & Hermes & 92 & 80 & 10.0 & 6.2 & 3.8 & 0.0 & 5.0 & 11.2 & 2.5 & 36.2 & 28.8 & 25.0 \\*
 & Goose & 84 & 80 & 7.5 & 50.0 & 0.0 & 3.8 & 3.8 & 11.2 & 2.5 & 18.8 & 11.2 & 25.0 \\*
 & Nanobot & 81 & 80 & 13.8 & 8.8 & 18.8 & 3.8 & 28.8 & 18.8 & 1.2 & 20.0 & 17.5 & 25.0 \\
\addlinespace[3pt]
\end{longtable}\endgroup

\begingroup\small
\setlength{\tabcolsep}{3.5pt}\renewcommand{\arraystretch}{1.12}
\begin{longtable}{@{}llrr*{10}{r}@{}}
\caption{Multimodal-task annotation profiles. Counts: $N_{\rm all}$, $n$; all remaining columns: percentages of $n$.}\label{tab:error_profiles_multimodal}\\
\toprule
Model & Harness & $N_{\rm all}$ & $n$ & E & T & R & P & S & V & O & None & Uncls. & Full \\
\midrule
\endfirsthead
\multicolumn{14}{l}{\textit{Multimodal-task annotation profiles (continued)}}\\
\toprule
Model & Harness & $N_{\rm all}$ & $n$ & E & T & R & P & S & V & O & None & Uncls. & Full \\
\midrule
\endhead
\midrule
\multicolumn{14}{r}{\textit{Continued on next page}}\\
\endfoot
\bottomrule
\endlastfoot
Gemma 4 & Claude Code & 12 & 12 & 0.0 & 33.3 & 25.0 & 0.0 & 0.0 & 83.3 & 16.7 & 0.0 & 8.3 & 0.0 \\*
 & Codex & 12 & 12 & 8.3 & 8.3 & 0.0 & 0.0 & 16.7 & 25.0 & 8.3 & 8.3 & 33.3 & 0.0 \\*
 & OpenCode & 12 & 12 & 25.0 & 25.0 & 0.0 & 0.0 & 8.3 & 50.0 & 0.0 & 0.0 & 16.7 & 0.0 \\*
 & OpenClaw & 12 & 12 & 25.0 & 8.3 & 0.0 & 0.0 & 8.3 & 58.3 & 16.7 & 8.3 & 16.7 & 0.0 \\*
 & Hermes & 12 & 12 & 8.3 & 16.7 & 8.3 & 0.0 & 16.7 & 41.7 & 8.3 & 8.3 & 16.7 & 0.0 \\*
 & Goose & 12 & 12 & 33.3 & 8.3 & 0.0 & 0.0 & 16.7 & 58.3 & 0.0 & 0.0 & 16.7 & 0.0 \\*
 & Nanobot & 12 & 12 & 8.3 & 16.7 & 8.3 & 0.0 & 8.3 & 75.0 & 8.3 & 0.0 & 16.7 & 0.0 \\
\addlinespace[3pt]
GLM Flash & Claude Code & 14 & 12 & 16.7 & 0.0 & 0.0 & 0.0 & 0.0 & 33.3 & 0.0 & 16.7 & 33.3 & 25.0 \\*
 & Codex & 12 & 12 & 8.3 & 16.7 & 0.0 & 0.0 & 0.0 & 16.7 & 16.7 & 41.7 & 16.7 & 33.3 \\*
 & OpenCode & 13 & 12 & 16.7 & 8.3 & 8.3 & 0.0 & 0.0 & 25.0 & 0.0 & 25.0 & 16.7 & 33.3 \\*
 & OpenClaw & 16 & 12 & 0.0 & 0.0 & 0.0 & 0.0 & 0.0 & 33.3 & 16.7 & 33.3 & 16.7 & 50.0 \\*
 & Hermes & 14 & 12 & 8.3 & 0.0 & 0.0 & 8.3 & 0.0 & 16.7 & 0.0 & 25.0 & 41.7 & 41.7 \\*
 & Goose & 15 & 12 & 8.3 & 0.0 & 0.0 & 8.3 & 0.0 & 41.7 & 8.3 & 16.7 & 16.7 & 33.3 \\*
 & Nanobot & 13 & 12 & 25.0 & 8.3 & 25.0 & 0.0 & 0.0 & 8.3 & 8.3 & 25.0 & 16.7 & 33.3 \\
\addlinespace[3pt]
GPT Luna & Claude Code & 14 & 12 & 16.7 & 0.0 & 0.0 & 0.0 & 0.0 & 50.0 & 0.0 & 25.0 & 8.3 & 25.0 \\*
 & Codex & 12 & 12 & 0.0 & 0.0 & 0.0 & 0.0 & 0.0 & 50.0 & 0.0 & 25.0 & 25.0 & 16.7 \\*
 & OpenCode & 12 & 12 & 25.0 & 0.0 & 0.0 & 0.0 & 8.3 & 41.7 & 0.0 & 8.3 & 33.3 & 8.3 \\*
 & OpenClaw & 13 & 12 & 16.7 & 0.0 & 0.0 & 0.0 & 0.0 & 25.0 & 0.0 & 41.7 & 16.7 & 33.3 \\*
 & Hermes & 12 & 12 & 16.7 & 8.3 & 25.0 & 0.0 & 0.0 & 16.7 & 0.0 & 16.7 & 25.0 & 16.7 \\*
 & Goose & 13 & 12 & 8.3 & 0.0 & 8.3 & 0.0 & 0.0 & 25.0 & 0.0 & 16.7 & 41.7 & 25.0 \\*
 & Nanobot & 12 & 12 & 0.0 & 8.3 & 0.0 & 0.0 & 0.0 & 41.7 & 0.0 & 41.7 & 8.3 & 25.0 \\
\addlinespace[3pt]
GPT Sol & Claude Code & 12 & 12 & 0.0 & 8.3 & 0.0 & 0.0 & 0.0 & 50.0 & 8.3 & 16.7 & 25.0 & 25.0 \\*
 & Codex & 14 & 12 & 0.0 & 0.0 & 0.0 & 0.0 & 8.3 & 50.0 & 0.0 & 16.7 & 33.3 & 25.0 \\*
 & OpenCode & 12 & 12 & 8.3 & 0.0 & 0.0 & 0.0 & 0.0 & 33.3 & 16.7 & 16.7 & 25.0 & 25.0 \\*
 & OpenClaw & 12 & 12 & 8.3 & 8.3 & 0.0 & 0.0 & 0.0 & 58.3 & 0.0 & 16.7 & 8.3 & 25.0 \\*
 & Hermes & 13 & 12 & 8.3 & 0.0 & 8.3 & 0.0 & 0.0 & 25.0 & 0.0 & 16.7 & 41.7 & 25.0 \\*
 & Goose & 12 & 12 & 0.0 & 0.0 & 0.0 & 0.0 & 0.0 & 16.7 & 0.0 & 16.7 & 66.7 & 8.3 \\*
 & Nanobot & 12 & 12 & 16.7 & 0.0 & 8.3 & 0.0 & 8.3 & 33.3 & 0.0 & 0.0 & 33.3 & 0.0 \\
\addlinespace[3pt]
GPT Terra & Claude Code & 12 & 12 & 8.3 & 0.0 & 0.0 & 0.0 & 0.0 & 33.3 & 0.0 & 16.7 & 50.0 & 8.3 \\*
 & Codex & 12 & 12 & 0.0 & 0.0 & 0.0 & 0.0 & 0.0 & 75.0 & 0.0 & 25.0 & 0.0 & 25.0 \\*
 & OpenCode & 13 & 12 & 16.7 & 0.0 & 0.0 & 0.0 & 0.0 & 33.3 & 8.3 & 41.7 & 8.3 & 33.3 \\*
 & OpenClaw & 14 & 12 & 8.3 & 8.3 & 0.0 & 0.0 & 0.0 & 25.0 & 8.3 & 50.0 & 0.0 & 33.3 \\*
 & Hermes & 12 & 12 & 8.3 & 0.0 & 25.0 & 0.0 & 0.0 & 33.3 & 8.3 & 8.3 & 41.7 & 16.7 \\*
 & Goose & 14 & 12 & 0.0 & 0.0 & 0.0 & 0.0 & 0.0 & 33.3 & 8.3 & 25.0 & 41.7 & 25.0 \\*
 & Nanobot & 13 & 12 & 25.0 & 0.0 & 8.3 & 0.0 & 8.3 & 25.0 & 0.0 & 25.0 & 16.7 & 33.3 \\
\addlinespace[3pt]
Kimi K3 & Claude Code & 13 & 12 & 16.7 & 16.7 & 0.0 & 0.0 & 8.3 & 16.7 & 0.0 & 41.7 & 8.3 & 33.3 \\*
 & Codex & 12 & 12 & 16.7 & 0.0 & 0.0 & 0.0 & 0.0 & 8.3 & 8.3 & 8.3 & 58.3 & 0.0 \\*
 & OpenCode & 13 & 12 & 8.3 & 0.0 & 0.0 & 8.3 & 0.0 & 33.3 & 0.0 & 25.0 & 25.0 & 33.3 \\*
 & OpenClaw & 12 & 12 & 25.0 & 8.3 & 0.0 & 0.0 & 0.0 & 50.0 & 0.0 & 25.0 & 0.0 & 25.0 \\*
 & Hermes & 13 & 12 & 25.0 & 8.3 & 8.3 & 0.0 & 8.3 & 41.7 & 0.0 & 25.0 & 8.3 & 33.3 \\*
 & Goose & 14 & 12 & 25.0 & 8.3 & 0.0 & 0.0 & 0.0 & 41.7 & 0.0 & 25.0 & 8.3 & 33.3 \\*
 & Nanobot & 14 & 12 & 8.3 & 0.0 & 33.3 & 8.3 & 16.7 & 33.3 & 0.0 & 16.7 & 8.3 & 33.3 \\
\addlinespace[3pt]
Qwen3.8 & Claude Code & 13 & 12 & 0.0 & 8.3 & 0.0 & 8.3 & 0.0 & 33.3 & 8.3 & 16.7 & 33.3 & 33.3 \\*
 & Codex & 14 & 12 & 8.3 & 0.0 & 0.0 & 8.3 & 0.0 & 33.3 & 0.0 & 25.0 & 25.0 & 33.3 \\*
 & OpenCode & 12 & 12 & 8.3 & 16.7 & 0.0 & 8.3 & 8.3 & 16.7 & 0.0 & 16.7 & 33.3 & 33.3 \\*
 & OpenClaw & 14 & 12 & 8.3 & 8.3 & 0.0 & 8.3 & 0.0 & 25.0 & 0.0 & 25.0 & 25.0 & 33.3 \\*
 & Hermes & 14 & 12 & 8.3 & 8.3 & 0.0 & 8.3 & 0.0 & 33.3 & 0.0 & 25.0 & 16.7 & 33.3 \\*
 & Goose & 13 & 12 & 8.3 & 0.0 & 0.0 & 0.0 & 0.0 & 33.3 & 8.3 & 33.3 & 16.7 & 25.0 \\*
 & Nanobot & 12 & 12 & 0.0 & 8.3 & 25.0 & 0.0 & 25.0 & 66.7 & 0.0 & 8.3 & 8.3 & 8.3 \\
\addlinespace[3pt]
\end{longtable}\endgroup

\section{Execution Case Studies}
\label{app:trajectory_cases}

The cases below follow three execution transitions: feedback into corrected actions, retained evidence into continued work, and task requirements into final artifacts. Same-task, same-model contrasts are complemented by a successful recovery and cross-model tool-use profiles. Request indices refer to provider traces; OpenCode may include a separate title-generation request. Case scores use the original 0--1 scale.

\subsection{Tool Feedback and Recovery}
\label{app:case_feedback}

\paragraph{Model-specific tool use.}
\label{app:tool_profiles}
Table~\ref{tab:tool_profiles} separates incompatible tool selection from its eventual recovery. The same Goose tool surface is associated with different observed behaviors across models. For example, DeepSeek Pro uses an image reader on the text contract in task 081, then recovers through a shell read. DeepSeek Flash attempts unavailable text-reading calls in task 222, receives the available-tool list, and switches to shell reads. Qwen also recovers image-reader errors in task 041 by using shell reads. These positive corrections coexist with the persistent erroneous calls illustrated in Table~\ref{tab:semantic_cases}. A separate Qwen interface audit finds 273 image-reader calls targeting common text-file extensions across 153 tasks. Interface friction and failure to recover are distinct observations; neither is inferred solely from the final score.

\begin{table}[!htbp]\centering\small
\setlength{\tabcolsep}{4pt}
\begin{tabular}{@{}lrrrr@{}}\toprule
 & \multicolumn{3}{c}{Goose} & OpenCode \\
\cmidrule(lr){2-4}\cmidrule(l){5-5}
Model & Wrong type & All tool errors & Recovered & Wrong type \\
\midrule
DS Flash & 21 & 25 & 24 & 0 \\
DS Pro & 28 & 33 & 31 & 0 \\
Gemma 4 & 0 & 9 & 8 & 0 \\
GLM-5.3 & 5 & 13 & 12 & 0 \\
GLM Flash & 13 & 16 & 15 & 0 \\
GPT Luna & 0 & 13 & 12 & 2 \\
GPT Sol & 1 & 13 & 11 & 0 \\
GPT Terra & 0 & 6 & 6 & 1 \\
Kimi K3 & 2 & 6 & 5 & 0 \\
Qwen3.8 & 35 & 41 & 40 & 0 \\
\bottomrule\end{tabular}
\caption{Tool-use labels in the general probability sample, with 80 runs per model--harness configuration. Wrong type is the codebook's wrong-tool-type subtype; Recovered counts recovery among all tool-error labels. The sampled tasks and score strata differ across configurations, so these are descriptive profiles rather than paired estimates of population incidence.}
\label{tab:tool_profiles}
\end{table}

\paragraph{From reasoning to execution.}
Table~\ref{tab:semantic_cases} compares Qwen executions on the same tasks. The cases were selected from attempts where Goose received zero credit and OpenCode received at least 0.80, followed by inspection of the task instructions, model responses, tool feedback, and final artifacts.

\begingroup\footnotesize
\setlength{\tabcolsep}{4pt}
\renewcommand{\arraystretch}{1.15}
\begin{longtable}{@{}>{\raggedright\arraybackslash}p{0.16\linewidth}*{3}{>{\raggedright\arraybackslash}p{\dimexpr0.28\linewidth-8pt\relax}}@{}}
\caption{Paired semantic and action-level evidence. The observations identify concrete differences in the recorded executions; scores in the first column are the official results. Full request and artifact evidence is preserved in the analysis logs.}\label{tab:semantic_cases}\\
\toprule
Task and scores & Goose trajectory & OpenCode trajectory & Observed distinction \\
\midrule
\endfirsthead
\multicolumn{4}{l}{\textit{Paired trajectory cases (continued)}}\\
\toprule
Task and scores & Goose trajectory & OpenCode trajectory & Observed distinction \\
\midrule
\endhead
\midrule
\multicolumn{4}{r}{\textit{Continued on next page}}\\
\endfoot
\bottomrule
\endlastfoot
137: Calibration-certificate recovery\newline Goose: 0.00\newline OpenCode: 0.92
& Request 10 identifies the wrong image tool and incomplete filename, but repeats both in its next calls. The run contains 17 explicit tool-error results. It eventually reads the controller and queries the database, but never writes the plan; six new artifacts are absent.
& Reads the controller at request 3 and writes the plan at request 8. Preview, apply, verify, staging, and replay follow. The primary outcome checks pass.
& A correction expressed in reasoning does not necessarily change the emitted operation. \\
\addlinespace[6pt]
105: Data-management package staging\newline Goose: 0.00\newline OpenCode: 0.94
& Identifies the main consent restrictions and repository routes, then repeatedly discusses reference syntax, repository alternatives, and future handoff details. No plan is written; six deliverables are absent.
& Writes the plan at request 6, builds at request 7, writes the handoff at request 9, and runs validation, staging, and repeated verification at request 10. The primary outcome checks pass.
& Settled business decisions need to become intermediate artifacts that execution can check. \\
\addlinespace[6pt]
125: Product-recall recovery\newline Goose: 0.00\newline OpenCode: 0.92
& Correctly identifies much of the scope and event logic, but continues reasoning about whether replay remains valid after state mutation. No plan is written; four new deliverables are absent.
& Writes the plan at request 6, corrects evidence references at request 7, previews and applies it, then reorders its lists and applies it again at request 10. The primary outcome checks pass.
& Actual replay provides evidence about stateful behavior that reasoning alone cannot supply. \\
\addlinespace[6pt]
206: Engineering-change propagation\newline Goose: 0.00\newline OpenCode: 1.00
& Begins a detailed recovery analysis before querying the decisive database records. After receiving them, it derives the plan again. A write call starts but is incomplete; six new artifacts are absent.
& Queries the database at request 5, derives and writes the plan, corrects revision-field types, and completes preview, apply, verify, staging, and replay.
& Acquiring decision-relevant state before expanding a plan reduces avoidable re-derivation. \\
\end{longtable}\endgroup

\paragraph{A completed recovery.}
\label{app:recovered_normalization}
Task 051 requires an offline normalization pipeline for customer CSV, order JSON, and note TXT inputs. The deliverables are normalized records, an index, a rejection ledger, and a batch summary. The task also requires deterministic ordering, globally unique accepted IDs, and an idempotent rerun. In the Gemma--Claude Code execution, the agent inspects its outputs, corrects missing note coverage, and obtains full credit. Table~\ref{tab:normalization_recovery} follows the observable sequence. The evidence packet abbreviates the implementation writes, so the case identifies the coverage defect and its correction without assigning an unseen parser-level cause.

\begin{table}[!htbp]
\centering\small
\setlength{\tabcolsep}{4pt}
\renewcommand{\arraystretch}{1.12}
\begin{tabular}{@{}>{\raggedright\arraybackslash}p{0.15\linewidth}>{\raggedright\arraybackslash}p{0.49\linewidth}>{\raggedright\arraybackslash}p{\dimexpr0.36\linewidth-16pt\relax}@{}}
\toprule
Requests & Observed execution & What changes \\
\midrule
4--5 & Reads generated files and ledgers. The summary reports 6 accepted records, 19 rejections, and 0 notes. & Output inspection exposes incomplete input coverage. \\
\addlinespace[3pt]
5--7 & Reads the note files, including their Case, Owner, and Status fields. & Acquires the concrete inputs needed for revision. \\
\addlinespace[3pt]
7--9 & Rewrites and executes the pipeline, then checks the outputs again: 9 accepted records, 9 rejections, and 3 notes. & The revision changes the delivered records and ledgers. \\
\addlinespace[3pt]
9--10 & Reruns the transformation and compares hashes of the output files; the command completes with no reported differences. & Checks repeatability of the revised outputs. \\
\bottomrule
\end{tabular}
\caption{A recovered coverage defect in a full-credit execution. Request indices follow the provider trace; the final oracle independently evaluates the artifacts.}
\label{tab:normalization_recovery}
\end{table}

\begin{figure}[H]\centering
\includegraphics[width=\linewidth]{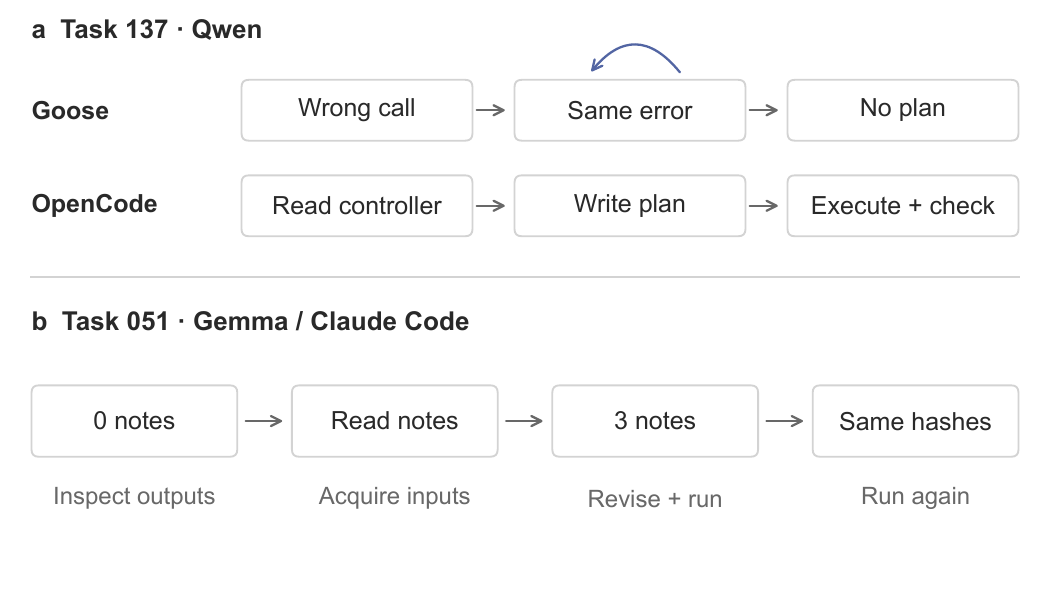}
\caption{Feedback and recovery in two distinct tasks. (a) Qwen--Goose repeats an invalid operation after identifying the mistake in task 137, while OpenCode writes and executes a plan. Goose later reads the controller but leaves the plan unwritten. (b) Gemma--Claude Code detects missing notes in task 051, revises the pipeline, and verifies the rerun.}
\label{fig:normalization_recovery}
\end{figure}

Unlike the repeated invalid calls in Table~\ref{tab:semantic_cases}, the observed discrepancy here leads to new source inspection, a changed implementation, and a check of the revised outputs. Figure~\ref{fig:normalization_recovery} juxtaposes these distinct responses to feedback.

\subsection{Evidence Retention and Context Recovery}
\label{app:case_context}

\paragraph{Retaining evidence needed for implementation.}
\label{app:evidence_retention}
In task 142 (trace-context propagation repair), the DeepSeek Flash--Nanobot run first writes implementation code at provider request 46, after repeated small-range reads. At one observed context transition, retained assistant reasoning grows from 65,760 to 142,765 characters while tool-result content decreases from 17,185 to 3,792. The inspected compaction implementation can replace recent tool results under context pressure. These counts describe retained text, not tokenizer-based token usage. The corresponding task uses 67 provider requests, compared with Codex's 12 and Goose's 14.

\begin{figure}[H]\centering
\includegraphics[width=\linewidth]{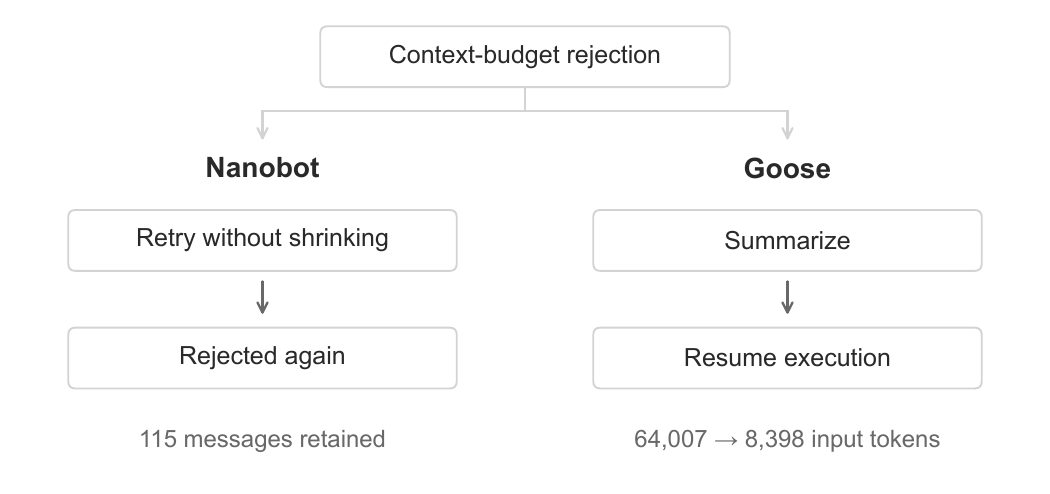}
\caption{Qwen executions on multimodal task 026 encounter context-budget rejection. Nanobot retries without materially reducing the request, whereas Goose summarizes and resumes. Input-token counts compare Goose's last successful request before rejection with its resumed request; the summary is a separate call.}
\label{fig:context_recovery_case}
\end{figure}

\paragraph{Resuming after a context-budget rejection.}
\label{app:multimodal_context_recovery}
The print-preview task (multimodal 026) starts without Node.js installed; preparing the required build dependencies is part of task execution. Paired Qwen--Nanobot and Qwen--Goose runs both encounter a context-budget rejection. Figure~\ref{fig:context_recovery_case} compares their recovery paths. Nanobot's two rejected requests retain 115 messages and the same output reservation, changing image placeholder text without materially reducing the request. Goose instead makes a separate summarization call and resumes execution with a summary and the original task; the recorded input decreases from 64,007 tokens in the preceding successful request to 8,398 in the resumed request. Earlier, Nanobot recovers compacted tool evidence through smaller reads. Neither run establishes a usable Node toolchain; Goose resumes execution but writes the distribution artifact by hand without a standard build.

\subsection{Requirements and Delivered Artifacts}
\label{app:case_delivery}

\paragraph{Policy decisions and supporting clauses.}
\label{app:policy_delivery}
Task 022 explicitly requires every materially applicable secondary clause, alongside the controlling clause. In case T104, the missing itemized receipt triggers exception EX-1 as well as the lodging rule. We inspect the final JSON artifacts of all 70 configurations and condition the comparison on passing the primary-rulings module. Of the 64 qualifying runs, 21 omit EX-1 from both the controlling and secondary references. Table~\ref{tab:policy_exception} gives the model-specific counts. The count tracks the explicitly required EX-1 reference in the final artifacts.

\begin{table}[!htbp]\centering\small
\begin{tabular}{@{}lcc@{}}\toprule
Model & Primary-rulings pass & EX-1 retained among these \\
\midrule
DS Flash & 7 / 7 & 1 / 7 \\
DS Pro & 7 / 7 & 3 / 7 \\
Gemma 4 & 5 / 7 & 0 / 5 \\
GLM-5.3 & 7 / 7 & 7 / 7 \\
GLM Flash & 7 / 7 & 4 / 7 \\
GPT Luna & 7 / 7 & 7 / 7 \\
GPT Sol & 7 / 7 & 7 / 7 \\
GPT Terra & 5 / 7 & 4 / 5 \\
Kimi K3 & 7 / 7 & 7 / 7 \\
Qwen3.8 & 5 / 7 & 3 / 5 \\
\bottomrule\end{tabular}
\caption{The same explicit exception requirement across models and harnesses. Counts refer to the recorded final artifacts; all seven harnesses are evaluated per model.}
\label{tab:policy_exception}
\end{table}

For DeepSeek Pro, both Codex and OpenCode pass the primary rulings, blocking-information, and output-package modules. OpenCode fails the cited-reasoning module on secondary-clause references. The omitted EX-1 reference in T104 is one concrete difference between their supporting reference sets. Codex's request 7 marks its plan complete after schema checks, but requests 8--11 revisit T106 and T109 and reconcile the JSON and CSV. The final outputs agree with OpenCode on all 11 case-level decisions, controlling clauses, and reason codes.

\begin{figure}[H]\centering
\includegraphics[width=\linewidth]{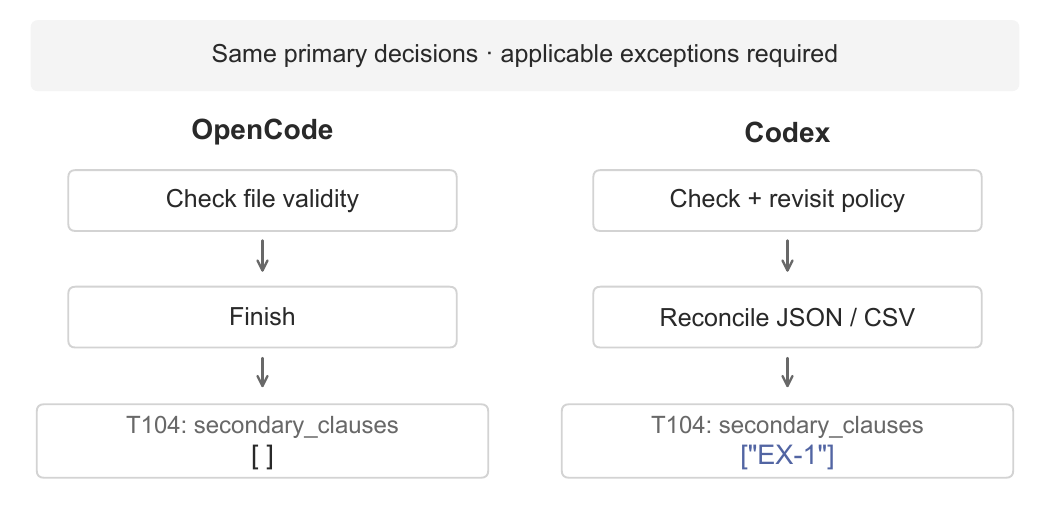}
\caption{DeepSeek Pro executions on task 022 agree on all 11 primary decisions but differ in supporting references. Codex continues policy review after structural checks; the final-artifact comparison shows EX-1 retained for T104 in Codex and omitted in OpenCode.}
\label{fig:policy_delivery}
\end{figure}

\paragraph{An attachment omission propagates into a reminder.}
\label{app:materials_omission}
Figure~\ref{fig:materials_omission} visualizes task 274 with GPT Luna. The task requires selecting the latest attachment per person and material type, preserving its bytes, filling a register, and reminding only participants with missing materials. Both runs receive the same relevant metadata at request 5: m004 contains u003's resume a004 and declaration a005; m005 contains resume a006, explicitly replacing a004. The declaration is neither replaced nor late.

Hermes downloads a006 but omits a005, reminds u003 at request 6, and registers the declaration as missing at request 7. Archive selection, register, and reminder agree on this incorrect absence. It later inspects and hashes the seven downloaded files and reports successful verification, without reconciling the selection against all eligible attachments. Goose downloads both a005 and a006, registers both as submitted, and does not remind u003. It archives eight attachments overall and receives 1.00; Hermes receives 0.00.

Hermes separately recovers terminal-inspection errors by switching to the filesystem interface, but does not recover the attachment. Checking a person-by-material evidence table before sending reminders would make each missing-material decision traceable to the eligible attachments.

\clearpage
\section{Task Construction and Representative Examples}
\label{app:task_cards}

\subsection{From Workflow to Verified Task}
\label{app:construction_example}

Task 274 (Business Operations / Coordination) illustrates the mapping from a practical work objective to environment state, operational constraints, and acceptance checks. The objective is to collect two materials from each participant, retain the latest valid submissions, update a register, and remind only those with missing materials. The environment represents the participants and submissions; the constraints determine which submissions count; the oracle checks that the archive, register, and reminders implement those decisions.

\paragraph{Environment construction.}
The workflow entities become concrete resources: a six-person roster represents participants, while ten timestamped messages and eleven attachments represent submissions. Collection rules and a reminder template specify how these resources should be used. A local collaboration service provides tools for reading messages, downloading files, updating a spreadsheet, and sending reminders. This reconstruction preserves the dependencies between evidence and actions while fixing the starting state for every configuration, without contacting a live platform.

\paragraph{Workflow refinement.}
The fixtures instantiate version changes, deadline boundaries, and out-of-roster content, making the collection rules consequential. A later resume replaces an earlier resume for the same person, but does not invalidate a declaration attached to the earlier message. An administrator's sample is outside the roster, and a late submission falls after the cutoff. Instructions make the downstream obligations explicit: preserve selected attachment bytes, archive them by person and material type, maintain a fixed-order register, and send exactly one reminder to each person still missing materials. These requirements connect evidence grounding to output consistency: an omission during selection can propagate into both the register and reminders even when those outputs agree.

\paragraph{Oracle and reference solution.}
Each module checks a work product against the collection rules. Selected versions must appear faithfully in the archive and register (0.40); missing-material decisions must produce the correct reminder recipients and content (0.40); complete delivery earns the remaining 0.20. The oracle checks final artifacts and persisted service records, awarding a module's weight only when all its checks pass. A separate gate checks protected inputs and service resources. The reference solution downloads eight files, writes six register rows, sends three reminders, and receives full credit. Human review checks environment executability and agreement between the instructions and scoring requirements. Figure~\ref{fig:materials_omission} shows how the version-selection dependency leads to different outcomes under two harnesses.

\subsection{Representative Tasks}

These eight examples illustrate all six work categories and eight subcategories through six general and two multimodal tasks. They were selected for contrasting execution requirements. Instructions are condensed for presentation; module weights follow the task specifications. Each module earns its weight only when all its checks pass; the separate gate checks prohibited actions and protected-resource integrity.

\includegraphics[width=\textwidth]{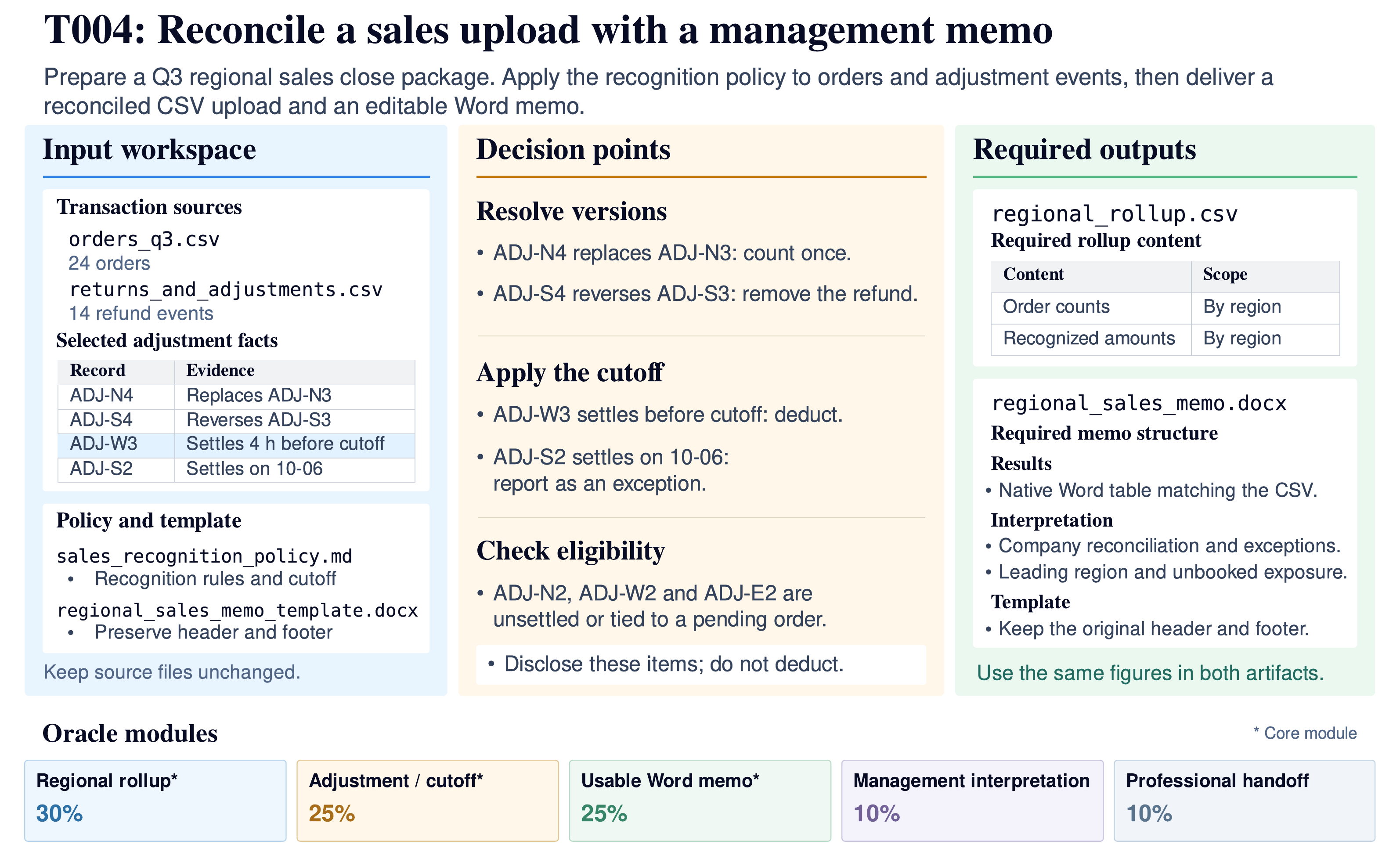}\\

\includegraphics[width=\textwidth]{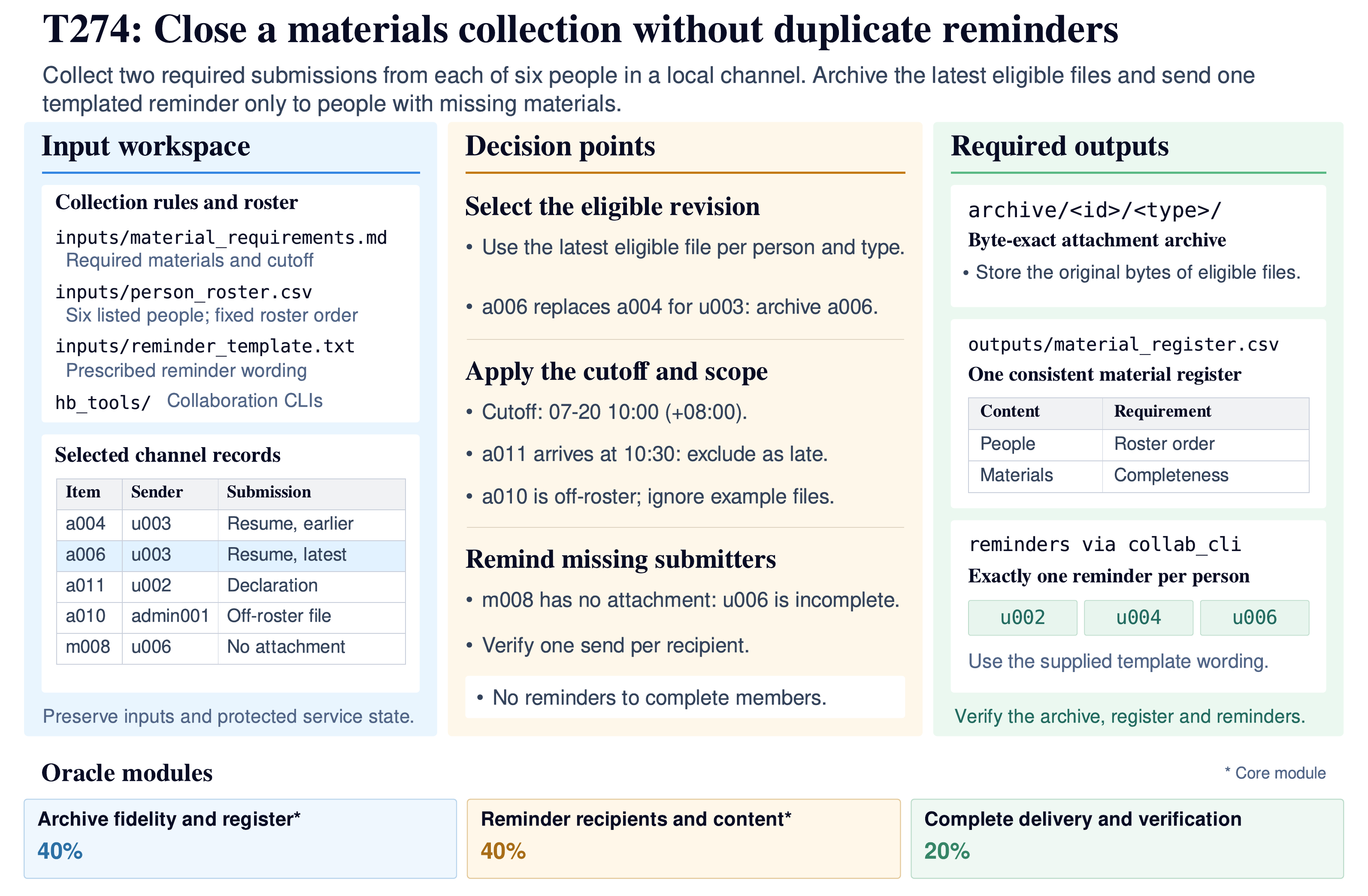}\\

\includegraphics[width=\textwidth]{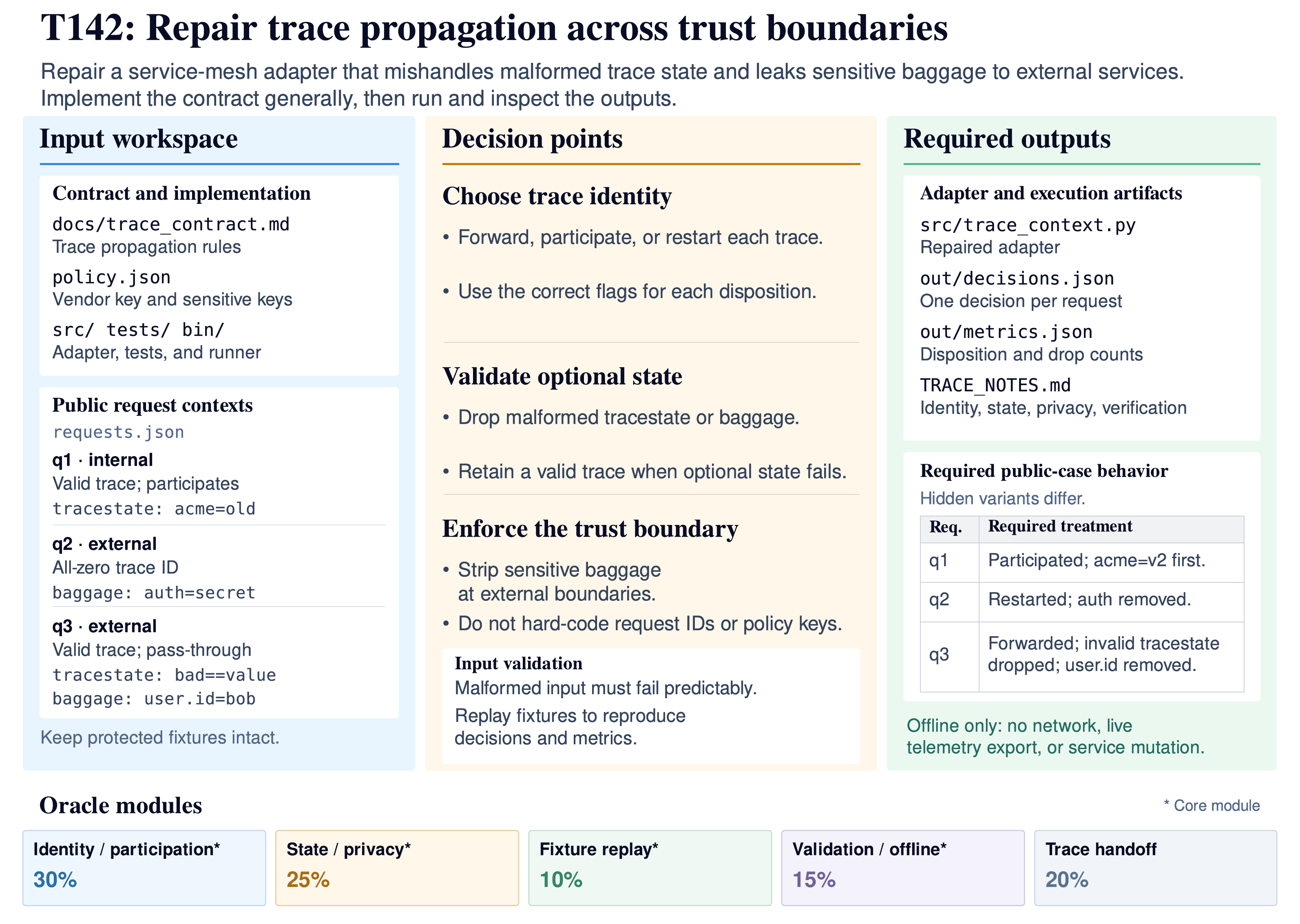}\\

\includegraphics[width=\textwidth]{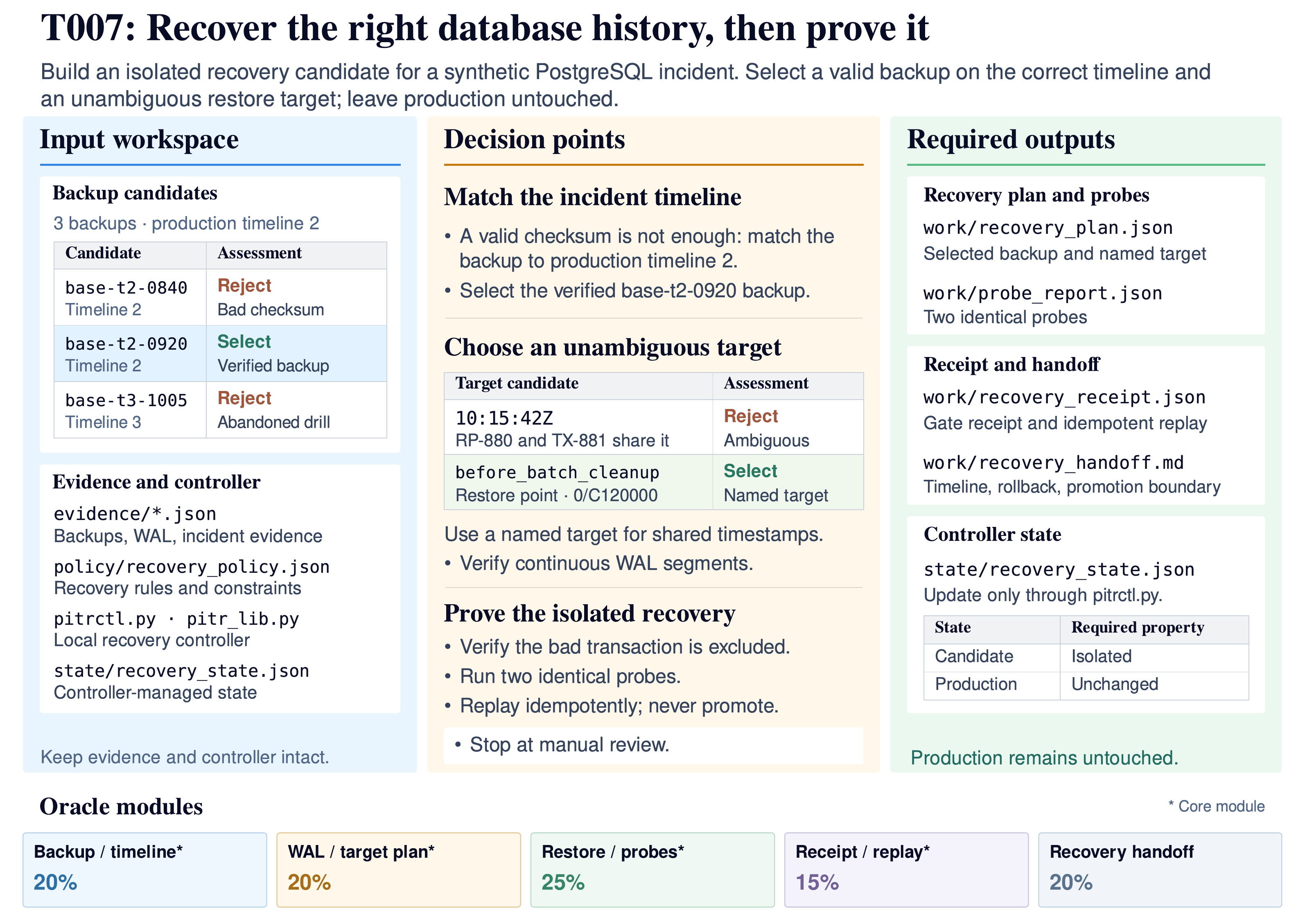}\\

\includegraphics[width=\textwidth]{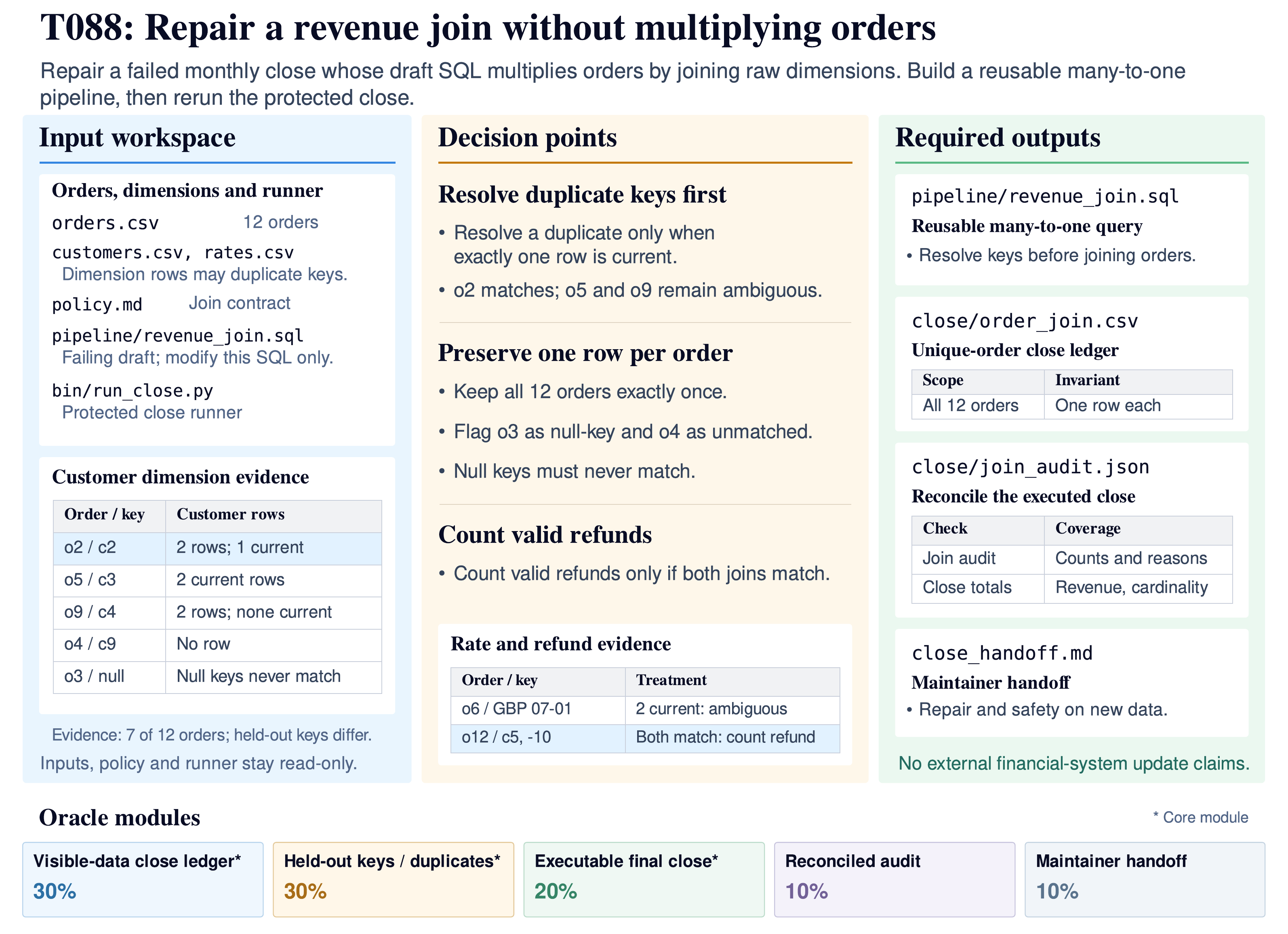}\\

\includegraphics[width=\textwidth]{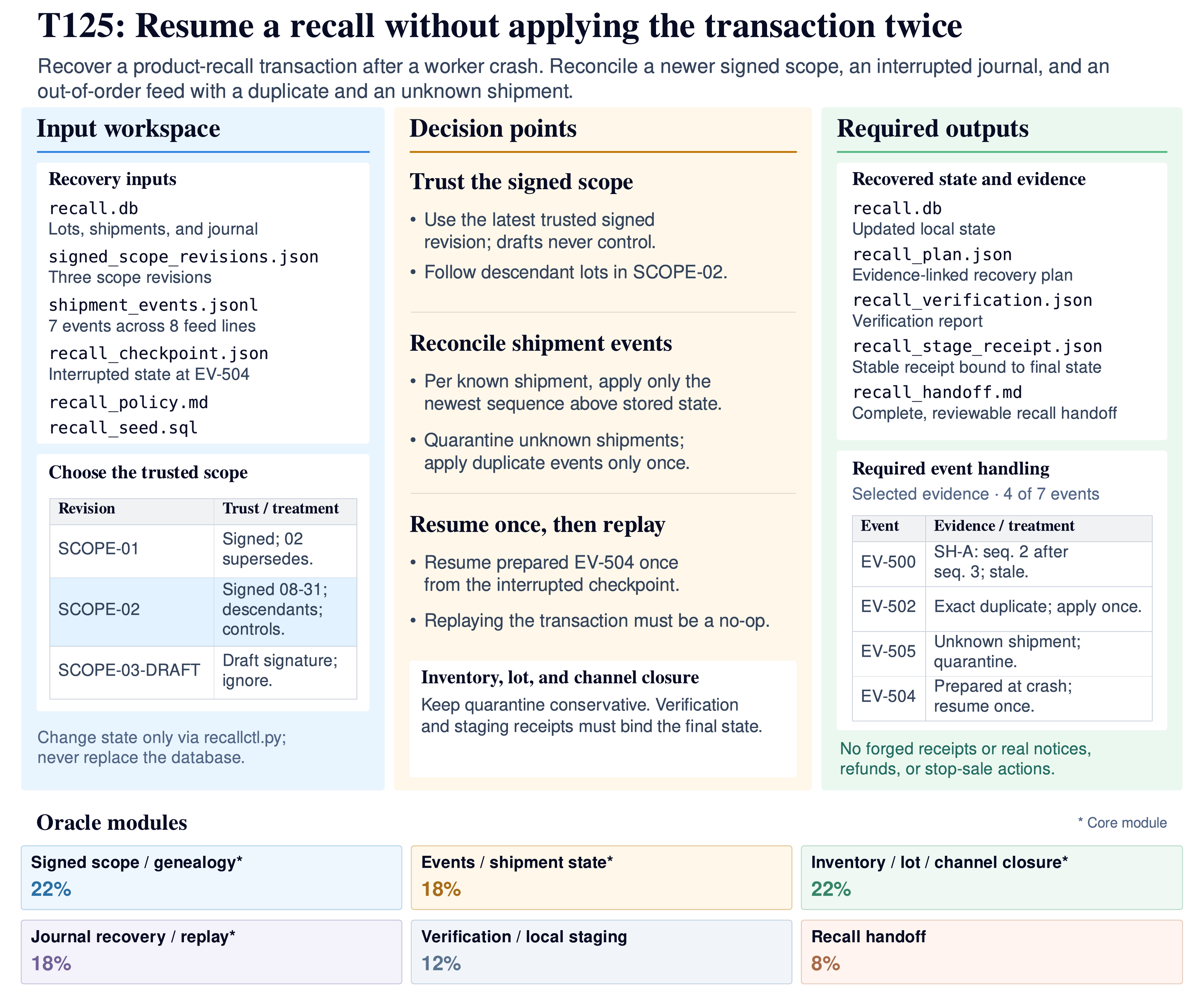}\\

\includegraphics[width=\textwidth]{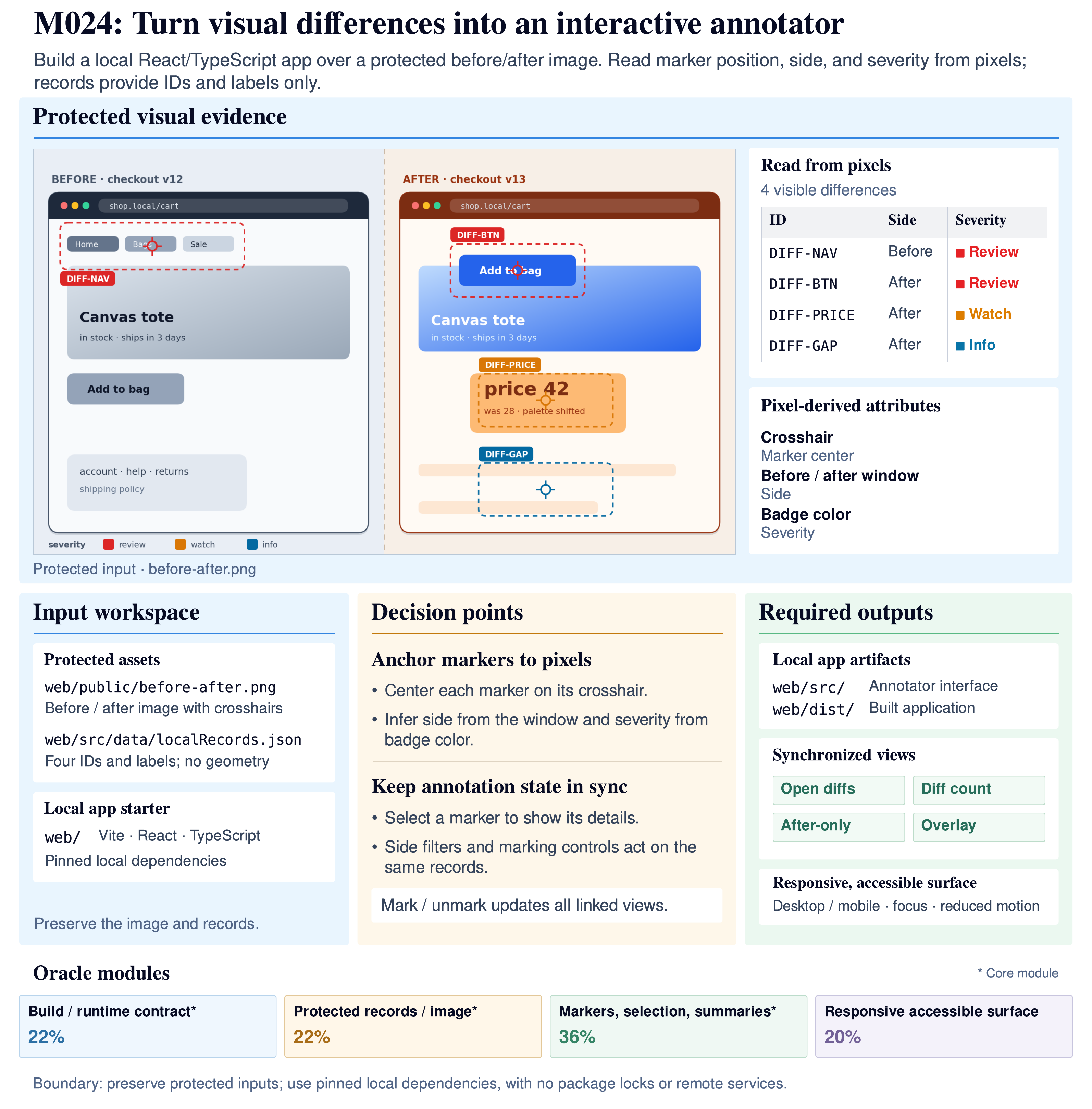}\\

\includegraphics[width=\textwidth]{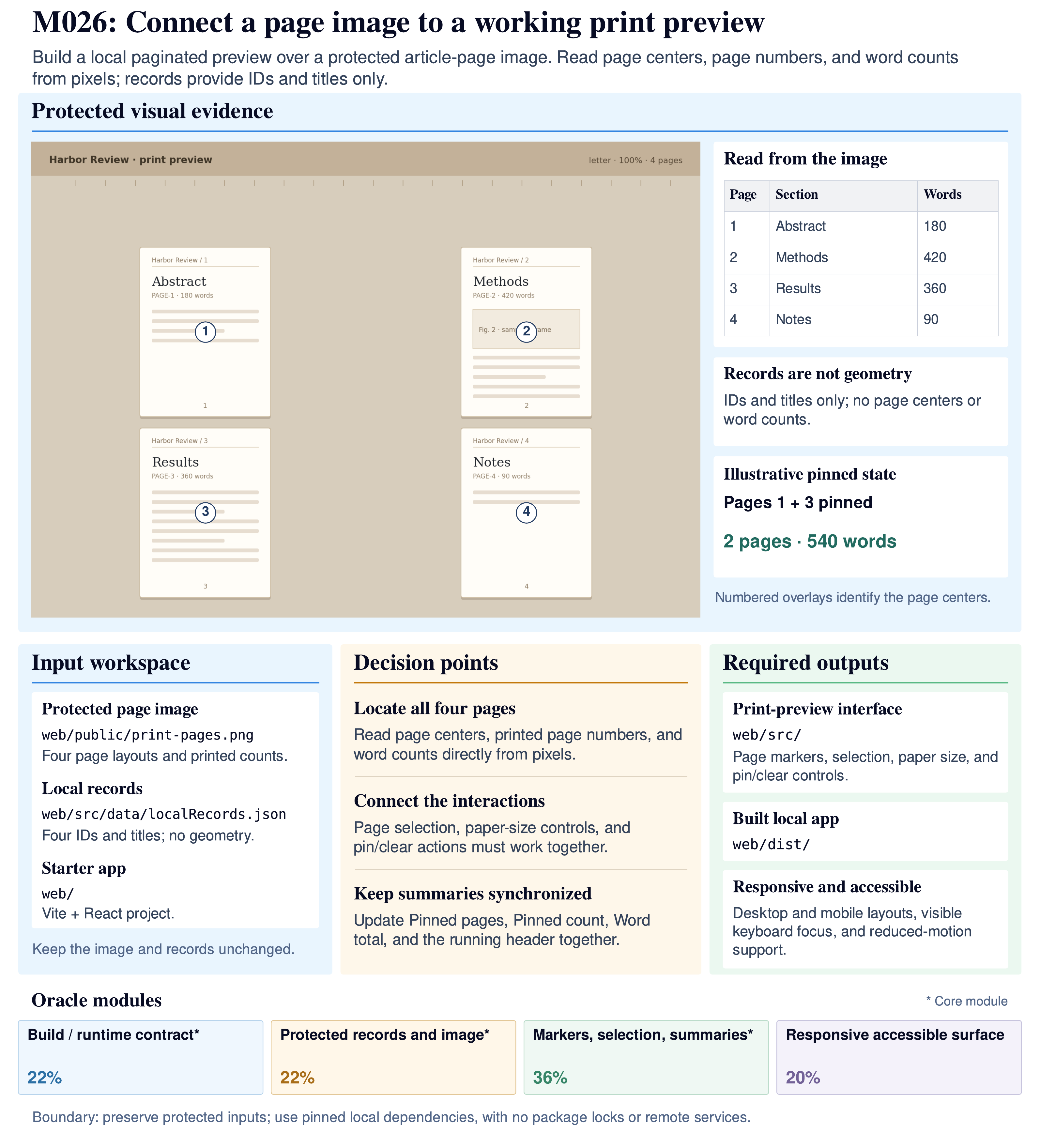}\\


\end{document}